\documentclass[manuscript,screen]{acmart}

\usepackage{natbib}
\usepackage{graphicx}
\usepackage{balance} 
\usepackage{amsmath,amsfonts}
\usepackage{color}

\usepackage{hyperref}
\usepackage{lscape}
\usepackage{array}
\usepackage{tabulary}
\usepackage{soul}
\usepackage[edges]{forest}
\usepackage{xcolor}
\usetikzlibrary{shapes.geometric}

\AtBeginDocument{%
  \providecommand\BibTeX{{%
    \normalfont B\kern-0.5em{\scshape i\kern-0.25em b}\kern-0.8em\TeX}}}

\setcopyright{acmcopyright}
\copyrightyear{2023}
\acmYear{2023}
\acmDOI{XXXXXXX.XXXXXXX}

\begin{document}

\title{Deep Learning for Time Series Classification and Extrinsic Regression: A~Current Survey}

\author[1]{Navid Mohammadi Foumani}
\authornotemark[1]
\email{navid.foumani@monash.edu.com}
\orcid{0000-0003-2475-6040}
\affiliation{%
  \institution{Monash~University}
  \country{Australia}
}
\author{Lynn Miller}
\email{lynn.miller1@monash.edu}
\affiliation{%
  \institution{Monash~University}
  \country{Australia}
}
\author{Chang Wei Tan}
\email{chang.tan@monash.edu}
\affiliation{%
  \institution{Monash~University}
  \country{Australia}
}
\author{Geoffrey I. Webb}
\email{geoff.webb@monash.edu}
\affiliation{%
  \institution{Monash~University}
  \country{Australia}
}
\author{Germain Forestier}
\email{germain.forestier@uha.fr}

\affiliation{%
  \institution{Monash~University}
  \country{Australia}
}
\affiliation{%
  \institution{IRIMAS, University of Haute-Alsace}
  \country{France}
}
\author{Mahsa Salehi}
\email{mahsa.salehi@monash.edu}
\affiliation{%
  \institution{Monash~University}
  \country{Australia}
}

\renewcommand{\shortauthors}{Foumani et al.}

\begin{abstract}
Time Series Classification and Extrinsic Regression are important and challenging machine learning tasks. Deep learning has revolutionized natural language processing and computer vision and holds great promise in other fields such as time series analysis where the relevant features must often be abstracted from the raw data but are not known a priori. This paper surveys the current state of the art in the fast-moving field of deep learning for time series classification and extrinsic regression. We review different network architectures and training methods used for these tasks and discuss the challenges and opportunities when applying deep learning to time series data. We also summarize two critical applications of time series classification and extrinsic regression, human activity recognition and satellite earth observation.


\end{abstract}






\maketitle

\section{Introduction}
Time series analysis has been identified as one of the ten most challenging research issues in the field of data mining in the 21st century~\cite{yang200610}. Time series classification (TSC) is a key time series analysis task~\cite{esling2012time}.
TSC builds a machine learning model to predict categorical class labels for data consisting of ordered sets of real-valued attributes. 
The many applications of time series analysis include human activity recognition~\cite{nweke2018deep,wang2019deep,chen2021deep}, diagnosis based on electronic health records~\cite{schirrmeister2017deep,rajkomar2018scalable}, and systems monitoring problems~\cite{bagnall2018uea}. The wide variety of dataset types in the University of California, Riverside (UCR)~\cite{dau2019ucr} and University of East Anglia (UEA)~\cite{bagnall2018uea} benchmark archive further illustrates the breadth of TSC applications. Time series extrinsic regression (TSER)~\cite{tan2021time} is the counterpart of TSC for which the output is numeric rather than categorical. It should be noted that the TSER is not a forecasting method but rather a method for understanding the relationship between the time series and the extrinsic variable. TSER is an emerging field with great potential to be used in a wide range of applications.

Deep learning has been very successful, especially in computer vision and natural language processing. Many modern applications integrate deep learning. 
Deep learning can autonomously learn informative features from raw data, eliminating the need for manual feature engineering. Consequently, there has been much interest in developing deep TSC and TSER due to their ability to learn relevant latent feature representations. It is worth noting that the majority of TSC and TSER research has focused on non-deep learning approaches. A recent benchmark \cite{middlehurst2023bake} shows that the deep learning method (InceptionTime \cite{fawaz2020inceptiontime}) is competitive but did not outperform the state of the art on benchmarking archives. 
One reason is that the popular UCR and UEA benchmarking archives were not designed for deep learning models. In particular, they are relatively small, while deep learning often excels when data quantities are large. 
Deep learning can also benefit from heightened compatibility with current hardware, particularly GPUs, leading to fast and efficient execution. Their exceptional scalability further allows seamless handling of growing data volumes and computational complexity, reinforcing their versatility in processing large datasets. 
Indeed, ConvTran~\cite{Foumani2023}, a recent deep architecture for TSC, outperforms one of the fastest conventional models, ROCKET~\cite{dempster2019rocket}, in terms of both speed and accuracy when there are more than 10k training samples.


A highly influential review paper on deep learning-based TSC~\cite{fawaz2019deep} was published in 2019. However, the field of research is very fast-moving, and that prior survey does not cover the current state of the art. For example, it does not include InceptionTime~\cite{fawaz2020inceptiontime}, a system that consistently outperforms ResNet~\cite{wang2017time}, the best performing system from the prior survey.
Nor does it cover attention models, which have received huge interest in recent years and have shown excellent capacity to model long-range dependencies in sequential data, and are well suited for time-series modeling~\cite{wen2022transformers}. Many attention variants have been proposed to address particular challenges in time series modeling and have been successfully applied to TSC~\cite{hao2020new,zerveas2021transformer,Foumani2023}. Moreover, the previous survey does not include self-supervised learning, which is emerging as a new paradigm~\cite{liu2021self}. Self-supervised learning induces supervision by designing pretext tasks instead of relying on predefined prior knowledge and has shown very promising results, especially in datasets with a low label regime~\cite{eldele2021time,yang2021voice2series,yue2022ts2vec,foumani2023series2vec}.

In light of the emergence of attention mechanisms, self-supervised learning, and various new network configurations for TSC, a systematic and comprehensive survey on deep learning in TSC would greatly benefit the time series community. This article aims to fill that gap by summarizing recent developments in deep learning-based time series analytics, specifically TSC and TSER. Following definitions and a brief introduction to the time series classification and extrinsic regression tasks, we propose a new taxonomy based on various methodological perspectives. Diverse architectures, including multilayer perceptrons (MLP), convolutional neural networks (CNN), recurrent neural networks (RNN), Graph Neural Network (GNN), and attention-based models, are discussed, along with refinements made to improve performance. Additionally, various types of self-supervised learning pretexts, such as contrastive learning and self-prediction, are explored. We also conduct a review of useful data augmentation and transfer learning strategies for time series data. Furthermore, we provide a summary of two key applications of TSC and TSER, namely Human Activity Recognition and Earth Observation.

\section{Background and Definitions}
This section begins by providing the necessary definitions and background information to understand the topic of training deep neural networks (DNNs) for TSC and TSER tasks. We begin by defining key terms and concepts, such as time series data and time series supervised learning. 
Finally, we present our proposed taxonomy of the different deep learning methods that have been used for TSC and TSER tasks. 

\subsection{Time series}
Time series data are sequences of data points indexed by time.
\vspace{-0.2cm}
\begin{definition}
\label{def:time series}
A time series $X$ is an ordered collection of $T$ pairs of measurements and timestamps,

$X=\{(x_1,t_1),(x_2,t_2), ..., (x_T,t_T)\}$, 
where $x_i\in\mathbb{R}^D$ and $t_1$ to $t_T$ are the timestamps for some measurements $x_1$ to $x_T$. 
\end{definition}

\noindent

Each $x_i$ is a $D$-dimensional vector of values, one for each feature captured in the series.  When $D=1$ the series is called \emph{univariate}.  When $D>1$ the series is called \emph{multivariate}.

\subsection{Time series supervised learning tasks}

This paper focuses on two time series learning tasks: time series extrinsic regression and time series classification.
Classification and regression are both supervised learning tasks that learn the relationship between a target variable and a set of time series.
We consider learning from a dataset $D=\left\{(X_1,Y_1), (X_2,Y_2),...,(X_N,Y_N)\right\}$ of $N$ time series where $Y_i$ denotes the target variable for each $X_i$. It is important to note that for ease of exposition, we assume in our discussion that the series are of the same length, but most methods extend trivially to the case of unequal-length series.
The main difference between TSER and TSC is that TSC predicts a categorical value for a time series from a set of finite categories, while TSER predicts a continuous value for a variable external to the input time series.
Typically $Y_i$ is a one hot encoded vector for TSC or a numeric value for TSER. 

In the context of deep learning, a supervised learning model is a neural network that executes the following functions to map the input time series to a target variable:

\vspace{-0.2cm}
\begin{equation}
f_L(\theta_L, X) = f_{L-1}(\theta_{L-1}, f_{L-2}(\theta_{L-2}, \ldots, f_1(\theta_1, X)))
\end{equation}

\noindent where $f_i$ represents the non-linear function and $\theta_i$ denotes the parameters at layer $i$. For TSC the neural network model is trained to map a time series dataset $D$ to a set of class labels $Y$ with $C$ class labels.
After training, the neural network outputs a vector of $C$ values that estimates the probability of a series $X$ belonging to each class.
This is typically achieved using the softmax activation function in the final layer of the neural network.
The softmax function estimates probabilities for all of the dependent classes such that they always sum to 1 across all classes.
The cross-entropy loss is commonly used for training neural networks with softmax outputs or classification type neural networks.

On the other hand, TSER trains the neural network model to map a time series dataset $D$ to a set of numeric values $Y$. 
Instead of outputting probabilities, a regression neural network outputs a numerical value for the time series. 
It is typically used with a linear activation function in the final layer of the neural network.
However, any non-linear functions with a single value output such as sigmoid, or ReLU can also be used.
A regression neural network typically trains using the mean square error or mean absolute error loss function.
However, depending on the distribution of the target variable and the choice of final activation functions, other loss functions can be used.

\subsection{TSC and TSER}

TSC is a fast-growing field, with hundreds of papers being published every year~\cite{bagnall2018uea,dau2019ucr,bagnall2017great,fawaz2019deep,ruiz2020great}. 
The majority of work in TSC are non-deep learning based. 
In this survey, we focus on deep learning approaches and refer interested readers to Appendix~\ref{App:non-deep} and benchmark papers \cite{middlehurst2023bake,bagnall2017great,ruiz2020great} for more details on non-deep learning approaches.
Most deep learning approaches to TSC have real-valued outputs that are mapped to a class label. TSER~\cite{tan2021time,tan2020monash} is a less widely studied task in which the predicted values are numeric, rather than categorical. 
While the majority of the architectures covered in this survey were designed for TSC, it is important to note that it is trivial to adapt most of them for TSER.

Deep learning-based TSC methods can be classified into two main types: generative and discriminative~\cite{langkvist2014review}. 
In the TSC community, generative methods are often considered model-based~\cite{bagnall2017great}, aiming to understand and model the joint probability distribution of input series $X$ and output labels $Y$, denoted as $p(X, Y)$. On the other hand, discriminative models focus on modeling the conditional probability of output labels $Y$ given input series $X$, expressed as $p(Y | X)$.

Generative models, such as the Stacked Denoising Auto-encoders (SDAE) have been proposed by Bengio et al.~\cite{bengio2013generalized} to identify the salient structure of input data distributions, and Hu et al.~\cite{hu2016transfer} used the same model for the pre-training phase before training a classifier for time series tasks. 
A universal neural network encoder has been developed to convert variable-length time series to a fixed-length representation~\cite{serra2018towards}. 
Also, a Deep Belief Network (DBN) combined with a transfer learning method was used in an unsupervised manner to model the latent features of time series~\cite{banerjee2019deep}. 
An Echo State Network (ESN) has been used to learn the appropriate time series representation by reconstructing the original raw time series prior to training the classifier \cite{aswolinskiy2018time}.
Generative Adversarial Networks (GANs) are one of the popular generative models that generate new examples by learning to discriminate between real and synthetic examples.
Various GANs have been developed for time series and have been reviewed in a recent survey \cite{GanSurvey2021}.
Often, implementing generative methods is more complex due to an additional step of training. 
Furthermore, generative methods are typically less efficient than discriminative methods, which directly map raw time series to class probability distributions.
Due to these barriers, researchers tend to focus on discriminative methods. 
Therefore, this survey mainly focuses on the end-to-end discriminative approaches.

\subsection{Taxonomy of Deep Learning in TSC and TSER}

\begin{figure}
\begin{forest}
  for tree={
    parent anchor=children,
    child anchor=parent,
    anchor=north,
    draw,
    align=left,
    inner sep=1.75pt,
    rounded corners=2pt, 
    font=\footnotesize,
  },
  where level=0{s sep=10pt,fill=red!5!white!80!green!40}{},
  where level=1{s sep=6pt,fill=red!60!white!80!yellow!40}{},
  where level=2{s sep=2pt,fill=red!40!white!80!blue!40}{},
  where level=3{s sep=2pt,fill=red!10!white!80!blue!40}{},
  where level=4{fill=red!5!white!80!green!40}{},
  forked edges,
  [\textbf{Deep Learning methods for Time Series Classification and Extrinsic Regression},
      [Supervised (Sec.\ref{sec:sup}), name=S, 
        s sep=6pt,
        [Multi-Layer\\Perceptron, name=MLP,rotate=0,anchor=north]
        [Convolutional\\Neural Network, name=CNN, for tree={grow'=0,folder,draw}, 
          [Adapted\\Convolutional\\Neural Network, name=ACNN]
          [Imaging Time\\ Series, name=ITS]
          [Multi-Scale\\ Operation, name=MSO]
        ]
        [Recurrent\\Neural\\Network, name=RNN,for tree={grow'=0,folder,draw}, 
          [Vanilla Recurrent\\Neural Network, name=RRN]
          [Long Short Term Memory, name=LSTM]
          [Gated Recurrent Unit, name=GRU]
          [Hybrid, name=RCNN]
        ]
        [Graph Neural\\Network, name=GNN,rotate=0,anchor=north]
        [Attention, name=Attn, for tree={grow'=0,folder,draw}, 
          [Self-Attention, name=SA]
          [Transformers, name=Trans]
        ]
      ]
      [Self-Supervised \\ (Sec.\ref{sec:ss}), name=SS, for tree={grow'=0,folder,draw}, 
        [Self-Prediction, name=SSCNN]
        [Contrastive-\\Learning, name=SSAttn]
        [Other pretext\\ tasks, name=SSGNN, 
        ]
      ]
      [Data \\ Augmentation \\(Sec.\ref{sec:aug}), name=DA, for tree={grow'=0,folder,draw}, 
          [Random\\ Transformations, name=RT]
          [Window methods, name=WM]
          [Averaging methods, name=AM]
        ]
      [Transfer \\ Learning \\(Sec.\ref{sec:TL}), name=TL,rotate=0,anchor=north ]
  ]
\end{forest}
\caption{Taxonomy of Deep Learning (DL) for TSC/TSER from the perspectives of network configuration and application domains.} 
\label{fig:TAX}
\vspace{-0.5cm}
\end{figure}
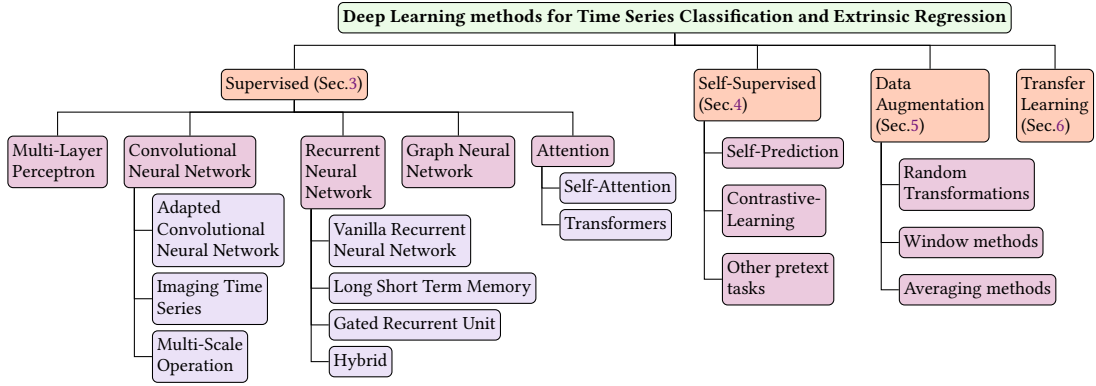


To provide an organized summary of the existing deep learning models for TSC, we propose a taxonomy that categorizes these models based on deep learning methods and application domains. This taxonomy is illustrated in Fig.~\ref{fig:TAX}. In section \ref{sec:sup}, we review various network architectures used for TSC, including multilayer perceptrons, convolutional neural networks, recurrent neural networks, graph neural networks, and attention-based models. We also discuss refinements made to these models to improve their performance on time series tasks. Additionally, various types of self-supervised learning pretexts, such as contrastive learning and self-prediction, are explored in section \ref{sec:ss}. We also conduct a review of useful data augmentation and transfer learning strategies
for time series data in section~\ref{sec:aug} and ~\ref{sec:TL}.
In addition to methods, we summarize key applications of TSC and TSER in section \ref{sec:app} of this paper. These applications include human activity recognition and satellite earth observation, which are important and challenging tasks that can benefit from the use of deep learning models. Overall, our proposed taxonomy and the discussions in these sections provide a comprehensive overview of the current state of the art in deep learning for time series analysis and outline future research directions.


\vspace{-0.3cm}
\section{Supervised Models} \label{sec:sup}
This section reviews the deep learning-based models for TSC and discusses their architectures by highlighting their strengths as well as limitations. More details on deep model architectures and their adaptations to time series data are available in Appendix~\ref{APx:DNN}.
\vspace{-0.3cm}
\subsection{Multi-Layer Perceptron (MLP)} \label{sec:MLP}
The most straightforward neural network architecture is a fully connected network (FC), also called a multilayer perceptron (MLP). The number of layers and neurons are defined as hyperparameters in MLP models. However, studies such as auto-adaptive MLP~\cite{del2021auto} have attempted to determine the number of neurons in the hidden layers automatically, based on the nature of the training time series data. This allows the network to adapt to the training data's characteristics and optimize its performance on the task at hand.

One of the main limitations of using multilayer perceptrons (MLPs) for time series data is that they are not well-suited to capturing the temporal dependencies in this type of data. MLPs are feedforward networks that process input data in a fixed and predetermined order without considering the temporal relationships between the input values. Various studies used MLPs alongside other feature extractors like Dynamic Time Warping (DTW) to address this problem~\cite{iwana2016robust,iwana2020dtw}. DTW-NN is a feedforward neural network that exploits DTW's elastic matching ability to dynamically align a layer's inputs to the weights instead of using a fixed and predetermined input-to-weight mapping. This weight alignment replaces the standard dot product within a neuron with DTW. In this way, the DTW-NN is able to tackle difficulties with time series recognition, such as temporal distortions and variable pattern length within a feedforward architecture~\cite{iwana2020dtw}.
Similarly, Symbolic Aggregate Approximation (SAX) is used to transform time series into a symbolic representation and produce sequences of words based on the symbolic representation~\cite{tabassum2022time}. The symbolic time series-based words are later used as input for training a two-layer MLP  for classification.

Although the models mentioned above attempt to resolve the shortage of capturing temporal dependencies in MLP models, they still have other limitations on capturing time-invariant features~\cite{wang2017time}. 
Additionally, MLP models do not have the ability to process input data in a hierarchical or multi-scale manner. Time series data often exhibits patterns and structures at different scales, such as long-term trends and short-term fluctuations. MLP models fail to capture these patterns, as they are only able to process input data in a single, fixed-length representation. In addition, MLPs may encounter difficulties when confronted with irregularly sampled time series data, where observations are not uniformly recorded in time. Many other deep learning models are better suited to handle time series data, such as recurrent neural networks (RNNs), convolutional neural networks (CNNs), and transformers, specifically designed to capture the temporal dependencies and patterns in time series data.

\vspace{-0.3cm}
\subsection{CNN based models}
Several improvements have been made to CNN since the success of AlexNet in 2012~\cite{krizhevsky2012imagenet} such as using deeper networks, applying smaller and more efficient convolutional filters, adding pooling layers to reduce the dimensionality of the feature maps, and utilizing batch normalization to improve the stability of training~\cite{gu2018recent}. They have been demonstrated to be very successful in many domains, such as computer vision, speech recognition, and natural language processing problems~\cite{lecun2015deep,gu2018recent}.
As a result of the success of CNN architectures in these various domains, researchers have also started adopting them for TSC. See table~\ref{tab:cnn} for a list of reviewed CNN models in this paper.

\subsubsection{Adapted CNNs for TSC and TSER}
This section presents the first category, which we refer to as Adapted CNNs for TSC and TSER. The papers discussed here are mostly adaptations without any particular preprocessing or mathematical characteristics, such as transforming the series to an image or using multi-scale convolution and therefore do not fit into one of the other categories. 

The first CNN for TSC was the Multi-Channel Deep Convolutional Neural Network (MC-DCNN)~\cite{zheng2014time}. 
It handles multivariate data by independently applying convolutions to each input channel. Each input dimension undergoes two convolutional stages with ReLU activation, followed by max pooling. The output from each dimension is concatenated and passed to a fully connected layer which is then fed to a final softmax classifier for classification. Similar to MC-DCNN, a three-layer convolution neural network was proposed for Human activity recognition (MC-CNN)\cite{yang2015deep}. Unlike the MC-DCNN, this model applies 1D convolutions to all input channels simultaneously to capture the temporal and spatial relationships in the early stages. The 2-stage version of MC-CNN architecture was used by Zhao et al.~\cite{zhao2017convolutional} on the earliest version of the UCR Time Series Data Mining Archive. The authors also conducted an ablation study to evaluate the performance of the CNN models with differing numbers of convolution filters and pooling types.

Fully Convolutional Networks (FCN)~\cite{long2015fully}, and Residual Networks (Resnet)~\cite{he2016deep} are two deep neural networks that are commonly used for image and video recognition tasks and have been adapted for end-to-end TSC~\cite{wang2017time}. FCNs are a variant of convolutional neural networks (CNNs) designed to operate on inputs of arbitrary size rather than being constrained to fixed-size inputs like traditional CNNs. This is achieved by replacing the fully connected layers in a traditional CNN with a Global Average Pooling (GAP)~\cite{long2015fully}. 
FCN was adapted for univariate TSC~\cite{wang2017time}, and similar to the original model, it contains three convolution blocks where each block contains a convolution layer followed by batch normalization and ReLU activation. Each block uses 128, 256, 128 filters with 8, 5, 3 filter lengths respectively. The output from the last convolution block is averaged with a GAP layer and passed to a final softmax classifier. 
The GAP layer has the property of reducing the spatial dimensions of the input while retaining the channel-wise information, which allows it to be used in conjunction with a class activation map (CAM)~\cite{zhou2016learning} to highlight the regions in the input that are most important for the predicted class. This can provide useful insights into how the network is making its predictions and help identify potential improvement areas. Similar to FCN, the Residual Network (ResNet) was also proposed in~\cite{wang2017time} for univariate TSC. ResNet is a deep architecture containing three residual blocks followed by a GAP layer and a softmax classifier.
It uses residual connections between blocks to reduce the vanishing gradient effect that affects deep learning models. The structure of each residual block is similar to the FCN architecture, containing three convolution layers followed by batch normalization and ReLU activation. Each convolution layer uses 64 filters with 8, 5, 3 filter lengths, respectively. ResNet was found to be one of the most accurate deep learning TSC architectures on 85 univariate TSC datasets~\cite{fawaz2019deep,bagnall2017great}. Additionally, integration of ResNet and FCN has been proposed to combine the strength of both networks~\cite{zou2019integration}.

In addition to adapting the network architecture, some research has focused on modifying the convolution kernel to suit TSC tasks better. Dilated convolutions neural networks (DCNNs)~\cite{li2018csrnet} are a type of CNN that uses dilated convolutions to increase the receptive field of the network without increasing the number of parameters. Dilated convolutions create gaps between elements of the kernel and perform convolution, thereby covering a larger area of the input. This allows the network to capture long-range dependencies in the data, making it well-suited to TSC tasks~\cite{yazdanbakhsh2019multivariate}. Recently, Disjoint-CNN~\cite{foumani2021disjoint} showed that factorization of 1D convolution kernels into disjoint temporal and spatial components yields accuracy improvements with almost no additional computational cost. Applying disjoint temporal convolution and then spatial convolution behaves similarly to Inverted Bottleneck~\cite{sandler2018mobilenetv2}. Like the Inverted Bottleneck, the temporal convolutions expand the number of input channels, and spatial convolutions later project the expanded hidden state back to the original size to capture the temporal and spatial interaction. 

\subsubsection{Imaging time series}
In TSC, a common approach is to convert the time series data into a fixed-length representation, such as a vector or matrix, which can then be input to a deep learning model. However, this can be challenging for time series data that vary in length or have complex temporal dependencies. One solution to this problem is to represent the time series data in an image-like format, where each time step is treated as a separate channel in the image. This allows the model to learn from the spatial relationships within the data rather than just the temporal relationships. In this context, the term \textit{spatial} refers to the relationships between different variables or features within a single time step of the time series.

As an alternative to using raw time series data as input, Wang and Oates encoded univariate time series data into different types of images that were then processed by a regular CNN~\cite{wang2015encoding}. This image-based framework initiated a new branch of deep learning approaches for time series, which consider image transformation as one of the feature engineering techniques. Wang and Oates presented two approaches for transforming a time series into an image. The first generates a Gramian Angular Field (GAF), while the second generates a Markov Transition Field (MTF). GAF represents time series data in a polar coordinate and uses various operations to convert these angles into a symmetry matrix and MTF encodes the matrix entries using the transition probability of a data point from one time step to another time step~\cite{wang2015encoding}. In both cases, the image generation increases the time series size, making the images potentially prohibitively large. Therefore they propose strategies to reduce their size without losing too much information. Afterward, the two types of images are combined in a two-channel image that is then used to produce better results than those achieved when using each image separately. Finally, a Tiled CNN model is applied to classify the time-series images. In other studies, a variety of transformation methods, including Recurrence Plots (RP)~\cite{hatami2018classification}, Gramian Angular Difference Field (GADF)~\cite{karimi2018scalable}, bilinear interpolation~\cite{zhao2019classify}, and Gramian Angular Summation Field (GASF)~\cite{yang2019sensor} have been proposed to transfer time series to input images expecting that the two-dimensional images could reveal features and patterns not found in the one-dimensional sequence of the original time series.

Hatami et al.~\cite{hatami2018classification} propose a representation method based on RP~\cite{kamphorst1987recurrence} to convert the time series to 2D images with a CNN model for TSC. In their study, time series are regarded as distinct recurrent behaviors such as periodicities and irregular cyclicities, which are the typical phenomena of dynamic systems. The main idea of using the RP method is to reveal at which points some trajectories return to a previous state. Finally, two-stage convolution and two fully connected layers are applied to classify the images generated by RP. Subsequently, pre-trained Inception v3~\cite{szegedy2016rethinking} was used to map the GADF images into a 2048-dimensional vector space.  The final stage used an MLP with three hidden layers, followed by a softmax activation function ~\cite{karimi2018scalable}.
Following the same framework, Chen and Shi~\cite{chen2019deep} adopted the Relative Position Matrix and VGGNet (RPMCNN) to classify time series data using transform 2D images. Their results showed promising performances by converting univariate time series data to 2D images using relative positions between two time stamps. Following the convention, three image encoding methods: GASF, GADF, and MTF, were used to encode MTS data into two-dimensional images~\cite{yang2019sensor}. They showed that the simple structure of ConvNet is sufficient for classification as it performed equally well with the complex structure of VGGNet. 

Overall, representing time series data as 2D images can be difficult because preserving the temporal relationships and patterns in the data can be challenging. This transformation can also result in a loss of information, making it difficult for the model to classify the data accurately. Chen and Shi~\cite{chen2019deep} have also shown that the specific transformation methods like  GASF, GADF, and MTF used in this process do not significantly improve the prediction outcome.
 
 \begin{table}[]
\setlength{\tabcolsep}{9pt}
\begin{tabular}{llcl}
\hline
Model & \multicolumn{1}{c}{Year} & Baseline Architecture & Other   features \\ \hline
\multicolumn{4}{l}{\textbf{Adapted}} \\ \hline
MC-DCNN~\cite{zheng2014time} & \multicolumn{1}{c}{2014} & 2-Stage Conv & Independent convolutions per Channel \\
MC-CNN~\cite{yang2015deep} & \multicolumn{1}{c}{2015} & 3-Stage Conv & 1D-Convolutions on all Channel \\
Zhao et al.~\cite{zhao2017convolutional} & \multicolumn{1}{c}{2015} & 2-Stage Conv & Similar architecture to MC-CNN \\
FCN~\cite{wang2017time} & \multicolumn{1}{c}{2017} & FCN & Using GAP instead of FC Layer  \\
ResNet~\cite{wang2017time} & \multicolumn{1}{c}{2017} & ResNet 9 & Using 3-Residual block \\ 
Res-CNN~\cite{zou2019integration} & \multicolumn{1}{c}{2019} & RezNet+FCN & Using 1-Residual block + FCN  \\ 
DCNNs~\cite{yazdanbakhsh2019multivariate} & \multicolumn{1}{c}{2019} & 4-Stage Conv & Using dilated convolutions \\ 
Disjoint-CNN~\cite{wang2017time} & \multicolumn{1}{c}{2021} & 4-Stage Conv & Disjoint temporal and spatial convolution \\

\hline
\multicolumn{4}{l}{\textbf{Series To Image}} \\ \hline
Wang\&Oates\cite{wang2015encoding} & \multicolumn{1}{c}{2015} & Tiled CNN & GAF, MT \\
Hatami et al.\cite{hatami2018classification} & \multicolumn{1}{c}{2017} & 2-Stage Conv & Recurrence Plots \\
Karimi et al.\cite{karimi2018scalable} & \multicolumn{1}{c}{2018} & Inception V3 & GADF \\
Zhao et al.~\cite{zhao2019classify} & \multicolumn{1}{c}{2019} & ResNet18, ShuffleNet V2 & Bilinear interpolation \\
RPMCNN~\cite{chen2019deep} & \multicolumn{1}{c}{2019} & VGGNet, 2-Stage Conv & Relative Position Matrix \\
Yang et al.~\cite{yang2019sensor} & 2019 & VGGNet & GASF, GADF, MTF\\ \hline

\multicolumn{4}{l}{\textbf{Multi-Scale Operation}} \\ \hline
MCNN~\cite{cui2016multi}& 2016 & 2-Stage Conv & Identity mapping, Smoothing, Down-sampling \\
t-LeNet~\cite{le2016data} & 2016 & 2-Stage Conv & Squeeze and Dilation \\
MVCNN~\cite{liu2018time}   & 2019 & 4-stage Conv & Inception V1 based \\ 
Brunel et al.~\cite{brunel2019cnn} & 2019 & Inception V1 &  \\
InceptionTime~\cite{fawaz2020inceptiontime} & 2019 & Inception V4 & Ensemble \\
EEG-inception~\cite{sun2021prototypical} & 2021 & InceptionTime &  \\
Inception-FCN~\cite{usmankhujaev2021time} & 2021 & InceptionTime + FCN &  \\
KDCTime~\cite{gong2022kdctime} & 2022 & InceptionTime & Knowledge Distillation, Label smoothing \\
LITE~\cite{ismail2023lite} & 2023 & InceptionTime & Multiplexing, dilated, and custom filters \\ \hline

\end{tabular}%
\caption{Summary of CNN models for time series classification and extrinsic regression}
\vspace{-1cm}
\label{tab:cnn}
\end{table}
\subsubsection{Multi-Scale Operation}
The papers discussed here apply a multi-scale convolutional kernel to the input series or apply regular convolutions on the input series at different scales.
Multi-scale CNNs (MCNN)~\cite{cui2016multi} and Time LeNet (t-LeNet)~\cite{le2016data} were considered the first models that preprocess the input series to apply convolution on multi-scale series rather than raw series. The design of both MCNNs and t-LeNet were inspired by computer vision models, which means that they were adapted from models originally developed for image recognition tasks. These models may not be well-suited to TSC tasks and may not perform as well as models specifically designed for this purpose. One potential reason for this is the use of progressive pooling layers in these models, commonly used in computer vision models, to reduce the input data size and make it easier to process. However, these pooling layers may not be as effective when applied to time series data and may limit the performance of the model.

MCNN has simple architecture and comprises two convolutions and a pooling layer, followed by a fully connected and softmax layer. However, this approach involves heavy data preprocessing.  Specifically, before any training, they use a sliding window to extract a time series subsequence, and later, the subsequence will undergo three transformations: (1) identity mapping, (2) down-sampling, and (3) smoothing, which results in the transformation of a univariate input time series into a multivariate one. Finally, the transformed output is fed to the CNN model to train a classifier~\cite{cui2016multi}. 
t-LeNet uses two data augmentation techniques: window slicing (WS) and window warping (WW), to prevent overfitting~\cite{le2016data}. The WS method is identical to MCNN’s data augmentation. The second data augmentation technique, WW, employs a warping technique that squeezes or dilates the time series. WS is also adopted to ensure that subsequences of the same length are extracted for training the network to deal with multi-length time series. Therefore, a given input time series of length $L$ is first dilated $(\times 2)$ and then squeezed $(\times 1/2 )$ using WW, resulting in three time series of length $L,2L,1/2L$ that are fed to WS to extract equal length subsequences for training.
Finally, as both MCNN and t-LeNet predict a class for each extracted subsequence, majority voting is applied to obtain the class prediction for the full time series.

Inception was first proposed by Szegedy et al.~\cite{szegedy2015going} for end-to-end image classification. Now the network has evolved to become Inception-v4, where Inception was coupled with residual connections to improve further the performance~\cite{szegedy2017inception}. Inspired by inception architecture, a multivariate convolutional neural network (MVCNN) is designed using multi-scale convolution kernels to find the optimal local construction~\cite{liu2018time}. MVCNN uses three scales of filters, including $2\times 2$, $3 \times 3$, and $5 \times 5$, to extract features of the interaction between sensors.
A one-dimensional Inception model was used for Supernovae classification using the light flux of a region in space as an input MTS for the network~\cite{brunel2019cnn}. However, the authors limited the conception of their Inception architecture to the first version of this model~\cite{szegedy2015going}. The Inception-ResNet~\cite{ronald2021isplinception} architecture includes convolutional layers, followed by Inception modules and residual blocks. The Inception modules are used to learn multiple scales and aspects of the data, allowing the network to capture more complex patterns. The residual blocks are then used to learn the residuals, or differences, between the input and output of the network, improving its performance.

InceptionTime~\cite{fawaz2020inceptiontime} explores much larger filters than any previously proposed network for TSC to reach state-of-the-art performance on the UCR benchmark. InceptionTime is an ensemble of five randomly initialized inception network models, each of which consists of two blocks of inception modules. 
Each inception module first reduces the dimensionality of a multivariate time series using a bottleneck layer with a length and stride of 1 while maintaining the same length. 
Then, 1D convolutions of different lengths are applied to the output of the bottleneck layer to extract patterns at different sizes.
In parallel, a max pooling layer followed by a bottleneck layer are also applied to the original time series to increase the robustness of the model to small perturbations.
The output from the convolution and max pooling layers are stacked to form a new multivariate time series which is then passed to the next layer.
Residual connections are used between each inception block to reduce the vanishing gradient effect. 
The output of the second inception block is passed to a GAP layer before feeding into a softmax classifier.

The strong performance of InceptionTime has inspired a number of extensions.
Like InceptionTime, EEG-inception~\cite{sun2021prototypical} uses several inception layers and residual connections as its backbone. Additionally, noise addition-based data augmentation of EEG signals is proposed, which increases the average accuracy. InceptionFCN~\cite{usmankhujaev2021time} focuses on combining two well-known deep learning techniques, namely the Inception module and the Fully Convolutional Network~\cite{usmankhujaev2021time}. In KDCTime~\cite{gong2022kdctime}, label smoothing (LSTime) and knowledge distillation (KDTime) were introduced for InceptionTime, automatically generated while compressing the inference model. Additionally, knowledge distillation with calibration (KDC) in KDCTime offers two calibrating strategies: KDC by translating (KDCT) and KDC by reordering (KDCR). LITE~\cite{ismail2023lite} addresses InceptionTime's complexity while preserving its TSC performance. Utilizing DepthWise Separable Convolutions, LITE incorporates multiplexing, dilated convolution, and custom filters~\cite{ismail2022deep} to enhance efficiency.

\subsection{Recurrent Neural Network}
Recurrent Neural Networks are types of neural networks built with internal memory to work with time series and sequential data. Conceptually similar to feed-forward neural networks (FFNs), RNNs differ in their ability to handle variable-length inputs and produce variable-length outputs.
\vspace{-0.25cm}
\subsubsection{Vanilla Recurrent Neural Networks (Vanilla RNNs)}
Recurrent neural networks for TSC have been proposed in~\cite{husken2003recurrent}. Using RNNs, the input series have been classified based on their dynamic behavior. They used sequence-to-sequence architecture in which each sub-series of input series is classified in the first step. Then the argmax function is applied to the entire output, and finally, the neuron with the highest rate specifies the classification result. In order to improve the model parallelization and capacity~\cite{dennis2019shallow} proposed a two-layer RNN. In the first layer, the input sequence is split into several independent RNNs to improve parallelization, followed by a second layer that utilizes the first layer's output to capture long-term dependencies~\cite{dennis2019shallow}. Further, RNNs have been used in some hierarchical architectures~\cite{fernandez2007sequence,hermans2013training}. Hermans and Schrauwen showed a deeper version of recurrent neural networks could perform hierarchical processing on complex temporal tasks and capture the time series structure more naturally than a shallow version~\cite{hermans2013training}. RNNs are usually trained iteratively using a procedure known as backpropagation through time (BPTT). When unfolded in time, RNNs look like very deep networks with shared parameters. With deeper neural layers in RNN and sharing weights across different RNN cells, the gradients are summed up at each time step to train the model. Thus, gradients undergo continuous matrix multiplication due to the chain rule and either shrink exponentially and have small values called vanishing gradients or blow up to a very large value, referred to as exploding gradients~\cite{pascanu2013difficulty}.  
These problems motivated the development of second-order methods for deep architectures named long short-term memory (LSTM)~\cite{hochreiter1997long} and Gated Recurrent Unit (GRU)~\cite{chung2014empirical}.
\vspace{-0.25cm}
\subsubsection{Long Short Term Memory (LSTM)}  
LSTM addresses the common vanishing/exploding gradient issue in vanilla RNNs by integrating memory cells with gate control into their state dynamics~\cite{hochreiter1997long}.
Due to its design nature, LSTM is suited to problems involving sequence data, such as language translation~\cite{sutskever2014sequence}, video representation learning~\cite{donahue2015long}, and image caption generation~\cite{karpathy2015deep}. The TSC problem is not an exception and mainly adopts a similar model to the language translation~\cite{sutskever2014sequence}. Sequence-to-Sequence with Attention (S2SwA)~\cite{tang2016sequence} incorporates two LSTMs, one encoder and one decoder, in a sequence-to-sequence fashion for TSC. In this model, the encoder LSTM accepts input time series of arbitrary lengths and extracts information from the raw data based on which the decoder LSTM constructs fixed-length sequences that can be regarded as automatically extracted features for classification.
\vspace{-0.25cm}
\subsubsection{Gated Recurrent Unit (GRU)}
GRU, another widely-used variant of RNNs, shares similarities with LSTM in its ability to control information flow and memorize context across multiple time steps~\cite{chung2014empirical}. Similar to S2SwA~\cite{tang2016sequence} sequence auto-encoder (SAE) based on GRU has been defined to deal with TSC problem~\cite{malhotra2017timenet}. A fixed-size output is produced by processing the various input lengths using GRU as the encoder and decoder. The model's accuracy was also improved by pre-training the parameters on massive unlabeled data.
\vspace{-0.25cm}
\subsubsection{Hybrid Models}
CNN's and RNNs are often combined for TSC because they have complementary strengths. As mentioned previously, CNNs are well-suited for learning from spatial relationships in data, such as the patterns and correlations between the channels of different time steps in a time series. This allows them to learn useful features from the time series data that can help improve the classification performance. RNNs, on the other hand, are well-suited for learning from temporal dependencies in data, such as the past values of a time series that can help predict its future values. This allows them to capture the dynamic nature of time series data and make more accurate predictions. Combining the strengths of CNNs and RNNs makes it possible to learn spatial and temporal features from the time series data, improving the model's performance for TSC. Additionally, the two models can be trained together, allowing them to learn from each other and improve the model's overall performance.

Various extensions like MLSTM-FCN~\cite{karim2019multivariate}, TapNet~\cite{zhang2020tapnet}, and SMATE~\cite{zuo2021smate}  were proposed later to deal with time-series data. MLSTM-FCN extends the univariate LSTM-FCN model~\cite{karim2017lstm} to the multivariate case. Like the LSTM-FCN, the multivariate version comprises LSTM blocks and fully convolutional blocks for extracting features from input series.
A squeeze and excite block is also added to the FCN block, and can execute a form of self-attention on the output feature maps of previous layers~\cite{karim2019multivariate}. Two further proposals for multivariate TSC are the Time series attentional prototype Network (TapNet) and Semi-Supervised Spatio-Temporal (SMATE) ~\cite{zhang2020tapnet,zuo2021smate}. These methods combine and seek to leverage the relative strengths of both traditional distance-based and deep-learning approaches.

MLSTM-FCN, TapNet, and SMATE were designed in dual-network architectures. The input is separately fed into the CNN and RNN models, and their output is concentrated before the fully connected layer for the final task. However, one branch cannot fully use the hidden states of the other during feature extraction since the final classification results are generated by concatenating the outputs of the two branches. That motivates different types of architecture that try layer-wise integration of CNN and RNN models. This motivates different architectures, such as GCRNN~\cite{lin2017gcrnn} and CNN-LSTM~\cite{mutegeki2020cnn}, which aim to integrate CNNs and RNNs in a layer-wise fashion.

While recurrent neural networks are commonly used for time series forecasting, only a few studies have applied them to TSC, mainly due to four reasons: (1) RNNs typically struggle with the gradient vanishing and exploding problem due to training on long-time series~\cite{pascanu2012understanding}. (2) RNNs are considered difficult to train and parallelize, so researchers are less likely to use them as they are computationally expensive~\cite{pascanu2013difficulty}. (3) Recurrent architectures are designed mainly to learn from the previous data to make predictions about the future~\cite{langkvist2014review}. (4) RNN models can fail to effectively capture and utilize long-range dependencies in long sequences~\cite{tang2016sequence}.

\vspace{-0.4cm}
\subsection{Attention based model} 
Despite the excellent performance of CNN models for capturing local temporal/spatial correlations, these models can not effectively capture and utilize long-range dependencies. Additionally, they only consider the local order of data points rather than the overall order of all data points. Therefore, many recent studies have embedded recurrent neural networks (RNN) such as LSTMs alongside the CNNs to capture this information~\cite{karim2017lstm,karim2019multivariate,zhang2020tapnet}. The disadvantage of RNN-based models is that they are computationally expensive, and their capability to capture long-range dependencies is limited~\cite{vaswani2017attention,hao2020new}. 
On the other hand, \textit{attention models} can capture long-range dependencies, and their broader receptive fields provide more contextual information, which can improve the models' learning capacity. 
The attention mechanism aims to enhance a network's representation ability by focusing on essential features and suppressing unnecessary ones. Not surprisingly, with the success of \textit{attention models} in natural language processing~\cite{vaswani2017attention,devlin2018bert}, many previous studies have attempted to bring the power of attention models into various domains such as computer vision~\cite{dosovitskiy2020image} and time series analysis~\cite{hao2020new,li2019enhancing,zhou2021informer,zerveas2021transformer,kostas2021bendr}. Table~\ref{tab:att} presents a list of the attention-based models reviewed in this paper.

\vspace{-0.25cm}
\subsubsection{Self-Attention}
Self-attention has been demonstrated to be effective in various natural language processing tasks due to its 
ability to capture long-term dependencies in text~\cite{vaswani2017attention}. 
Recently, it has also been shown to be effective for TSC tasks~\cite{hao2020new,yuan2018muvan,hsieh2021explainable,chen2021multi}. 
As we mentioned, the self-attention module is embedded in the encoder-decoder models to improve the model performance. However, only the encoder and the self-attention module have been used for TSC. 
Early models of TSC follow the same backbone of natural language processing models and use the Recurrent-based models such as RNN~\cite{yuan2018novel}, GRU\cite{yuan2018muvan} and LSTM\cite{liang2018geoman,hu2020multistage} for encoding the input series. For example, the Multi-View Attention Network (MuVAN) applies bidirectional GRUs independently to each input dimension as the encoder and then feeds all the representations into a self-attention bock~\cite{yuan2018muvan}.

As a result of the excellent performance of the CNN models, many studies have attempted to encode the time series using CNN before applying attention~\cite{hao2020new,hsieh2021explainable,cheng2020novel,xiao2021rtfn}. 
Cross Attention Stabilized Fully Convolutional Neural Network (CA-SFCN)~\cite{hao2020new} and Locality Aware eXplainable Convolutional ATtention network (LAXCAT)~\cite{hsieh2021explainable} applied the self-attention mechanism to leverage the long-term dependencies for the MTSC task. CA-SFCN combines FCN and two types of self-attention - temporal attention (TA) and variable attention (VA), which interact to capture the long-range dependencies and variables interactions. LAXCAT also used temporal and variable attention to identify informative variables and the time intervals where they have informative patterns for classification. WaveletDTW Hybrid attEntion Networks (WHEN)~\cite{wang2023wavelet} integrate two attention mechanisms, namely wavelet attention and DTW attention, into the BiLSTM to enhance model performance. In wavelet attention, they leverage wavelets to compute attention scores, specifically targeting the analysis of dynamic frequency components in nonstationary time series. Simultaneously, DTW attention employs the DTW distance to calculate attention scores, addressing the challenge of time distortion in multiple time series.

Several self-attention models have been developed to improve network performance~\cite{jaderberg2015spatial,woo2018cbam}, including Squeeze-and-Excitation (SE)~\cite{hu2018squeeze}, which focuses on channel attention and is often used to classify time series data~\cite{karim2019multivariate,chen2021multi,wang2021time}. 
The SE block allows the whole network to use global information to selectively focus on the informative feature maps and suppress less important ones~\cite{hu2018squeeze}.
More importantly, the SE block can increase the quality of the shared lower-level representations in the early layers and becomes increasingly specialized when responding to different inputs in later layers. The weight of each feature map is automatically learned at each layer of the network, and the SE block can boost feature discrimination throughout the whole network. Multi-scale Attention Convolutional Neural Network (MACNN)~\cite{chen2021multi} applies the different kernel size convolutions to capture different scales of information along the time axis by generating feature maps at differing scales. Then an SE block is used to enhance useful feature maps and suppress less useful ones by automatically learning each feature map's importance. 

\vspace{-0.25cm}
\subsubsection{Transformers}

The impressive performance of multi-headed attention has led to numerous attempts to adapt multi-headed attention to the TSC domain. Transformers for classification usually employ a simple encoder structure consisting of attention and feed-forward layers. SAnD (Simply Attend and Diagnose)~\cite{song2018attend} architecture adopted a multi-head attention mechanism similar to a vanilla transformer~\cite{vaswani2017attention} to classify clinical time series for the first time. The model uses both positional encoding and a dense interpolation embedding technique to incorporate temporal order into representation learning. In another study that classified vibration signals~\cite{jin2021end}, time-frequency features such as Frequency Coefficients and Short Time Fourier Transformation (STFT) spectrums are used as input embeddings to the transformers. A multi-head attention-based model was applied to raw optical satellite TSC using Gaussian Process Interpolation~\cite{Rasmussen2004} embedding and outperformed convolution, and recurrent neural networks~\cite{allam2021paying}.

Gated Transformer Networks (GTN)~\cite{liu2021gated} use two-tower multi-headed attention to capture the discriminative information from the input series. Also, they merged the output of two towers using a learnable matrix named gating. To enhance locality awareness of transformers for TSC, flexible multi-head linear attention (FMLA)~\cite{zhao2022rethinking} integrates deformable convolutional blocks and online knowledge distillation, as well as a random mask to reduce noise. For each TSC dataset, AutoTransformer searches for the suitable network architecture using the neural architecture search (NAS) algorithm before feeding the output to the multi-headed attention blocks. 
ConvTran~\cite{Foumani2023} currently stands as the state of the art in multivariate TSC. They conducted a review of existing absolute and relative position encoding methods in TSC.  Based on the limitations of the current position encodings for time series, they introduced two novel ones named tAPE and eRPE for absolute and relative positions, respectively. Integrating these proposed position encodings into a transformer block and combining them with a convolution layer, they presented a novel deep-learning framework for multivariate time series classification—ConvTran.


\begin{table}
\setlength{\tabcolsep}{12pt}
\begin{tabular}{lccc}
\hline
\multicolumn{1}{c}{\textbf{Model}} &
  \textbf{Year} &
  \multicolumn{1}{c}{\textbf{Embedding}} &
  \multicolumn{1}{c}{\textbf{Attention}} \\ \hline
MuVAN~\cite{yuan2018muvan}              & 2018 & Bi-GRU              & Self-attention         \\ 
ChannelAtt~\cite{yuan2018novel}         & 2018 & RNN                 & Self-attention         \\ 
GeoMAN~\cite{liang2018geoman}           & 2018 & LSTM                & Self-attention         \\ 
Multi-Stage-Att~\cite{hu2020multistage} & 2020 & LSTM                & Self-attention         \\ 
CT\_CAM~\cite{cheng2020novel}            & 2020 & FCN + Bi-GRU        & Self-attention         \\ 
CA-SFCN~\cite{hao2020new}               & 2020 & FCN                 & Self-attention         \\ 
RTFN~\cite{xiao2021rtfn}               & 2021 & CNN + LSTM                 & Self-attention         \\ 
LAXCAT~\cite{hsieh2021explainable}      & 2021 & CNN                 & Self-attention         \\ 
MACNN~\cite{chen2021multi}              & 2021 & Multi-scale CNN     & Squeeze-and-Excitation \\ 
WHEN~\cite{wang2023wavelet}              & 2023 & CNN + BiLSTM     & Self-attention \\ 

\hline 
SAnD~\cite{song2018attend} &            2018 &  \begin{tabular}[c]{@{}l@{}} Linear Embedding \end{tabular} & Multi-Head \\ 
T2~\cite{allam2021paying}               & 2021 & \begin{tabular}[c]{@{}l@{}} Gaussian Process Regression + 1D Conv\end{tabular} & Multi-Head             \\ 
GTN~\cite{liu2021gated}                 & 2021 & Linear Embedding & Multi-Head             \\ 
TRANS\_tf~\cite{jin2021end}             & 2021   &  time-frequency features & Multi-Head \\ 
FMLA~\cite{zhao2022rethinking}          & 2022 & Deformable CNN      & Multi-Head             \\ 
AutoTransformer~\cite{ren2022autotransformer} & 2022 & Multi-scale CNN + NAS &  Multi-Head \\

ConvTran~\cite{Foumani2023} & 2023 & Disjoint-CNN &  Multi-Head \\
\hline
\end{tabular}
\caption{Summary of Attention-based Models for Time Series Classification and Extrinsic Regression}
\label{tab:att}
\vspace{-1 cm}
\end{table}

\vspace{-0.3cm}
\subsection{Graph Neural Networks} 

While both CNNs and RNNs perform well on Euclidean data, many time series problems have data that are more naturally represented as graphs~\cite{Jin2023graph}. For example, in a network of sensors, the sensors may be irregularly spaced, instead of the sensors forming a regular grid. A graph representation of data collected by this network can model this irregular layout more accurately than can be done using a Euclidean space. However, using standard deep learning algorithms to learn from graph structures is challenging~\cite{Wu2021graph}. For example, nodes may have a varying number of neighbouring nodes, making it difficult to apply a convolution operation.

Graph Neural Networks (GNNs)~\cite{Scarselli2009graph} are methods that adapt deep learning techniques to the graph domain. Much of the early research using GNNs for time series analysis concentrated on forecasting tasks~\cite{Jin2023graph}. However, recent works consider GNNs for TSC~\cite{Xi2023graph,Liu2023graph} and TSER~\cite{Bloemheuvel2023graph} tasks. A list of the GNN models reviewed in this paper is provided in table \ref{table:gnns}. Time2Graph+~\cite{Cheng2021graph} transforms each time series into a shapelet graph. Shapelets are extracted from the time series and form the graph nodes. The graph edges are weighted based on transition probabilities between the two shapelets. Once the input graphs have been constructed, a graph attention network is used to create a representation of the time series that is fed into a classifier.

\begin{table}
\begin{tabulary}{0.96\textwidth \tymin=2cm}{p{2.7cm} L L p{4.3cm}}
\hline
\multicolumn{1}{c}{\textbf{Model}} &
  \textbf{Year} &
  \multicolumn{1}{c}{\textbf{GNN Type}} &
  \multicolumn{1}{c}{\textbf{Other Components}} \\ \hline
\noalign{\smallskip}
TGCN~\cite{covert2019temporal} & 2019 & Graph convolutional network & 1D-CNN \\
\noalign{\smallskip}
DGCNN~\cite{Song2020graph} & 2020 & Graph convolutional network & 1x1 CNN \\
\noalign{\smallskip}
GraphSleepNet~\cite{jia2020graphsleepnet} & 2020 & Graph convolutional network & Temporal attention \\
\noalign{\smallskip}
T\mbox{-}GCN~\cite{ma2021deep} & 2021 & Graph convolutional network & GRU \\
\noalign{\smallskip}
MRF-GCN~\cite{li2020multireceptive} & 2021 & Graph convolutional network & Fast Fourier Transforms (FFT) \\
\noalign{\smallskip}
Nhu et al.~\cite{nhu2021graph} & 2021 & Graph convolutional network & 1D-CNN \\
\noalign{\smallskip}
DCRNN~\cite{Tang2021graph} & 2021 & Graph convolutional network & GRU \\
\noalign{\smallskip}
Time2Graph+~\cite{Cheng2021graph} & 2021 & Graph attention & Shapelet transform \\
\noalign{\smallskip}
RAINDROP~\cite{Zhang2021graph} & 2021 & Graph guided network & Temporal attention \\
\noalign{\smallskip}
STEGON~\cite{Censi2021graph} & 2021 & Graph attention & 1D-CNN \\
\noalign{\smallskip}
Azevedo et al.~\cite{Azevedo2022graph} & 2022 & Graph network block with pooling & 1D-CNN \\
\noalign{\smallskip}
MTPool~\cite{Duan2022graph} & 2022 & Variational Graph Pooling & 1D-CNN \\
\noalign{\smallskip}
SimTSC~\cite{Zha2022graph} & 2022 & Graph convolutional network & DTW, ResNet \\
\noalign{\smallskip}
Tulczyjew et al.~\cite{Tulczyjew2022graph} & 2022 & Graph convolutional network & Adaptive pooling \\
\noalign{\smallskip}
C-DGAM~\cite{Sun2023graph} & 2023 & Graph attention & 1D-CNN with attention \\
\noalign{\smallskip}
Dufourg et al.~\cite{Dufourg2023graph} & 2023 & Spatio-temporal graph & Simple Linear Iterative Clustering \\
\noalign{\smallskip}
TISER-GCN~\cite{Bloemheuvel2023graph} & 2023 & Graph convolutional network & 1D-CNN \\
\noalign{\smallskip}
TodyNet~\cite{Liu2023graph} & 2023 & Dynamic graph neural network & 1D-CNN \\
\noalign{\smallskip}
LB-SimTSC~\cite{Xi2023graph} & 2023 & Graph convolutional network & Lower-bound DTW, ResNet \\
\noalign{\smallskip}
\hline
\noalign{\smallskip}
\end{tabulary}
\caption{Summary of graph neural network models for time series classification and extrinsic regression}
\label{table:gnns}
\vspace{-0.9 cm}
\end{table}
SimTSC~\cite{Zha2022graph} constructs a pairwise similarity graph where each time series forms a node and edge weights are computed based on the DTW distance measure. Node attributes are generated using a feature vector encoder. GNN operations are used to enhance the node features based on similarities between adjacent time series. These representations are then used for the final classification step, which produces a classification for each node. LB-SimTSC~\cite{Xi2023graph} replaces the expensive DTW computation with the LB-Keogh lower-bounding method~\cite{Keogh2005exact}.

Spatiotemporal GNNs model both spatial (or inter-variable) and temporal dependencies using two modules that work in tandem. The spatial module models the dependencies between the time series by applying graph convolutions over a GNN (graph convolutional networks or GCNs~\cite{Kipf2016graph}). The temporal module models the dependencies within the time series using an RNN~\cite{ma2021deep,Tang2021graph}, 1D-CNN~\cite{Censi2021graph,Azevedo2022graph}, Attention~\cite{Zhang2021graph,Sun2023graph}, or a combination of these~\cite{Jin2023graph}. The features extracted from the graph layers are then fed into the classification or regression layers to make either a single prediction~\cite{Tang2021graph,Azevedo2022graph,Zhang2021graph,Sun2023graph} or a prediction for each node~\cite{ma2021deep,Censi2021graph}.

Spatiotemporal GCNs are often used to analyse sensor arrays, where the graph structure models the physical layout of the sensors. A common example is electroencephalogram (EEG) data, where the location of EEG electrodes is represented as a graph that is used to analyse the EEG signal. Some of these applications are epilepsy detection~\cite{nhu2021graph}, seizure detection~\cite{covert2019temporal,Tang2021graph}, emotion recognition~\cite{Song2020graph}, and sleep classification~\cite{jia2020graphsleepnet}. Besides EEG, GCNs have also been applied to engineering applications such as machine fault diagnosis~\cite{li2020multireceptive}, slope deformation prediction~\cite{ma2021deep} and seismic activity prediction~\cite{Bloemheuvel2023graph}.
MTPool~\cite{Duan2022graph} uses a spatiotemporal GCN for multivariate time series classification. In this study, each channel in the time series is represented by a node in the graph and the graph edges model the correlations between the channels. The GCN is combined with temporal convolutions and a hierarchical graph pooling technique.
Spatiotemporal GNNs have also been used for object-based image analysis~\cite{Censi2021graph} and semantic segmentation~\cite{Tulczyjew2022graph} of image time series. However, these assume the labels and spatial relationships are static over time. In many cases these may both change. Spatiotemporal \emph{graphs} (STGs), which include temporal edges as well as spatial edges, can model these dynamic relationships~\cite{Dufourg2023graph}. In STGs, each node represents an object at one timestamp. Spatial edges connect the object to adjacent objects and temporal edges connect two objects in consecutive images if they have common pixels.




\section{Self-supervised Models} \label{sec:ss}
Obtaining labeled data for large time series datasets poses significant costs and challenges. Machine learning models trained on large labeled time series datasets often exhibit superior performance compared to models trained on sparsely labeled datasets, small datasets with limited labels, or those without supervision, leading to suboptimal performance across various time series machine learning tasks \cite{yue2022ts2vec,yang2022unsupervised}. As a result, rather than depending on high-quality annotations for large datasets, researchers and practitioners are increasingly shifting their focus toward self-supervised representation learning for time series.

Self-supervised representation learning, a subfield of machine learning, focuses on learning representations from data without explicit supervision \cite{foumani2023series2vec}. In contrast to supervised learning, which relies on labeled data, self-supervised learning methods utilize the inherent structure of the data to learn valuable representations in an unsupervised manner. The learned representations can then be used for a variety of downstream tasks including classification, anomaly detection, and forecasting. This survey specifically emphasizes classification as a downstream task. We categorized self-supervised learning approaches for TSC into three groups based on the pretext. Table~\ref{tab:SS} shows a list of the self-supervised models reviewed in this paper.
\vspace{-0.4cm}
\subsection{Contrastive Learning}
Contrastive learning involves model learning to differentiate between positive and negative time series examples. Time-Contrastive Learning (TCL)~\cite{hyvarinen2016unsupervised}, Scalable Representation Learning (SRL or T-Loss)~\cite{franceschi2019unsupervised} and Temporal Neighborhood Coding (TNC) \cite{tonekaboni2021unsupervised} apply a subsequence-based sampling and assume that distant segments are negative pairs and neighbor segments are positive pairs. TNC takes advantage of the local smoothness of a signal’s generative process to define neighborhoods in time with stationary properties to further improve the sampling quality for the contrastive loss function. TS2Vec \cite{yue2022ts2vec} uses contrastive learning to obtain robust contextual representations for each timestamp hierarchically. It involves randomly sampling two overlapping subseries from input and encouraging consistency of contextual representations on the common segment. The encoder is optimized using both temporal contrastive loss and instance-wise contrastive loss. 

In addition to the subsequence-based methods, other models employ instance-based sampling~\cite{eldele2021time,wickstrom2022mixing,yang2022timeclr,yang2022unsupervised,zhang2022self,meng2023mhccl}, treating each sample individually to generate positive and negative samples for contrastive loss. 
Time-series Temporal and Contextual Contrasting (TS-TCC)~\cite{eldele2021time} uses weak and strong augmentations to transform the input series into two views and then uses a temporal contrasting module to learn robust temporal representations. The contrasting contextual module is then built upon the contexts from the temporal contrasting module and aims to maximize similarity among contexts of the same sample while minimizing similarity among contexts of different samples. Similarly, TimeCLR~\cite{yang2022timeclr} introduces DTW data augmentation to enhance robustness against phase shift and amplitude change phenomena.
Bilinear Temporal-Spectral Fusion (BTSF)~\cite{yang2022unsupervised} uses simple dropout as the augmentation method and aims to incorporate spectral information into the feature representation. Similarly, Time-Frequency Consistency (TF-C) \cite{zhang2022self} is a self-supervised learning method that leverages the frequency domain to achieve better representation. It proposes that the time-based and frequency-based representations, learned from the same time series sample, should be more similar to each other in the time-frequency space compared to representations of different time series samples.

\begin{table}[]
\setlength{\tabcolsep}{9pt}
\begin{tabular}{llcl}
\hline
Model & \multicolumn{1}{c}{Year} & \textbf{Encoder Backbones} &  \\ \hline

\textbf{Contrastive Learning} &  &  & \textbf{Other features}\\ \hline
TCL~\cite{hyvarinen2016unsupervised}& 2016 & MLP &Sequence-based contrast \\
T-Loss/SRL~\cite{franceschi2019unsupervised}& 2019 & Causal CNN & Sequence-based contrast  \\
TNC~\cite{tonekaboni2021unsupervised}& 2021 & Bidirectional RNN &Sequence-based contrast   \\
TS-TCC~\cite{eldele2021time} & 2021& CNN + Transformers&Instance/Sequence-based contrast \\
MCL~\cite{wickstrom2022mixing} & 2021 &FCN & Instance-based contrast\\
TimeCLR~\cite{yang2022timeclr} & 2021 & InceptionTime & Instance-based contrast \\
TS2Vec~\cite{yue2022ts2vec} & 2021& Dilated CNN &Sequence-based contrast  \\
BTSF~\cite{yang2022unsupervised}& 2022& Causal CNN& Instance-based contrast  \\
TF-C~\cite{zhang2022self} & 2022 & ResNets &Instance-based contrast \\
MHCCL~\cite{meng2023mhccl} & 2023 & ResNet &Instance-based contrast \\
\hline
\multicolumn{4}{l}{\textbf{Self-Prediction}}  \\ \hline
BENDR~\cite{kostas2021bendr} & 2021 & CNN + Transformers &  Sequence masking\\
Voice2Series~\cite{yang2021voice2series} & 2021 & CNN+Transformers & Binary masking \\
TST~\cite{zerveas2021transformer}& 2021 & Transformers & Binary masking \\
TARNet~\cite{chowdhury2022tarnet}& 2022 & Transformers & Binary masking \\
TimeMAE~\cite{cheng2023timemae}& 2023 & CNN + Transformers & Sequence masking  \\
CRT~\cite{zhang2023self}& 2023 & Transformers & Sequence masking \\
  \hline

\multicolumn{4}{l}{\textbf{Other Pretext tasks}} \\ \hline
PHIT~\cite{ismail2023finding}& 2023& H-InceptionTime & \\
Series2Vec~\cite{foumani2023series2vec}& 2023 & Disjoint CNN & Similarity based representation learning  \\
 \hline

\end{tabular}%
\caption{Summary of self-supervised models for time series classification and extrinsic regression}
\vspace{-0.9cm}
\label{tab:SS}
\end{table}

\vspace{-0.4cm}
\subsection{Self-Prediction}
The primary objective of self-prediction-based self-supervised models is to reconstruct the input or representation of input data. Studies have explored using transformer-based self-supervised learning methods for TSC~\cite{kostas2021bendr,yang2021voice2series,zerveas2021transformer,chowdhury2022tarnet,cheng2023timemae,zhang2023self}, following the success of models like BERT~\cite{devlin2018bert}. BErt-inspired Neural Data Representations (BENDER)\cite{kostas2021bendr} uses the transformer structure to model EEG sequences and shows that it can effectively handle massive amounts of EEG data recorded with differing hardware. Another study, Voice-to-Series with Transformer-based Attention (V2Sa)\cite{yang2021voice2series}, utilizes a large-scale pre-trained speech processing model for TSC.

Transformer-based Framework (TST)\cite{zerveas2021transformer} and TARNet~\cite{chowdhury2022tarnet} adapts vanilla transformers to the multivariate time series domain and uses a self-prediction-based self-supervised pre-training approach with masked data. These studies demonstrate the potential of using transformer-based self-supervised learning methods for TSC. 

\vspace{-0.3cm}
\subsection{Other Pretext tasks}

While many pretext tasks in self-supervised learning are typically contrastive or self-predictive, specific tasks are tailored for time series data. In image-based self-supervised learning, synthetic transformations (augmentation) of an image are created, and the model learns to contrast the image and its transforms with other images in the training data, which works well for object interpretation. However, time series analysis fundamentally differs from vision or natural language processing concerning the definition of meaningful self-supervised learning tasks. 

Guided by this insight, Foumani et al.~\cite{foumani2023series2vec} introduce Series2Vec, a novel self-supervised representation learning approach. Unlike other contrastive self-supervised methods in time series, which carry the risk of positive sample variants being less similar to the anchor sample than series in the negative set, Series2Vec is trained to predict the similarity between two series in both temporal and spectral domains through a self-supervised task. Series2Vec relies primarily on the consistency of the unsupervised similarity step, rather than the intrinsic quality of the similarity measurement, without the need for hand-crafted data augmentation. Pre-trained H-InceptionTime (PHIT)~\cite{ismail2023finding} is pre-trained using a novel pretext task designed to identify the originating dataset of each time series sample. The objective is to generate flexible convolution filters that can be applied across diverse datasets. Furthermore, PHIT demonstrates its capability to mitigate overfitting in small datasets.

\vspace{-0.3cm}
\section{Data augmentation} \label{sec:aug}

In the field of deep learning, the concept of data augmentation has emerged as an important tool for enhancing performance, particularly in scenarios where the availability of training data is limited~\cite{shorten2019survey}. Originally proposed in computer vision, data augmentation involves a variety of transformations to images, such as cropping, rotating, flipping, and applying filters like blurring and sharpening. These transformations serve to introduce a diverse range of scenarios within the training data, thereby aiding in the development of more robust and generalizable models.
However, the direct application of these image-based augmentation techniques to time series data often proves to be inadequate or inappropriate. Operations like rotation may disrupt the intrinsic temporal structure of time series data.

The challenge of overfitting is particularly pronounced in the field of deep learning models for TSC. These models are characterized by a high number of trainable parameters, which can lead to a model that performs well on training data but fails to generalize to unseen data. In such cases, data augmentation can be a valuable strategy. It offers an alternative to the costly and sometimes impractical approach of collecting additional real-world data. By generating synthetic samples from existing datasets, we can effectively augment the size and variety of our training data.
The following details different investigated methods to produce synthetic time series for data augmentation.

\vspace{-0.25cm}    
\subsubsection*{Random Transformations}
Several augmentations have been developed for the magnitude domain. Jittering, as explored by Um et al.~\cite{um2017data}, involves the addition of random noise to the time series. Another method, flipping~\cite{rashid2019window}, reverses the time series values.  Scaling is a technique where the time series is multiplied by a factor from a Gaussian distribution. Magnitude warping, which shares similarities with scaling, distorts the series along a curve that varies smoothly.
For time domain transformations, permutation algorithms play a significant role. For example, the slicing transformation involves removing sub-sequence from the series. There are also various warping methods like Random Warping~\cite{iwana2021time}, Time Warping~\cite{um2017data}, Time Stretching~\cite{nguyen2020improving}, and Time Perturbation~\cite{vachhani2018data}, each introducing different forms of distortion to the time series.
Finally, in the frequency domain, transformations often utilize the Fourier transform. For example, Gao et al.~\cite{gao2020robusttad} introduce perturbations to both the magnitude and phase spectrum following a Fourier transform.
\vspace{-0.25cm}    
\subsubsection*{Window methods}
A primary approach in window methods is to create new time series by combining segments from various series of the same class. This technique effectively enriches the data pool with a variety of samples.
Window slicing, as introduced by Cui et al.~\cite{cui2016multiscale} involves dividing a time series into smaller segments, with each segment retaining the class label of the original series. These segments are then used to train classifiers, offering a detailed view of the data. During classification, each segment is evaluated individually, and a collective decision on the final label is reached through a voting system among the slices.
Another technique is window warping, based on the DTW algorithm. This method adjusts segments of a time series along the temporal axis, either stretching or compressing them. This introduces variability in the time dimension of the data. Le Guennec et al.~\cite{leguennec2016data} work provides examples of the application of both window slicing and window warping, showcasing their effectiveness in enhancing the diversity and representativeness of time series datasets.

\vspace{-0.25cm}    
\subsubsection*{Averaging methods}
Averaging methods in time series data augmentation combine multiple series to form a new, unified series. This process is more difficult than it might seem, as it requires careful consideration of factors like noise and distortions in both the time and magnitude aspects of the data.
In this context, weighted Dynamic Time Warping (DTW) Barycenter Averaging (wDBA) introduced by Forestier et al.~\cite{forestier2017generating} provides an averaging method by aligning time series in a way that accounts for their temporal dynamics. The practical application of wDBA is illustrated in the study by Ismail Fawaz et al.~\cite{fawaz2018data}, where it is employed in conjunction with a ResNet classifier, demonstrating its effectiveness.
Additionally, the research conducted by Terefe et al.~\cite{terefe2020time} uses an auto-encoder for averaging a set of time series. This method represents a more advanced approach in time series data augmentation, exploiting the auto-encoder's capacity for learning and reconstructing data to generate averaged representations of time series.

\vspace{-0.25cm}    
\subsubsection*{Selection of data augmentation methods}
The selection of the appropriate data augmentation technique is critical and must be adapted to the specific characteristics of the dataset and the architecture of the neural network being used. Studies like those conducted by Iwana et al.~\cite{iwana2021empirical}, Pialla et al.~\cite{pialla2022data} and Gao et al~\cite{gao2023data} highlight the complexity of this task. These studies demonstrate that the effectiveness of augmentation techniques can vary significantly across different datasets and neural network architectures. Consequently, a method that proves effective in one scenario may not necessarily yield similar results in another. To this end, practitioners in the field of TSC must engage in a careful and informed process of method selection and tuning. While the array of available data augmentation techniques offers a comprehensive toolkit for tackling the challenges of limited data and overfitting, their successful application depends heavily on a nuanced understanding of both the methods themselves and the specific demands of the task at hand.

\vspace{-0.25cm}    
\section{Transfer learning} \label{sec:TL}

Transfer learning, initially popularized in the field of computer vision, is increasingly becoming relevant in the domain of TSC. In computer vision, this approach involves using a pre-trained network, typically on large datasets like ImageNet~\cite{deng2009imagenet}, as a starting point rather than initiating with random network weights. This method is also related to the concept of foundation or base models, which are large-scale machine learning models trained on extensive data, often using self-supervised or semi-supervised learning. These models are adaptable to a wide array of tasks, showcasing their versatility.
The principle of transfer learning is also closely associated with domain adaptation which focuses on applying a model trained on a source data distribution to a different, but related, target data distribution. This approach is crucial in leveraging pre-trained models for various applications, particularly in scenarios where data is scarce or specific to certain domains.

In the context of TSC, insights have been contributed by the work of Ismail Fawaz et al.~\cite{fawaz2018transfer}, who conducted a study using the UCR archive.  Their extensive experiments demonstrated that transfer learning could lead to positive or negative outcomes, depending on the chosen datasets for transfer. This finding underscores the importance of the relationship between source and target datasets in transfer learning efficacy. Ismail Fawaz et al.~\cite{fawaz2018transfer} also introduced an approach to predict the success of transfer learning in TSC by using DTW to measure similarities between datasets. This metric serves as a guide to select the most appropriate source dataset for a given target dataset, thereby enhancing accuracy in a majority of cases.

Other researchers have also explored transfer learning in TSC. Spiegel~\cite{spiegel2016transfer} work on using dissimilarity spaces to enrich feature representations in TSC set a precedent for employing unconventional data sources. This approach of enhancing learning with diverse data types finds a parallel in Li et al.~\cite{li2020deep} method, which leverages sensor modality labels from various fields to train a deep network, emphasizing the importance of versatile data in transfer learning.
Building on the concept of data diversity, Rotem et al.~\cite{rotem2022transfer} pushed the boundaries further by generating a synthetic univariate time series dataset for transfer learning. This synthetic dataset, used for regression tasks, underscores the potential of artificial data in overcoming the limitations of real-world datasets.
Furthermore, Senanayaka et al.~\cite{senanayaka2022similarity} introduced the similarity-based multi-source transfer learning (SiMuS-TL) approach. By establishing a 'mixed domain' to model similarities among various sources, Senanayaka et al. demonstrated the effectiveness of carefully selected and related data sources in transfer learning.
Finally, Kashiparekh et al.~\cite{kashiparekh2019convtimenet} with their ConvTimeNet (CTN) focused on the adaptability of pre-trained networks across diverse time scales.

While the explored studies collectively advance our understanding of transfer learning in TSC, the field remains open for further investigation. A key challenge lies in determining the most suitable source models for transfer, a task complicated by the relative scarcity of large, curated, and annotated datasets in time series analysis compared to the field of computer vision. This restricts the utility of transfer learning in TSC, as the availability of extensive and diverse datasets is crucial for developing robust and generalizable models. Furthermore, the question of developing filters that are generic enough to be effective across a wide range of applications remains unresolved. This aspect is critical for the success of transfer learning, as the applicability of a pre-trained model to new tasks depends on the universality of its learned features.
Additionally, the strategy of whether to freeze certain layers of the network during transfer or to fine-tune the entire network is another area that warrants deeper exploration.


\section{Applications - recent developments and challenges} \label{sec:app}


TSC and TSER techniques have been used to analyze and model time-dependent data in a wide range of applications.
These include human activity recognition, Earth observation, medical diagnosis including Electroencephalogram (EEG) \cite{MerlinPraveena2022app} and Electrocardiogram (ECG) \cite{Liu2021app} monitoring, air quality and pollution prediction \cite{Zaini2022app,Zhang2022app}, structural and machine health monitoring \cite{Toh2020app,Thoppil2021app}, Industrial Internet of Things (IIOT) \cite{Ren2023app}, energy consumption and anomaly detection \cite{Himeur2021app}, and bio-acoustics \cite{Stowell2022app}.

Due to the extensive range of applications that use TSC and TSER, it is infeasible to cover them all in detail in a single review. 
Therefore, in this survey, we focus on just two applications -- human activity recognition and satellite Earth observation. (References to recent reviews have been provided for the other applications mentioned above.)
These are two important but quite different domains and were chosen to give the reader an idea of the diverseness of time series use in deep learning. 
The following sections provide an overview of the use of TSC and TSER, the latest developments, and challenges in these two applications.

\subsection{Human Activity Recognition}

Human activity recognition (HAR), is the identification or monitoring of human activity through the analysis of data collected by sensors or other instruments~\cite{Gupta2022survey}. The recent growth of wearable technologies and the Internet of Things has resulted not only in the collection of large volumes of activity data~\cite{Ramanujam2021survey}, but also easy deployment of applications utilising this data to improve the safety and quality of human life~\cite{chen2021deep,Gupta2022survey}. HAR is therefore an important field of research with applications including healthcare, fitness monitoring, smart homes~\cite{Lockhart2012appl}, and assisted living~\cite{Tapia2004appl}.

Devices used to collect HAR data can be categorised as visual or sensor-based~\cite{chen2021deep,wang2019deep}. Sensor-based devices can be further categorised as object sensors (for example RFIDs embedded into objects), ambient sensors (motion sensors, WiFi or Bluetooth devices in fixed locations) and wearable sensors~\cite{wang2019deep}, including smartphones~\cite{nweke2018deep}. However, the majority of HAR studies use data from wearable sensors or visual devices~\cite{Gupta2022survey}. Additionally, human activity recognition from visual device data requires the use of computer vision techniques and is therefore out of scope for this review. Accordingly, this section reviews wearable sensor-based methods of HAR. For reviews of vision-based HAR, refer to Kong and Fu\cite{Kong2022vision} or Zhang et al.~\cite{Zhang2019vision}.

The main sensors used in wearable devices are accelerometers, gyroscopes and magnetic sensors~\cite{Ordonez2016deep}, which each collect three-dimensional spatial data over time. Inertial measurement units (IMUs) are wearable devices that combine all three sensors in one unit~\cite{Reiss2012pamap,Zhang2012uschad}. Wearable device studies typically collect data from multiple IMUs located on different parts of the body~\cite{Roggen2010opp,Sztyler2017position}. To create a dataset suitable for HAR modelling, the sensor data is split into (usually equally size) time windows~\cite{Lara2013survey}. The task is then to learn a function that maps the multi-variate sensor data for each time window to a set of activities. Thus, the data forms multi-variate time series suited to TSC.

Given the broad scope of our survey, this section necessarily only provides a brief overview of the studies using deep learning for HAR. However, there are several surveys that provide a more in-depth review of machine learning and deep learning for HAR. Lara and Labrador~\cite{Lara2013survey} provide a comprehensive introduction to HAR, including machine learning methods used and the principal issues and challenges. Both Nweke et al.~\cite{nweke2018deep} and Wang et al.~\cite{wang2019deep} provide a summary of deep learning methods, highlighting their advantages and limitations. Chen et al.~\cite{chen2021deep} discuss challenges in HAR and the appropriate deep learning methods for addressing each challenge. They also provide a comprehensive list of publicly-available HAR datasets. Gu et al.~\cite{Gu2022survey} focus on deep learning methods, reviewing preprocessing and evaluation techniques as well as the deep learning models.

The deep learning methods used for HAR include both CNNs and RNNs, as well as hybrid CNN-RNN models. While some of the models include an attention module, we did not find any studies proposing a full attention or transformer model. A summary of the studies reviewed and the type of model built is provided in table \ref{tab:harpapers}. Hammerla et al.~\cite{Hammerla2016deep} compared several deep learning model for HAR, include three LSTM variants, a CNN model, and DNN model. They found a bi-directional LSTM performed best on naturalistic datasets where long-term effects are important. However, they found some applications need to focus on short-term movement patterns and suggested CNNs are more appropriate for these applications. Thus, research across all model types is beneficial for the on-going development of models for HAR applications.

{
\setlength{\tabcolsep}{4pt}
\begin{table}[!htp]
\centering
\begin{tabulary}{\textwidth \tymin=2cm}{L p{1cm} L L}
\hline
Model & Year & Embedding & Other features \\
\hline
Zeng et al.~\cite{Zeng2014cnn}          & 2014 & CNN       &  \\
DCNN~\cite{Jiang2015cnn}                & 2015 & CNN       & Discrete Fourier Transform \\
Yang et al.~\cite{Yang2015cnn}          & 2015 & CNN       &  \\
DeepConvLSTM~\cite{Ordonez2016deep}     & 2016 & CNN, LSTM &  \\
Hammerla et al.~\cite{Hammerla2016deep} & 2016 & CNN, LSTM & Bi-directional \\
Ronao et al.~\cite{Ronao2016har}        & 2016 & CNN       &  \\
Guan and Pl{\"o}tz~\cite{Guan2017ensemble} & 2017 & LSTM   & Ensemble \\
Lee et al.~\cite{Lee2017har}            & 2017 & CNN       &  \\
Murad and Pyun~\cite{Murad2017deep}     & 2017 & LSTM      & Uni- \& bi-directional \\
Ignatov~\cite{Ignatov2018realtime}      & 2018 & CNN       & Statistical features \\
Moya Rueda et al.~\cite{Rueda2018cnn}   & 2018 & CNN       &  \\
Yao et al.~\cite{Yao2018cnn}            & 2018 & CNN       & Fully convolutional \\
Zeng et al.~\cite{Zeng2018rnn}          & 2018 & LSTM      & 2 attention layers \\
AttnSense~\cite{Ma2019attnsense}        & 2019 & CNN, GRU  & Fast Fourier Transform, 2 attention layers\\
InnoHAR~\cite{Xu2019innohar}            & 2019 & CNN, GRU  & Inception \\
Zhang et al.~\cite{Zhang2020novel}      & 2020 & CNN       & Attention \\
Challa et al.~\cite{Challa2021multi}    & 2021 & CNN, LSTM & Bi-directional \\
CNN-biGRU~\cite{Mekruksavanich2021deep} & 2021 & CNN, GRU  & Bi-directional \\
DEBONAIR~\cite{Chen2021har}             & 2021 & ConvLSTM  &  \\
Mekruksavanich and Jitpattanakul~\cite{Mekruksavanich2021lstm} & 2021 & CNN, LSTM &  \\
Mekruksavanich and Jitpattanakul~\cite{Mekruksavanich2021biometric} & 2021 & CNN, LSTM & Ensemble \\
Nafea et al.~\cite{Nafea2021sensor}     & 2021 & CNN, LSTM & Bi-directional \\
Singh et al.~\cite{Singh2021convlstm}   & 2021 & CNN, LSTM & Attention \\
Wang et al.~\cite{Wang2022deep}         & 2022 & CNN       &  \\
Xu et al.~\cite{Xu2022deform}           & 2022 & CNN, Resnet & Deformable convolutions \\
\hline
\noalign{\smallskip}
\end{tabulary}
\caption{Summary of HAR deep learning models}
\vspace{-0.7 cm}
\label{tab:harpapers}
\end{table}
}

Many of the papers reviewed in this section used commonly available datasets to build and evaluate their models. A summary of the most commonly used datasets is provided in section \ref{har-datasets} of the Appendix.

\vspace{-0.25cm}
\subsubsection{Convolutional neural networks}
One of the most common types of convolutional kernels for HAR is the $k \times 1$ kernel. This kernel convolves $k$ time steps together, moving along each time series in the input features in turn~\cite{Wang2022deep}, so while weights are shared between the input features, there is no mixing between features.
The outputs from the final convolutional layer are flattened and processed by fully-connected layers before the final classification is made. Ronao et al.~\cite{Ronao2016har} performed a comprehensive evaluation of CNN models for HAR, evaluating the effect of changing the number of layers, filters and filter sizes. The input data was collected from smartphone accelerometer and gyroscope sensors. Ignatov~\cite{Ignatov2018realtime} used a one-layer CNN, and augmented the extracted features with statistical features before being passed to fully-connected layers. The architecture was effective with short time series (1 second) so useful for real time activity modelling.


One drawback of the above method is that it forces weight-sharing across all the input features. This may not be optimal, especially when using data collected from multiple devices. In this case, using a separate CNN for each device~\cite{Rueda2018cnn} allows independent weighting of the features. Similarly, as each sensor is typically tri-axial, a separate CNN can be used for each axis~\cite{Zeng2014cnn,Zhang2020novel}. The features extracted by each CNN are then concatenated and processed either by fully-connected layers~\cite{Zeng2014cnn} or an attention head~\cite{Zhang2020novel}.

While the above two methods are the most common, other studies have proposed alternative CNNs for HAR. DCNN~\cite{Jiang2015cnn} pre-processes the sensor data using a Discrete Fourier Transform to convert IMU data to frequency signals, then uses two-dimensional convolutions to extract combined temporal and frequency features. Lee et al.~\cite{Lee2017har} pre-processed the tri-axial accelerometer data to a magnitude vector, which was then processed in parallel by CNNs with varying kernel sizes, extracting features at different scales. Xu et al.~\cite{Xu2022deform} used deformable convolutions~\cite{Dai2017deform} in both a 2D-CNN and a ResNet model and found these models performed better than their non-deformable counterparts. Yao et al.~\cite{Yao2018cnn} proposed a fully convolutional model using two-dimensional temporal and feature convolutions. Their model has two advantages as (1) it handles arbitrary length input sequences and (2) it makes a prediction for each timestep, which avoids the need to pre-process the data into windows and can detect transitions between activities.
\vspace{-0.25cm}
\subsubsection{Recurrent neural networks}
Several long short-term memory (LSTM) models have been proposed for HAR.
Murad and Pyun~\cite{Murad2017deep} designed and compared three multi-layered LSTMs, a uni-directional LSTM, a bi-directional LSTM, and a ``cascading'' LSTM, which has a bi-directional first layer, followed by uni-directional layers. In each case the output from all time steps is used as input to the classification layer.
Zeng et al.~\cite{Zeng2018rnn} added two attention layers to an LSTM, a sensor attention layer before the LSTM and a temporal attention layer after the LSTM. They include a regularisation term they called ``continuous attention'' to smooth the transition between attention weights.
Guan and Pl{\"o}tz~\cite{Guan2017ensemble} created an ensemble of LSTM models by saving the models at every training epoch, then selecting the best ``M'' models based on validation set results, thus aiming to reduce model variance.
\vspace{-0.25cm}
\subsubsection{Hybrid models}
Many recent studies have focussed on hybrid models, combining both CNNs and RNNs.
DeepConvLSTM~\cite{Ordonez2016deep} comprises four temporal convolutional layers followed by two LSTM layers, which the authors found to perform better than an equivalent CNN (replacing the LSTM layers with fully-connected layers). As the LSTM layers have fewer parameters than fully-connected layers, the DeepConvLSTM model was also much smaller.
Singh et al.~\cite{Singh2021convlstm} used a CNN to encode the spatial data (i.e. the sensor readings at each timestamp) followed by a single LSTM layer to encode the temporal data, then a self-attention layer to weight the time steps. They found this model performed better than an equivalent model using temporal convolutions in the CNN layers.
Challa et al.~\cite{Challa2021multi} proposed using three 1D-CNNs with different kernel sizes in parallel, followed by 2 bi-directional LSTM layers and a fully-connected layer.
Nafea et al.~\cite{Nafea2021sensor} also used 1D-CNNs with different kernel sizes and bi-directional LSTMs. However, they used separate branches for the CNNs and LSTMs, merging the features extracted in each branch for the final fully connected layer.
Mekruksavanich and Jitpattanakul~\cite{Mekruksavanich2021lstm} compared a 4-layer CNN-LSTM model with a smaller CNN-LSTM model and LSTM models, finding the extra convolutional layers improved performance over the smaller models.
DEBONAIR~\cite{Chen2021har} is another multi-layered model. It uses parallel 1D-CNNs, each having different kernel, filter, and pooling sizes to extract different types of features associated with different types of activity. These are followed by a combined 1D-CNN, then two LSTM layers.
Mekruksavanich and Jitpattanakul~\cite{Mekruksavanich2021biometric} ensembled four different models: a CNN, an LSTM, a CNN-LSTM, and a ConvLSTM model. They aimed to produce a model for boimetric user identification that could not only identify the activity being performed, but also the participant performing the activity.

A few hybrid models use GRUs instead of LSTMs.
InnoHAR~\cite{Xu2019innohar} is a modified DeepConvLSTM~\cite{Ordonez2016deep}, replacing the four CNN layers with inception layers and the two LSTM layers with GRU layers. The authors found this inception model performed better than both the original DeepConvLSTM model and a straight CNN model~\cite{Yang2015cnn}. 
AttnSense~\cite{Ma2019attnsense} uses a Fast Fourier transform to generate frequency features which are then convolved separately for each time step. Attention layers are used to weight the extracted frequency features. These are then passed through a GRU with temporal attention to extract temporal features.
CNN-BiGRU~\cite{Mekruksavanich2021deep} uses a CNN layer to extract spatial features from the sensor data, then one or more GRU layers extract temporal features. The final section of the model is a fully-connected module consisting of one or more hidden layers and a softmax output layer.
\vspace{-0.4cm}
\subsection{Satellite Earth Observation}

Ever since NASA launched the first Landsat satellite in 1972~\cite{Wulder2008landsat}, Earth-observing satellites have been recording images of the Earth’s surface, providing 50 years of continuous Earth observation (EO) data that can be used to estimate environmental variables informing us about the state of the Earth. Instruments on board the satellites record reflected or emitted electromagnetic radiation from the Earth's surface and vegetation~\cite{Emery2017emr}. The regular, repeated observations from these instruments form satellite image time series (SITS) that are useful for analysing the dynamic properties of some variables, such as plant phenology. The main modalities used for SITS analysis are multispectral spectrometers and spectroradiometers, which observe the visible and infrared (IR) frequencies and Synthetic Aperture Radar (SAR) systems which emit a microwave signal and measure the backscatter. A list of the main satellites and instruments used in the studies reviewed is provided in section \ref{earth-observation} of the Appendix.

Raw data collected by satellite instruments needs to be pre-processed before being used in machine learning. This is frequently done by the data providers to produce analysis ready datasets (ARD). With the increasing availability of compatible ARD datasets from sources such as Google Earth Engine~\cite{Gorelick2017gee} and various data cubes~\cite{Giuliani2017cube,Lewis2017cube}, models combining data from multiple data sources (multi-modal) are becoming more common. These data sources make it straightforward to obtain data that are co-registered (spatially aligned and with the same resolution and projection), thus avoiding the need for complex pre-processing.

Satellite image time series can be processed either (1) as two-dimensional temporal and spectral data, processing each pixel independently and ignoring the spatial dimensions, or (2) as four-dimensional data, including the two spatial dimensions, thus models extract spatio-temporal features. This latter method allows estimates to be made at pixel, patch, or object level, however it requires either more complex models, or spatial features to be extracted in a pre-processing step. Feature extraction can be as simple as extracting the mean value for each band. However, both clustering (TASSEL,~\cite{Ienco2020tassel}), and neural-network based methods, such as the Pixel-Set Encoder~\cite{Garnot2020tae} have been used for more complex feature extraction.

The most common use of SITS deep learning is for the classification of the Earth's surface by land cover and agricultural land by crop types. The classes used can range from very broad land cover categories (such as forest, grasslands, agriculture) through to specific crops types. Other classification tasks include identifying specific features, such as sink-holes~\cite{Kulshrestha2022hole}, burnt areas~\cite{Ban2020fire}, flooded areas~\cite{Rambour2020flood}, roads~\cite{KamdemDeTeyou2020road}, deforestation~\cite{Matosak2022forest}, vegetation quality~\cite{Minh2017vege} and forest understory and litter types~\cite{Labenski2022under}.

Extrinsic regression tasks are less common than classification tasks, but several recent studies have investigated methods of estimating water content in vegetation, as measured by the variable Live Fuel Moisture Content (LFMC)~\cite{Rao2020lfmc,Zhu2020lfmc,Miller2022lfmc,Xie2022lfmc}. Other regression tasks include estimating the wood volume of forests~\cite{Lahssini2022wood} by using a hybrid CNN-MLP model combining a time series of Sentinel-2 images with a single LiDAR image and crop yield~\cite{Sun2020yield} which uses a hybrid of CNN and LSTM. 

Many different approaches to learning from SITS data have been studied, with studies using all the main deep learning architectures, adapting them for multi-modal learning, and combining architectures in hybrid and ensemble models. The rest of this section reviews the architectures that have been used to model SITS data. A summary of these papers and architectures is provided in table \ref{table:sits}.

{
\setlength{\tabcolsep}{11pt}
\begin{table}[!htp]
    \centering
    \begin{tabular}{l l l l}
    \hline
    Model & Year & Embedding & Other features \\
    \hline
    \multicolumn{4}{l}{Crop type classification} \\
    \hline
    TAN~\cite{Li2019tan} & 2019 & 2D-CNN \& GRU & Attention --- temporal \\
    TGA~\cite{Li2020tga} & 2020 & 2D-CNN & Attention --- squeeze and excitation \\
    3D-CNN~\cite{Ji20183dcnn} & 2018 & 3D-CNN & \\
    DCM~\cite{Xu2020dcm} & 2020 & LSTM & Self-attention \\
    HierbiLSTM~\cite{Barriere2022lstm} & 2022 & LSTM & Self-attention \\
    L-TAE~\cite{Garnot2020ltae} & 2020 & MLP & Attention --- temporal \\
    PSE-TAE~\cite{Garnot2020tae,Ofori-Ampofo2021att} & 2020 & MLP & Attention --- temporal optionally Multi-modal\\
    SITS-BERT~\cite{Yuan2021sitsbert} & 2021 & & Pre-trained transformer \\
    \hline
    \multicolumn{4}{l}{Land Cover classification} \\
    \hline
    1D-CNN~\cite{DiMauro2017tiselc} & 2017 & 1D-CNN \& MLP & Hybrid model \\
    1D \& 2D-CNNs~\cite{Kussul2017cnn} & 2017 & 1D-CNN; 2D-CNN & Ensemble model \\
    TempCNN~\cite{Pelletier2019tempcnn} & 2019 & 1D-CNN & \\
    TASSEL~\cite{Ienco2020tassel} & 2020 & 1D-CNN & Self-attention \\
    TSI~\cite{Dou2021tsi} & 2021 & 1D-CNN; LSTM & Ensemble model \\
    TWINNS~\cite{Ienco2019twinns} & 2019 & 2D-CNN \& GRU & Attention --- temporal;  Multi-modal \\
    DuPLO~\cite{Interdonato2019duplo} & 2019 & 2D-CNN \& GRU & Attention --- temporal \\
    Sequential RNN~\cite{Russwurm2018seqrnn} & 2018 & 2D-FCN \& LSTM & Hybrid model \\
    FG-UNET~\cite{Stoian2019fgunet} & 2019 & UNet \& 2D-CNN & Hybrid model \\
    LSTM~\cite{Ienco2017rnn} & 2017 & LSTM & \\
    HOb2sRNN~\cite{Gbodjo2020hob2srnn} & 2020 & GRU & Attention --- temporal \\
    OD2RNN~\cite{Ienco2019od2rnn} & 2019 & GRU & Attention --- temporal; Multi-modal \\
    SITS-Former~\cite{Yuan2022sitsformer} & 2022 & 3D-CNN & Pre-trained transformer \\
    \hline
    \multicolumn{4}{l}{Other classification tasks} \\
    \hline
    Deforestation~\cite{Matosak2022forest} & 2022 & U-Net \& LSTM & Hybrid model \\
    Flood detection~\cite{Rambour2020flood} & 2020 & Resnet \& GRU & Hybrid model \\
    Forest understory~\cite{Labenski2022under} & 2022 & 2D-CNN \& LSTM & Ensemble model \\
    Road detection~\cite{KamdemDeTeyou2020road} & 2020 & U-Net \& convLSTM & Hybrid model \\
    Vegetation quality~\cite{Minh2017vege} & 2017 & LSTM; GRU & \\
    \hline
    \multicolumn{4}{l}{Extrinsic regression tasks} \\
    \hline
    TempCNN-LFMC~\cite{Zhu2020lfmc} & 2021 & 1D-CNN & \\
    Multi-tempCNN~\cite{Miller2022lfmc} & 2022 & 1D-CNN & Multi-modal, ensemble  model \\
    LFMC estimation~\cite{Rao2020lfmc} & 2020 & LSTM & Multi-modal \\
    LFMC estimation~\cite{Xie2022lfmc} & 2022 & 1D-CNN \& LSTM & Multi-modal, hybrid, ensemble \\
    MLDL-net~\cite{Sun2020yield} & 2020 & 2D-CNN \& LSTM & Hybrid model \\
    SSTNN~\cite{Qiao2021yield} & 2021 & 3D-CNN \& LSTM & Hybrid model \\
    MMFVE~\cite{Lahssini2022wood} & 2022 & 2D-CNN & Hybrid model \\
    \hline
    \noalign{\smallskip}
    \end{tabular}
    \caption{Summary of SITS deep learning models}
    \vspace{-1cm}
    \label{table:sits}
\end{table}
}
\vspace{-0.25cm}
\subsubsection{Recurrent Neural Networks (RNNs)}

One of the first papers to use RNNs for land cover classification was Ienco et al.~\cite{Ienco2017rnn}, who showed an LSTM model out-performed non deep learning methods such as Random Forest (RF) and Support Vector Machines (SVM). However, they also showed that the performance of both RF and SVM improves if trained on features extracted by the LSTM model, and in some cases were more accurate than the straight LSTM model. Rao et al.~\cite{Rao2020lfmc} used an extrinsic regression LSTM model to estimate LFMC in the western United States.

More commonly, however, RNNs are combined with an attention layer to allow the model to focus on the most important time steps. The OD2RNN model~\cite{Ienco2019od2rnn}, used separate GRU layers followed by attention layers to process Sentinel-1 and Sentinel-2 data, combining the features extracted by each source for the final fully-connected layers. HOb2sRNN~\cite{Gbodjo2020hob2srnn} refined OD2RNN by using a hierarchy of land cover classifications; the model was pretrained using broad land cover classifications, then further trained using the finer-grained classifications. DCM~\cite{Xu2020dcm} and HierbiLSTM~\cite{Barriere2022lstm} both use a bi-directional LSTM, processing the time series in both directions, followed by a self-attention transformer for a pixel-level crop-mapping model. All these studies found adding the attention layers improved model performance over a straight GRU or LSTM model.
\vspace{-0.25cm}
\subsubsection{Convolutional Neural Networks (CNNs)}
While many authors have claimed that RNNs out-perform CNNs for land cover and crop type classification, most of these comparisons are to 2-dimensional CNNs (2D-CNN), that ignore the temporal ordering of SITS data~\cite{Pelletier2019tempcnn}. However, other studies show using 1-dimensional CNNs (1D-CNNs) to extract temporal information or 3-dimensional CNNs (3D-CNNs) to extract spatio-temporal information are both effective methods of learning from SITS data.
TempCNN~\cite{Pelletier2019tempcnn} consists of three 1D convolutional layers. The output from the final convolutional layer is passed through a fully-connected layer, then the final softmax classification layer. TASSEL~\cite{Ienco2020tassel}, is an adaptation of TempCNN for OBIA classification, using TempCNN models to process features extracted from the objects, followed by an attention layer to weight the convolved features. TempCNN has also been adapted for extrinsic regression~\cite{Zhu2020lfmc} and used for LFMC estimation~\cite{Zhu2020lfmc,Miller2022lfmc,Xie2022lfmc}.

2D-CNNs are mainly used to extract spatial or spatio-temporal features for both pixel and object classification. The model input is usually 4-dimensional and the data is convolved spatially, with two main methods used to handle the temporal dimension. In the first method, each time step is convolved separately and the extracted features are merged in later stages of the model~\cite{Li2020tga}. In the second method, the time steps and channels are flattened to form a large multivariate image~\cite{Kussul2017cnn,Lahssini2022wood}. FG-UNet~\cite{Stoian2019fgunet} is a fully-convolutional model that combines both the above methods, first grouping time steps by threes to produce images with 30 channels (10 spectral $\times$ 3 temporal), which are passed through both U-Net and 2D-CNN layers.

Ji et al.~\cite{Ji20183dcnn} used a three-dimensional CNN (3D-CNN) to convolve the spatial and temporal dimensions together, combining the strengths of 1D-CNN and 2D-CNNs. The study found a 3D-CNN crop classification model performed significantly better than the 2D-CNN, again showing the importance of the temporal features. Another study, SSTNN~\cite{Qiao2021yield} obtained good results for crop yield prediction by using a 3D-CNN to convolve the spatial and spectral dimensions, extracting spatio-spectral features for each time step. These features were then processed by LSTM layers to perform the temporal modelling.
\vspace{-0.25cm}    
\subsubsection{Transformer and Attention Models}

As an alternative to including attention layers with a CNN or RNN, several studies have designed models that process temporal information using only attention layers. PSE-TAE~\cite{Garnot2020tae} used a modified transformer called a temporal attention encoder (TAE) for crop mapping and found the TAE performed better than either a CNN or an RNN. L-TAE~\cite{Garnot2020ltae} replaced the TAE with a light-weight transformer which is both computationally efficient and more accurate than the full TAE. Ofori-Ampofo et al.~\cite{Ofori-Ampofo2021att} adapted the TAE model for multi-modal inputs, using Sentinel-1 and Sentinel-2 data for crop type mapping. Rußwurm and Körner\cite{Russwurm2020att} compared a self-attention model with RNN and CNN architectures. They found that this model was more robust to noise than either RNN or CNN and suggested self-attention is suitable for processing raw, cloud-affected satellite data.

Building on the success of pre-trained transformers for natural language processing (NLP) such as BERT~\cite{devlin2018bert}, pre-trained transformers have been proposed for EO tasks~\cite{Yuan2021sitsbert}. Earth observation tasks are particularly suited for pre-trained models as large quantities of EO data are readily available, while labelled data can be difficult to obtain~\cite{Tuia2016da}, especially in remote locations. SITS-BERT~\cite{Yuan2021sitsbert} is an adaptation of BERT~\cite{devlin2018bert} for pixel-based SITS classification. For the pretext task, random noise is added to the pixels, and the model is trained to identify and remove this noise. The pre-trained model is then further trained for required tasks such as crop type or land cover mapping. SITS-Former~\cite{Yuan2022sitsformer} modifies SITS-BERT for patch classification by using 3D-Conv layers to encode the spatial-spectral information, which is then passed through the temporal attention layers. The pretext task used for SITS-Former is to predict randomly masked pixels.
\vspace{-0.25cm}
\subsubsection{Hybrid Models}

A common use of hybrid models is to use a CNN to extract spatial features and an RNN to extract temporal features. Garnot et al.~\cite{Garnot2019time} compared a straight 2D-CNN model (thus ignoring the temporal aspect), a straight GRU model (thus ignoring the spatial aspect) and a combined 2D-CNN and GRU model (thus using both spatial and temporal information) and found the combined model gave the best results, demonstrating that both the spatial and temporal dimensions provide useful information for land cover mapping and crop classification. DuPLO~\cite{Interdonato2019duplo} was one of the first models to exploit this method, running a CNN and ConvGRU model in parallel, then fusing the outputs using a fully-connected network for the final classifier. During training, an auxiliary classifier for each component was used to enhance the discriminative power. TWINNS~\cite{Ienco2019twinns} extended DuPLO to a multi-modal model, using time series of both Sentinel-1 (SAR) and Sentinel-2 (Optical) images. Each modality was processed by separate CNN and convGRU models, then the output features from all four models were fused for classification.

Other hybrid models include 
Li et al.~\cite{Li2019tan}, who used a CNN for spatial and spectral unification of Landsat-8 and Sentinel-2 images which were then processed by a GRU.
MLDL-Net~\cite{Sun2020yield} is a 2D-CNN extrinsic regression model, using CNNs to extract time step features, which are then passed through an LSTM model to extract temporal features. Fully connected layers combine the feature sets to predict crop yield.
Rußwurm and Körner~\cite{Russwurm2018seqrnn} extracted temporal features first, using a bi-directional LSTM, then used a fully-convolutional 2D-CNN to incorporate spatial information and classify each pixel in the input patch. 
\vspace{-0.25cm}
\subsubsection{Ensemble Models}

One of the easiest ways to ensemble DL models is to train multiple homogeneous models, that vary only in the random weight initialisation~\cite{Fawaz2019ensembles}. Di Mauro et al.~\cite{DiMauro2017tiselc} ensembled 100 LULC models with different weight initialisations by averaging the softmax predictions. They found this produced a more stable and stronger classifier that outperformed the individual models. Multi-tempCNN~\cite{Miller2022lfmc}, a model for LFMC estimation, is an ensemble of homogeneous models for extrinsic regression. The authors suggested that as an additional benefit, the variance of the individual model predictions can be used to obtain a measure of uncertainty of the estimates. TSI~\cite{Dou2021tsi} also ensembles a set of homogeneous models, but instead of relying on random weight initialisation to introduce model diversity, the time series are segmented and models trained on each segment.

Other methods create ensembles of heterogeneous models. Kussul et al~\cite{Kussul2017cnn} compared ensembles of 1D-CNNs and 2D-CNNs models for land cover classification. Each model in the ensemble used a different number of filters, so finding different feature sets useful for classification. 
Xie et al.~\cite{Xie2022lfmc} ensembled three heterogeneous models --- a causal temporal convolutional neural network (TCN), an LSTM, and a hybrid TCN-LSTM model --- for an extrinsic regression model to estimate LFMC. The ensembles were created using stacking~\cite{Wolpert1992stack}. The authors compared this method to boosting their TCN-LSTM model, using Adaboost~\cite{Freund1996ada} to create a three-member ensemble, and found that stacking a diverse set of models out-performed boosting.
\vspace{-0.25cm}
\subsubsection{EO Surveys and Reviews}
This survey is one of very few that include a section focusing specifically on deep learning TSC and TSER tasks using SITS data. However, there are other reviews that provide further information about related topics.
Gomez et al.~\cite{Gomez2016review} is an older review highlighting the importance role of SITS data for land cover classification.
Zhu et al.~\cite{Zhu2017review} reviewed the advances and challenges in DL for remote sensing, and the resources available that are potentially useful to help DL address some of the major challenges facing humanity.
Ma et al.~\cite{Ma2019review} studies the role of deep learning in Earth observation using remotely sensed data. It covers a broad range of tasks including image fusion, image segmentation and object-based analysis, as well as classification tasks.
Yuan et al.~\cite{Yuan2020review} provide a review of DL applications for remote sensing, comparing the role of DL versus physical modelling of environmental variables and highlighting challenges in DL for remote sensing that need to be addressed.
Chaves et al.~\cite{Chaves2020review} reviewed recent research using Landsat 8 and/or Sentinel-2 data for land cover mapping. While not focused on SITS DL methods, the review notes the growing importance of these methods.
Moskolai et al.~\cite{Moskolai2021apps} is a review of forecasting applications using DL with SITS data that provides an analysis of the main DL architectures that are relevant for classification as well as forecasting.

\vspace{-0.2cm}
\section{Conclusion} 
\vspace{-0.2cm}
In conclusion, this survey paper has discussed a variety of deep network architectures for time series classification and extrinsic regression tasks, including multilayer perceptrons, convolutional neural networks, recurrent neural networks, and attention-based models. We have also highlighted refinements that have been made to improve the performance of these models on time series tasks. Additionally, we have discussed two critical applications of time series classification and regression, human activity recognition and satellite Earth observation. Overall, using deep network architectures and refinements has enabled significant progress in the field of time series classification and will continue to be essential for addressing a wide range of real-world problems. We hope this survey will stimulate further research using deep learning techniques for time series classification and extrinsic regression. Additionally, we provide a carefully curated collection of sources, available at \url{https://github.com/Navidfoumani/TSC_Survey}, to further support the research community.
\vspace{-0.25cm}
\begin{acks}
This work was supported by an Australian Government Research Training Program (RTP) scholarship.
\end{acks}
\balance
\vspace{-0.3cm}

\balance
\bibliographystyle{IEEEtran}
\bibliography{references}

\begin{thebibliography}{100}
\providecommand{\url}[1]{#1}
\csname url@samestyle\endcsname
\providecommand{\newblock}{\relax}
\providecommand{\bibinfo}[2]{#2}
\providecommand{\BIBentrySTDinterwordspacing}{\spaceskip=0pt\relax}
\providecommand{\BIBentryALTinterwordstretchfactor}{4}
\providecommand{\BIBentryALTinterwordspacing}{\spaceskip=\fontdimen2\font plus
\BIBentryALTinterwordstretchfactor\fontdimen3\font minus \fontdimen4\font\relax}
\providecommand{\BIBforeignlanguage}[2]{{%
\expandafter\ifx\csname l@#1\endcsname\relax
\typeout{** WARNING: IEEEtran.bst: No hyphenation pattern has been}%
\typeout{** loaded for the language `#1'. Using the pattern for}%
\typeout{** the default language instead.}%
\else
\language=\csname l@#1\endcsname
\fi
#2}}
\providecommand{\BIBdecl}{\relax}
\BIBdecl

\bibitem{yang200610}
Q.~Yang and X.~Wu, ``10 challenging problems in data mining research,'' \emph{Int. J. Inf. Tech. \& Decision Making}, vol.~5, no.~04, pp. 597--604, 2006.

\bibitem{esling2012time}
P.~Esling and C.~Agon, ``Time-series data mining,'' \emph{ACM Computing Surveys (CSUR)}, vol.~45, no.~1, pp. 1--34, 2012.

\bibitem{nweke2018deep}
H.~F. Nweke, Y.~W. Teh, M.~A. Al-Garadi, and U.~R. Alo, ``Deep learning algorithms for human activity recognition using mobile and wearable sensor networks: State of the art and research challenges,'' \emph{Expert Systems with Applications}, vol. 105, pp. 233--261, 2018.

\bibitem{wang2019deep}
J.~Wang, Y.~Chen, S.~Hao, X.~Peng, and L.~Hu, ``Deep learning for sensor-based activity recognition: A survey,'' \emph{Pattern recognition letters}, vol. 119, pp. 3--11, 2019.

\bibitem{chen2021deep}
K.~Chen, D.~Zhang, L.~Yao, B.~Guo, Z.~Yu, and Y.~Liu, ``Deep learning for sensor-based human activity recognition: Overview, challenges, and opportunities,'' \emph{ACM Computing Surveys (CSUR)}, vol.~54, no.~4, pp. 1--40, 2021.

\bibitem{schirrmeister2017deep}
R.~T. Schirrmeister, J.~T. Springenberg, L.~D.~J. Fiederer, M.~Glasstetter, K.~Eggensperger, M.~Tangermann, F.~Hutter, W.~Burgard, and T.~Ball, ``{Deep learning with convolutional neural networks for EEG decoding and visualization},'' \emph{Human brain mapping}, vol.~38, no.~11, pp. 5391--5420, 2017.

\bibitem{rajkomar2018scalable}
A.~Rajkomar, E.~Oren, K.~Chen, A.~M. Dai, N.~Hajaj, M.~Hardt, P.~J. Liu, X.~Liu, J.~Marcus, M.~Sun \emph{et~al.}, ``Scalable and accurate deep learning with electronic health records,'' \emph{NPJ digital medicine}, vol.~1, no.~1, pp. 1--10, 2018.

\bibitem{bagnall2018uea}
A.~Bagnall, H.~A. Dau, J.~Lines, M.~Flynn, J.~Large, A.~Bostrom, P.~Southam, and E.~Keogh, ``{The UEA multivariate time series classification archive, 2018},'' \emph{arXiv preprint:1811.00075}, 2018.

\bibitem{dau2019ucr}
H.~A. Dau, A.~Bagnall, K.~Kamgar, C.-C.~M. Yeh, Y.~Zhu, S.~Gharghabi, C.~A. Ratanamahatana, and E.~Keogh, ``{The UCR time series archive},'' \emph{IEEE/CAA Journal of Automatica Sinica}, vol.~6, no.~6, pp. 1293--1305, 2019.

\bibitem{tan2021time}
C.~W. Tan, C.~Bergmeir, F.~Petitjean, and G.~I. Webb, ``Time series extrinsic regression,'' \emph{Data Min. Knowl. Discov.}, vol.~35, no.~3, pp. 1032--1060, 2021.

\bibitem{middlehurst2023bake}
M.~Middlehurst, P.~Sch{\"a}fer, and A.~Bagnall, ``Bake off redux: a review and experimental evaluation of recent time series classification algorithms,'' \emph{arXiv preprint arXiv:2304.13029}, 2023.

\bibitem{fawaz2020inceptiontime}
H.~I. Fawaz, B.~Lucas, G.~Forestier, C.~Pelletier, D.~F. Schmidt, J.~Weber, G.~I. Webb, L.~Idoumghar, P.-A. Muller, and F.~Petitjean, ``Inceptiontime: Finding alexnet for time series classification,'' \emph{Data Min. Knowl. Discov.}, vol.~34, no.~6, pp. 1936--1962, 2020.

\bibitem{Foumani2023}
N.~M. Foumani, C.~W. Tan, G.~I. Webb, and M.~Salehi, ``Improving position encoding of transformers for multivariate time series classification,'' \emph{Data Min. Knowl. Discov.}, Sep 2023.

\bibitem{dempster2019rocket}
A.~Dempster, F.~Petitjean, and G.~I. Webb, ``{ROCKET: exceptionally fast and accurate time series classification using random convolutional kernels},'' \emph{Data Min. Knowl. Discov.}, vol.~34, no.~5, pp. 1454--1495, 2020.

\bibitem{fawaz2019deep}
H.~I. Fawaz, G.~Forestier, J.~Weber, L.~Idoumghar, and P.-A. Muller, ``Deep learning for time series classification: a review,'' \emph{Data Min Knowl Discov}, vol.~33, no.~4, pp. 917--963, 2019.

\bibitem{wang2017time}
Z.~Wang, W.~Yan, and T.~Oates, ``Time series classification from scratch with deep neural networks: A strong baseline,'' in \emph{2017 International joint conference on neural networks (IJCNN)}.\hskip 1em plus 0.5em minus 0.4em\relax IEEE, 2017, pp. 1578--1585.

\bibitem{wen2022transformers}
Q.~Wen, T.~Zhou, C.~Zhang, W.~Chen, Z.~Ma, J.~Yan, and L.~Sun, ``Transformers in time series: A survey,'' \emph{arXiv preprint:2202.07125}, 2022.

\bibitem{hao2020new}
Y.~Hao and H.~Cao, ``A new attention mechanism to classify multivariate time series,'' in \emph{29th Int. Joint Conf. Artificial Intelligence}, 2020.

\bibitem{zerveas2021transformer}
G.~Zerveas, S.~Jayaraman, D.~Patel, A.~Bhamidipaty, and C.~Eickhoff, ``A transformer-based framework for multivariate time series representation learning,'' in \emph{27th ACM SIGKDD Conference on Knowledge Discovery \& Data Mining}, 2021, pp. 2114--2124.

\bibitem{liu2021self}
X.~Liu, F.~Zhang, Z.~Hou, L.~Mian, Z.~Wang, J.~Zhang, and J.~Tang, ``Self-supervised learning: Generative or contrastive,'' \emph{IEEE transactions on knowledge and data engineering}, vol.~35, no.~1, pp. 857--876, 2021.

\bibitem{eldele2021time}
E.~Eldele, M.~Ragab, Z.~Chen, M.~Wu, C.~K. Kwoh, X.~Li, and C.~Guan, ``Time-series representation learning via temporal and contextual contrasting,'' in \emph{Proceedings of the Thirtieth International Joint Conference on Artificial Intelligence, {IJCAI-21}}, 2021, pp. 2352--2359.

\bibitem{yang2021voice2series}
C.-H.~H. Yang, Y.-Y. Tsai, and P.-Y. Chen, ``Voice2series: Reprogramming acoustic models for time series classification,'' in \emph{Int. conf. mach. learn.}\hskip 1em plus 0.5em minus 0.4em\relax PMLR, 2021, pp. 11\,808--11\,819.

\bibitem{yue2022ts2vec}
Z.~Yue, Y.~Wang, J.~Duan, T.~Yang, C.~Huang, Y.~Tong, and B.~Xu, ``Ts2vec: Towards universal representation of time series,'' in \emph{AAAI}, vol.~36, no.~8, 2022, pp. 8980--8987.

\bibitem{foumani2023series2vec}
N.~M. Foumani, C.~W. Tan, G.~I. Webb, and M.~Salehi, ``Series2vec: Similarity-based self-supervised representation learning for time series classification,'' \emph{arXiv preprint arXiv:2312.03998}, 2023.

\bibitem{bagnall2017great}
A.~Bagnall, J.~Lines, A.~Bostrom, J.~Large, and E.~Keogh, ``The great time series classification bake off: a review and experimental evaluation of recent algorithmic advances,'' \emph{Data Min. Knowl. Discov.}, vol.~31, no.~3, pp. 606--660, 2017.

\bibitem{ruiz2020great}
A.~P. Ruiz, M.~Flynn, J.~Large, M.~Middlehurst, and A.~Bagnall, ``The great multivariate time series classification bake off: a review and experimental evaluation of recent algorithmic advances,'' \emph{Data Min. Knowl. Discov.}, pp. 1--49, 2020.

\bibitem{tan2020monash}
C.~W. Tan, C.~Bergmeir, F.~Petitjean, and G.~I. Webb, ``{Monash University, UEA, UCR time series regression archive},'' \emph{arXiv preprint:2006.10996}, 2020.

\bibitem{langkvist2014review}
M.~L{\"a}ngkvist, L.~Karlsson, and A.~Loutfi, ``A review of unsupervised feature learning and deep learning for time-series modeling,'' \emph{Pattern Recognition Letters}, vol.~42, pp. 11--24, 2014.

\bibitem{bengio2013generalized}
Y.~Bengio, L.~Yao, G.~Alain, and P.~Vincent, ``Generalized denoising auto-encoders as generative models,'' \emph{Advances neural inf. process. syst.}, vol.~26, 2013.

\bibitem{hu2016transfer}
Q.~Hu, R.~Zhang, and Y.~Zhou, ``Transfer learning for short-term wind speed prediction with deep neural networks,'' \emph{Renewable Energy}, vol.~85, pp. 83--95, 2016.

\bibitem{serra2018towards}
J.~Serr{\`a}, S.~Pascual, and A.~Karatzoglou, ``Towards a universal neural network encoder for time series.'' in \emph{CCIA}, 2018, pp. 120--129.

\bibitem{banerjee2019deep}
D.~Banerjee, K.~Islam, K.~Xue, G.~Mei, L.~Xiao, G.~Zhang, R.~Xu, C.~Lei, S.~Ji, and J.~Li, ``A deep transfer learning approach for improved post-traumatic stress disorder diagnosis,'' \emph{Knowledge and Information Systems}, vol.~60, no.~3, pp. 1693--1724, 2019.

\bibitem{aswolinskiy2018time}
W.~Aswolinskiy, R.~F. Reinhart, and J.~Steil, ``Time series classification in reservoir-and model-space,'' \emph{Neural Processing Letters}, vol.~48, no.~2, pp. 789--809, 2018.

\bibitem{GanSurvey2021}
\BIBentryALTinterwordspacing
E.~Brophy, Z.~Wang, Q.~She, and T.~Ward, ``Generative adversarial networks in time series: A systematic literature review,'' \emph{ACM Comput. Surv.}, vol.~55, no.~10, feb 2023. [Online]. Available: \url{https://doi.org/10.1145/3559540}
\BIBentrySTDinterwordspacing

\bibitem{del2021auto}
F.~A. Del~Campo, M.~C.~G. Neri, O.~O.~V. Villegas, V.~G.~C. S{\'a}nchez, H.~d. J.~O. Dom{\'\i}nguez, and V.~G. Jim{\'e}nez, ``Auto-adaptive multilayer perceptron for univariate time series classification,'' \emph{Expert Systems with Applications}, vol. 181, p. 115147, 2021.

\bibitem{iwana2016robust}
B.~K. Iwana, V.~Frinken, and S.~Uchida, ``A robust dissimilarity-based neural network for temporal pattern recognition,'' in \emph{2016 15th International Conference on Frontiers in Handwriting Recognition (ICFHR)}.\hskip 1em plus 0.5em minus 0.4em\relax IEEE, 2016, pp. 265--270.

\bibitem{iwana2020dtw}
------, ``{DTW-NN: A novel neural network for time series recognition using dynamic alignment between inputs and weights},'' \emph{Knowledge-Based Systems}, vol. 188, p. 104971, 2020.

\bibitem{tabassum2022time}
N.~Tabassum, S.~Menon, and A.~Jastrzebska, ``Time-series classification with safe: Simple and fast segmented word embedding-based neural time series classifier,'' \emph{Information Processing \& Management}, vol.~59, no.~5, p. 103044, 2022.

\bibitem{krizhevsky2012imagenet}
A.~Krizhevsky, I.~Sutskever, and G.~E. Hinton, ``Imagenet classification with deep convolutional neural networks,'' \emph{Advances neural inf. process. syst.}, vol.~25, pp. 1097--1105, 2012.

\bibitem{gu2018recent}
J.~Gu, Z.~Wang, J.~Kuen, L.~Ma, A.~Shahroudy, B.~Shuai, T.~Liu, X.~Wang, G.~Wang, J.~Cai \emph{et~al.}, ``Recent advances in convolutional neural networks,'' \emph{Pattern recognition}, vol.~77, pp. 354--377, 2018.

\bibitem{lecun2015deep}
Y.~LeCun, Y.~Bengio, and G.~Hinton, ``Deep learning,'' \emph{nature}, vol. 521, no. 7553, pp. 436--444, 2015.

\bibitem{zheng2014time}
Y.~Zheng, Q.~Liu, E.~Chen, Y.~Ge, and J.~L. Zhao, ``Time series classification using multi-channels deep convolutional neural networks,'' in \emph{International Conference on Web-Age Information Management}.\hskip 1em plus 0.5em minus 0.4em\relax Springer, 2014, pp. 298--310.

\bibitem{yang2015deep}
J.~Yang, M.~N. Nguyen, P.~P. San, X.~L. Li, and S.~Krishnaswamy, ``Deep convolutional neural networks on multichannel time series for human activity recognition,'' in \emph{Twenty-fourth international joint conference on artificial intelligence}, 2015.

\bibitem{zhao2017convolutional}
B.~Zhao, H.~Lu, S.~Chen, J.~Liu, and D.~Wu, ``Convolutional neural networks for time series classification,'' \emph{Journal of Systems Engineering and Electronics}, vol.~28, no.~1, pp. 162--169, 2017.

\bibitem{long2015fully}
J.~Long, E.~Shelhamer, and T.~Darrell, ``Fully convolutional networks for semantic segmentation,'' in \emph{IEEE conf. comp. vision patt. recognit.}, 2015, pp. 3431--3440.

\bibitem{he2016deep}
K.~He, X.~Zhang, S.~Ren, and J.~Sun, ``Deep residual learning for image recognition,'' in \emph{IEEE conf. comp. vision patt. recognit.}, 2016, pp. 770--778.

\bibitem{zhou2016learning}
B.~Zhou, A.~Khosla, A.~Lapedriza, A.~Oliva, and A.~Torralba, ``Learning deep features for discriminative localization,'' in \emph{IEEE conf. comp. vision patt. recognit.}, 2016, pp. 2921--2929.

\bibitem{zou2019integration}
X.~Zou, Z.~Wang, Q.~Li, and W.~Sheng, ``Integration of residual network and convolutional neural network along with various activation functions and global pooling for time series classification,'' \emph{Neurocomputing}, vol. 367, pp. 39--45, 2019.

\bibitem{li2018csrnet}
Y.~Li, X.~Zhang, and D.~Chen, ``Csrnet: Dilated convolutional neural networks for understanding the highly congested scenes,'' in \emph{IEEE conf. comp. vision patt. recognit.}, 2018, pp. 1091--1100.

\bibitem{yazdanbakhsh2019multivariate}
O.~Yazdanbakhsh and S.~Dick, ``Multivariate time series classification using dilated convolutional neural network,'' \emph{arXiv preprint:1905.01697}, 2019.

\bibitem{foumani2021disjoint}
S.~N.~M. Foumani, C.~W. Tan, and M.~Salehi, ``Disjoint-cnn for multivariate time series classification,'' in \emph{2021 Int. Conf. Data Min. Workshops (ICDMW)}.\hskip 1em plus 0.5em minus 0.4em\relax IEEE, 2021, pp. 760--769.

\bibitem{sandler2018mobilenetv2}
M.~Sandler, A.~Howard, M.~Zhu, A.~Zhmoginov, and L.-C. Chen, ``{MobileNet-V2: Inverted residuals and linear bottlenecks},'' in \emph{IEEE conf. comp. vision patt. recognit.}, 2018, pp. 4510--4520.

\bibitem{wang2015encoding}
Z.~Wang and T.~Oates, ``Encoding time series as images for visual inspection and classification using tiled convolutional neural networks,'' in \emph{Workshops at the twenty-ninth AAAI conference on artificial intelligence}, 2015.

\bibitem{hatami2018classification}
N.~Hatami, Y.~Gavet, and J.~Debayle, ``Classification of time-series images using deep convolutional neural networks,'' in \emph{Tenth international conference on machine vision (ICMV 2017)}, vol. 10696.\hskip 1em plus 0.5em minus 0.4em\relax SPIE, 2018, pp. 242--249.

\bibitem{karimi2018scalable}
S.~Karimi-Bidhendi, F.~Munshi, and A.~Munshi, ``Scalable classification of univariate and multivariate time series,'' in \emph{2018 IEEE International Conference on Big Data (Big Data)}.\hskip 1em plus 0.5em minus 0.4em\relax IEEE, 2018, pp. 1598--1605.

\bibitem{zhao2019classify}
Y.~Zhao and Z.~Cai, ``Classify multivariate time series by deep neural network image classification,'' in \emph{2019 2nd China Symposium on Cognitive Computing and Hybrid Intelligence (CCHI)}.\hskip 1em plus 0.5em minus 0.4em\relax IEEE, 2019, pp. 93--98.

\bibitem{yang2019sensor}
C.-L. Yang, Z.-X. Chen, and C.-Y. Yang, ``Sensor classification using convolutional neural network by encoding multivariate time series as two-dimensional colored images,'' \emph{Sensors}, vol.~20, no.~1, p. 168, 2019.

\bibitem{kamphorst1987recurrence}
J.-P. E. S.~O. Kamphorst, D.~Ruelle \emph{et~al.}, ``Recurrence plots of dynamical systems,'' \emph{Europhysics Letters}, vol.~4, no.~9, p.~17, 1987.

\bibitem{szegedy2016rethinking}
C.~Szegedy, V.~Vanhoucke, S.~Ioffe, J.~Shlens, and Z.~Wojna, ``Rethinking the inception architecture for computer vision,'' in \emph{IEEE conf. comp. vision patt. recognit.}, 2016, pp. 2818--2826.

\bibitem{chen2019deep}
W.~Chen and K.~Shi, ``A deep learning framework for time series classification using relative position matrix and convolutional neural network,'' \emph{Neurocomputing}, vol. 359, pp. 384--394, 2019.

\bibitem{cui2016multi}
Z.~Cui, W.~Chen, and Y.~Chen, ``Multi-scale convolutional neural networks for time series classification,'' \emph{arXiv preprint:1603.06995}, 2016.

\bibitem{le2016data}
A.~Le~Guennec, S.~Malinowski, and R.~Tavenard, ``Data augmentation for time series classification using convolutional neural networks,'' in \emph{ECML/PKDD workshop on advanced analytics and learning on temporal data}, 2016.

\bibitem{liu2018time}
C.-L. Liu, W.-H. Hsaio, and Y.-C. Tu, ``Time series classification with multivariate convolutional neural network,'' \emph{IEEE Transactions on Industrial Electronics}, vol.~66, no.~6, pp. 4788--4797, 2018.

\bibitem{brunel2019cnn}
A.~Brunel, J.~Pasquet, J.~PASQUET, N.~Rodriguez, F.~Comby, D.~Fouchez, and M.~Chaumont, ``A cnn adapted to time series for the classification of supernovae,'' \emph{Electronic imaging}, vol. 2019, no.~14, pp. 90--1, 2019.

\bibitem{sun2021prototypical}
J.~Sun, S.~Takeuchi, and I.~Yamasaki, ``Prototypical inception network with cross branch attention for time series classification,'' in \emph{2021 International Joint Conference on Neural Networks (IJCNN)}.\hskip 1em plus 0.5em minus 0.4em\relax IEEE, 2021, pp. 1--7.

\bibitem{usmankhujaev2021time}
S.~Usmankhujaev, B.~Ibrokhimov, S.~Baydadaev, and J.~Kwon, ``Time series classification with inceptionfcn,'' \emph{Sensors}, vol.~22, no.~1, p. 157, 2021.

\bibitem{gong2022kdctime}
X.~Gong, Y.-W. Si, Y.~Tian, C.~Lin, X.~Zhang, and X.~Liu, ``Kdctime: Knowledge distillation with calibration on inceptiontime for time-series classification,'' \emph{Inf. Sci.}, vol. 613, pp. 184--203, 2022.

\bibitem{ismail2023lite}
A.~Ismail-Fawaz, M.~Devanne, S.~Berretti, J.~Weber, and G.~Forestier, ``Lite: Light inception with boosting techniques for time series classification,'' in \emph{2023 IEEE 10th International Conference on Data Science and Advanced Analytics (DSAA)}.\hskip 1em plus 0.5em minus 0.4em\relax IEEE, 2023, pp. 1--10.

\bibitem{szegedy2015going}
C.~Szegedy, W.~Liu, Y.~Jia, P.~Sermanet, S.~Reed, D.~Anguelov, D.~Erhan, V.~Vanhoucke, and A.~Rabinovich, ``Going deeper with convolutions,'' in \emph{IEEE conf. comp. vision patt. recognit.}, 2015, pp. 1--9.

\bibitem{szegedy2017inception}
C.~Szegedy, S.~Ioffe, V.~Vanhoucke, and A.~A. Alemi, ``Inception-v4, inception-resnet and the impact of residual connections on learning,'' in \emph{Thirty-first AAAI conference on artificial intelligence}, 2017.

\bibitem{ronald2021isplinception}
M.~Ronald, A.~Poulose, and D.~S. Han, ``isplinception: an inception-resnet deep learning architecture for human activity recognition,'' \emph{IEEE Access}, vol.~9, pp. 68\,985--69\,001, 2021.

\bibitem{ismail2022deep}
A.~Ismail-Fawaz, M.~Devanne, J.~Weber, and G.~Forestier, ``Deep learning for time series classification using new hand-crafted convolution filters,'' in \emph{2022 IEEE International Conference on Big Data (Big Data)}.\hskip 1em plus 0.5em minus 0.4em\relax IEEE, 2022, pp. 972--981.

\bibitem{husken2003recurrent}
M.~H{\"u}sken and P.~Stagge, ``Recurrent neural networks for time series classification,'' \emph{Neurocomputing}, vol.~50, pp. 223--235, 2003.

\bibitem{dennis2019shallow}
D.~Dennis, D.~A.~E. Acar, V.~Mandikal, V.~S. Sadasivan, V.~Saligrama, H.~V. Simhadri, and P.~Jain, ``Shallow rnn: accurate time-series classification on resource constrained devices,'' \emph{Advances neural inf. process. syst.}, vol.~32, 2019.

\bibitem{fernandez2007sequence}
S.~Fern{\'a}ndez, A.~Graves, and J.~Schmidhuber, ``Sequence labelling in structured domains with hierarchical recurrent neural networks,'' in \emph{20th International Joint Conference on Artificial Intelligence, IJCAI 2007}, 2007.

\bibitem{hermans2013training}
M.~Hermans and B.~Schrauwen, ``Training and analysing deep recurrent neural networks,'' \emph{Advances neural inf. process. syst.}, vol.~26, 2013.

\bibitem{pascanu2013difficulty}
R.~Pascanu, T.~Mikolov, and Y.~Bengio, ``On the difficulty of training recurrent neural networks,'' in \emph{Int. conf. mach. learn.}\hskip 1em plus 0.5em minus 0.4em\relax PMLR, 2013, pp. 1310--1318.

\bibitem{hochreiter1997long}
S.~Hochreiter and J.~Schmidhuber, ``Long short-term memory,'' \emph{Neural computation}, vol.~9, no.~8, pp. 1735--1780, 1997.

\bibitem{chung2014empirical}
J.~Chung, C.~Gulcehre, K.~Cho, and Y.~Bengio, ``Empirical evaluation of gated recurrent neural networks on sequence modeling,'' \emph{arXiv preprint:1412.3555}, 2014.

\bibitem{sutskever2014sequence}
I.~Sutskever, O.~Vinyals, and Q.~V. Le, ``Sequence to sequence learning with neural networks,'' \emph{Advances neural inf. process. syst.}, vol.~27, 2014.

\bibitem{donahue2015long}
J.~Donahue, L.~Anne~Hendricks, S.~Guadarrama, M.~Rohrbach, S.~Venugopalan, K.~Saenko, and T.~Darrell, ``Long-term recurrent convolutional networks for visual recognition and description,'' in \emph{IEEE conf. comp. vision patt. recognit.}, 2015, pp. 2625--2634.

\bibitem{karpathy2015deep}
A.~Karpathy and L.~Fei-Fei, ``Deep visual-semantic alignments for generating image descriptions,'' in \emph{IEEE conf. comp. vision patt. recognit.}, 2015, pp. 3128--3137.

\bibitem{tang2016sequence}
Y.~Tang, J.~Xu, K.~Matsumoto, and C.~Ono, ``Sequence-to-sequence model with attention for time series classification,'' in \emph{2016 IEEE 16th Int. Conf. Data Min. Workshops (ICDMW)}.\hskip 1em plus 0.5em minus 0.4em\relax IEEE, 2016, pp. 503--510.

\bibitem{malhotra2017timenet}
P.~Malhotra, V.~TV, L.~Vig, P.~Agarwal, and G.~Shroff, ``Timenet: Pre-trained deep recurrent neural network for time series classification,'' \emph{arXiv preprint:1706.08838}, 2017.

\bibitem{karim2019multivariate}
F.~Karim, S.~Majumdar, H.~Darabi, and S.~Harford, ``{Multivariate LSTM-FCNs for time series classification},'' \emph{Neural Networks}, vol. 116, pp. 237--245, 2019.

\bibitem{zhang2020tapnet}
X.~Zhang, Y.~Gao, J.~Lin, and C.-T. Lu, ``Tapnet: Multivariate time series classification with attentional prototypical network,'' in \emph{AAAI Conference on Artificial Intelligence}, vol.~34, no.~04, 2020, pp. 6845--6852.

\bibitem{zuo2021smate}
J.~Zuo, K.~Zeitouni, and Y.~Taher, ``Smate: Semi-supervised spatio-temporal representation learning on multivariate time series,'' in \emph{2021 IEEE International Conference on Data Mining (ICDM)}.\hskip 1em plus 0.5em minus 0.4em\relax IEEE, 2021, pp. 1565--1570.

\bibitem{karim2017lstm}
F.~Karim, S.~Majumdar, H.~Darabi, and S.~Chen, ``{LSTM fully convolutional networks for time series classification},'' \emph{IEEE access}, vol.~6, pp. 1662--1669, 2017.

\bibitem{lin2017gcrnn}
S.~Lin and G.~C. Runger, ``Gcrnn: Group-constrained convolutional recurrent neural network,'' \emph{IEEE transactions on neural networks and learning systems}, vol.~29, no.~10, pp. 4709--4718, 2017.

\bibitem{mutegeki2020cnn}
R.~Mutegeki and D.~S. Han, ``{A CNN-LSTM approach to human activity recognition},'' in \emph{2020 IEEE Int. Conf. Comput. Intell. Commun. Technol. (ICAIIC)}.\hskip 1em plus 0.5em minus 0.4em\relax IEEE, 2020, pp. 362--366.

\bibitem{pascanu2012understanding}
R.~Pascanu, T.~Mikolov, and Y.~Bengio, ``Understanding the exploding gradient problem,'' \emph{CoRR, abs/1211.5063}, vol.~2, no. 417, p.~1, 2012.

\bibitem{vaswani2017attention}
A.~Vaswani, N.~Shazeer, N.~Parmar, J.~Uszkoreit, L.~Jones, A.~N. Gomez, {\L}.~Kaiser, and I.~Polosukhin, ``Attention is all you need,'' \emph{Advances neural inf. process. syst.}, vol.~30, 2017.

\bibitem{devlin2018bert}
J.~Devlin, M.-W. Chang, K.~Lee, and K.~Toutanova, ``{BERT: Pre-training of deep bidirectional transformers for language understanding},'' in \emph{Proceedings of NAACL-HLT 2019}, vol.~1.\hskip 1em plus 0.5em minus 0.4em\relax Stroudsburg, PA, USA: Association for Computational Linguistics, 2019, pp. 4171--4186.

\bibitem{dosovitskiy2020image}
A.~Dosovitskiy, L.~Beyer, A.~Kolesnikov, D.~Weissenborn, X.~Zhai, T.~Unterthiner, M.~Dehghani, M.~Minderer, G.~Heigold, S.~Gelly \emph{et~al.}, ``{An image is worth 16x16 words: Transformers for image recognition at scale},'' \emph{arXiv preprint:2010.11929}, 2020.

\bibitem{li2019enhancing}
S.~Li, X.~Jin, Y.~Xuan, X.~Zhou, W.~Chen, Y.-X. Wang, and X.~Yan, ``Enhancing the locality and breaking the memory bottleneck of transformer on time series forecasting,'' \emph{Advances neural inf. process. syst.}, vol.~32, 2019.

\bibitem{zhou2021informer}
H.~Zhou, S.~Zhang, J.~Peng, S.~Zhang, J.~Li, H.~Xiong, and W.~Zhang, ``Informer: Beyond efficient transformer for long sequence time-series forecasting,'' in \emph{Proceedings of AAAI}, 2021.

\bibitem{kostas2021bendr}
D.~Kostas, S.~Aroca-Ouellette, and F.~Rudzicz, ``Bendr: using transformers and a contrastive self-supervised learning task to learn from massive amounts of eeg data,'' \emph{Frontiers in Human Neuroscience}, vol.~15, 2021.

\bibitem{yuan2018muvan}
Y.~Yuan, G.~Xun, F.~Ma, Y.~Wang, N.~Du, K.~Jia, L.~Su, and A.~Zhang, ``Muvan: A multi-view attention network for multivariate temporal data,'' in \emph{2018 IEEE International Conference on Data Mining (ICDM)}.\hskip 1em plus 0.5em minus 0.4em\relax IEEE, 2018, pp. 717--726.

\bibitem{hsieh2021explainable}
T.-Y. Hsieh, S.~Wang, Y.~Sun, and V.~Honavar, ``Explainable multivariate time series classification: A deep neural network which learns to attend to important variables as well as time intervals,'' in \emph{14th ACM International Conference on Web Search and Data Mining}, 2021, pp. 607--615.

\bibitem{chen2021multi}
W.~Chen and K.~Shi, ``Multi-scale attention convolutional neural network for time series classification,'' \emph{Neural Networks}, vol. 136, pp. 126--140, 2021.

\bibitem{yuan2018novel}
Y.~Yuan, G.~Xun, F.~Ma, Q.~Suo, H.~Xue, K.~Jia, and A.~Zhang, ``A novel channel-aware attention framework for multi-channel eeg seizure detection via multi-view deep learning,'' in \emph{2018 IEEE EMBS International Conference on Biomedical \& Health Informatics (BHI)}.\hskip 1em plus 0.5em minus 0.4em\relax IEEE, 2018, pp. 206--209.

\bibitem{liang2018geoman}
Y.~Liang, S.~Ke, J.~Zhang, X.~Yi, and Y.~Zheng, ``Geoman: Multi-level attention networks for geo-sensory time series prediction,'' in \emph{IJCAI}, vol. 2018, 2018, pp. 3428--3434.

\bibitem{hu2020multistage}
J.~Hu and W.~Zheng, ``Multistage attention network for multivariate time series prediction,'' \emph{Neurocomputing}, vol. 383, pp. 122--137, 2020.

\bibitem{cheng2020novel}
X.~Cheng, P.~Han, G.~Li, S.~Chen, and H.~Zhang, ``A novel channel and temporal-wise attention in convolutional networks for multivariate time series classification,'' \emph{IEEE Access}, vol.~8, pp. 212\,247--212\,257, 2020.

\bibitem{xiao2021rtfn}
Z.~Xiao, X.~Xu, H.~Xing, S.~Luo, P.~Dai, and D.~Zhan, ``Rtfn: a robust temporal feature network for time series classification,'' \emph{Inf. Sci.}, vol. 571, pp. 65--86, 2021.

\bibitem{wang2023wavelet}
J.~Wang, C.~Yang, X.~Jiang, and J.~Wu, ``When: A wavelet-dtw hybrid attention network for heterogeneous time series analysis,'' in \emph{Proceedings of the 29th ACM SIGKDD Conference on Knowledge Discovery and Data Mining}, 2023, pp. 2361--2373.

\bibitem{jaderberg2015spatial}
M.~Jaderberg, K.~Simonyan, A.~Zisserman \emph{et~al.}, ``Spatial transformer networks,'' \emph{Advances neural inf. process. syst.}, vol.~28, 2015.

\bibitem{woo2018cbam}
S.~Woo, J.~Park, J.-Y. Lee, and I.~S. Kweon, ``Cbam: Convolutional block attention module,'' in \emph{European conference on computer vision}, 2018, pp. 3--19.

\bibitem{hu2018squeeze}
J.~Hu, L.~Shen, and G.~Sun, ``Squeeze-and-excitation networks,'' in \emph{IEEE conf. comp. vision patt. recognit.}, 2018, pp. 7132--7141.

\bibitem{wang2021time}
T.~Wang, Z.~Liu, T.~Zhang, and Y.~Li, ``Time series classification based on multi-scale dynamic convolutional features and distance features,'' in \emph{2021 2nd Asia Symposium on Signal Processing (ASSP)}.\hskip 1em plus 0.5em minus 0.4em\relax IEEE, 2021, pp. 239--246.

\bibitem{song2018attend}
H.~Song, D.~Rajan, J.~Thiagarajan, and A.~Spanias, ``Attend and diagnose: Clinical time series analysis using attention models,'' in \emph{AAAI conference on artificial intelligence}, vol.~32, no.~1, 2018.

\bibitem{jin2021end}
C.-c. Jin and X.~Chen, ``An end-to-end framework combining time--frequency expert knowledge and modified transformer networks for vibration signal classification,'' \emph{Expert Systems with Applications}, vol. 171, p. 114570, 2021.

\bibitem{Rasmussen2004}
C.~E. Rasmussen, \emph{Gaussian Processes in Machine Learning}.\hskip 1em plus 0.5em minus 0.4em\relax Berlin, Heidelberg: Springer Berlin Heidelberg, 2004, pp. 63--71.

\bibitem{allam2021paying}
T.~Allam~Jr and J.~D. McEwen, ``Paying attention to astronomical transients: Photometric classification with the time-series transformer,'' \emph{arXiv preprint:2105.06178}, 2021.

\bibitem{liu2021gated}
M.~Liu, S.~Ren, S.~Ma, J.~Jiao, Y.~Chen, Z.~Wang, and W.~Song, ``Gated transformer networks for multivariate time series classification,'' \emph{arXiv preprint:2103.14438}, 2021.

\bibitem{zhao2022rethinking}
B.~Zhao, H.~Xing, X.~Wang, F.~Song, and Z.~Xiao, ``Rethinking attention mechanism in time series classification,'' \emph{arXiv preprint:2207.07564}, 2022.

\bibitem{ren2022autotransformer}
Y.~Ren, L.~Li, X.~Yang, and J.~Zhou, ``Autotransformer: Automatic transformer architecture design for time series classification,'' in \emph{Pacific-Asia Conference on Knowledge Discovery and Data Mining}.\hskip 1em plus 0.5em minus 0.4em\relax Springer, 2022, pp. 143--155.

\bibitem{Jin2023graph}
\BIBentryALTinterwordspacing
M.~Jin, H.~Y. Koh, Q.~Wen, D.~Zambon, C.~Alippi, G.~I. Webb, I.~King, and S.~Pan, ``{A Survey on Graph Neural Networks for Time Series: Forecasting, Classification, Imputation, and Anomaly Detection},'' \emph{arXiv}, vol.~14, no.~8, pp. 1--27, jul 2023. [Online]. Available: \url{http://arxiv.org/abs/2307.03759}
\BIBentrySTDinterwordspacing

\bibitem{Wu2021graph}
Z.~Wu, S.~Pan, F.~Chen, G.~Long, C.~Zhang, and P.~S. Yu, ``{A Comprehensive Survey on Graph Neural Networks},'' \emph{IEEE Transactions on Neural Networks and Learning Systems}, vol.~32, no.~1, pp. 4--24, jan 2021.

\bibitem{Scarselli2009graph}
F.~Scarselli, M.~Gori, A.~C. Tsoi, M.~Hagenbuchner, and G.~Monfardini, ``{The Graph Neural Network Model},'' \emph{IEEE Transactions on Neural Networks}, vol.~20, no.~1, pp. 61--80, jan 2009.

\bibitem{Xi2023graph}
W.~Xi, A.~Jain, L.~Zhang, and J.~Lin, ``{LB-SimTSC: An Efficient Similarity-Aware Graph Neural Network for Semi-Supervised Time Series Classification},'' \emph{arXiv}, jan 2023.

\bibitem{Liu2023graph}
H.~Liu, X.~Liu, D.~Yang, Z.~Liang, H.~Wang, Y.~Cui, and J.~Gu, ``{TodyNet: Temporal Dynamic Graph Neural Network for Multivariate Time Series Classification},'' \emph{arXiv}, vol.~XX, no.~Xx, pp. 1--10, apr 2023.

\bibitem{Bloemheuvel2023graph}
S.~Bloemheuvel, J.~van~den Hoogen, D.~Jozinovi{\'{c}}, A.~Michelini, and M.~Atzmueller, ``{Graph neural networks for multivariate time series regression with application to seismic data},'' \emph{International Journal of Data Science and Analytics}, vol.~16, no.~3, pp. 317--332, sep 2023.

\bibitem{Cheng2021graph}
Z.~Cheng, Y.~Yang, S.~Jiang, W.~Hu, Z.~Ying, Z.~Chai, and C.~Wang, ``{Time2Graph+: Bridging Time Series and Graph Representation Learning via Multiple Attentions},'' \emph{IEEE Transactions on Knowledge and Data Engineering}, vol.~35, no.~2, pp. 1--1, 2021.

\bibitem{covert2019temporal}
I.~C. Covert, B.~Krishnan, I.~Najm, J.~Zhan, M.~Shore, J.~Hixson, and M.~J. Po, ``Temporal graph convolutional networks for automatic seizure detection,'' in \emph{Machine Learning for Healthcare Conference}.\hskip 1em plus 0.5em minus 0.4em\relax PMLR, 2019, pp. 160--180.

\bibitem{Song2020graph}
T.~Song, W.~Zheng, P.~Song, and Z.~Cui, ``{EEG Emotion Recognition Using Dynamical Graph Convolutional Neural Networks},'' \emph{IEEE Transactions on Affective Computing}, vol.~11, no.~3, pp. 532--541, jul 2020.

\bibitem{jia2020graphsleepnet}
Z.~Jia, Y.~Lin, J.~Wang, R.~Zhou, X.~Ning, Y.~He, and Y.~Zhao, ``Graphsleepnet: Adaptive spatial-temporal graph convolutional networks for sleep stage classification.'' in \emph{IJCAI}, 2020, pp. 1324--1330.

\bibitem{ma2021deep}
Z.~Ma, G.~Mei, E.~Prezioso, Z.~Zhang, and N.~Xu, ``A deep learning approach using graph convolutional networks for slope deformation prediction based on time-series displacement data,'' \emph{Neural Computing and Applications}, vol.~33, no.~21, pp. 14\,441--14\,457, 2021.

\bibitem{li2020multireceptive}
T.~Li, Z.~Zhao, C.~Sun, R.~Yan, and X.~Chen, ``Multireceptive field graph convolutional networks for machine fault diagnosis,'' \emph{IEEE Transactions on Industrial Electronics}, vol.~68, no.~12, pp. 12\,739--12\,749, 2020.

\bibitem{nhu2021graph}
D.~Nhu, M.~Janmohamed, P.~Perucca, A.~Gilligan, P.~Kwan, T.~O'Brien, C.~Tan, and L.~Kuhlmann, ``Graph convolutional network for generalized epileptiform abnormality detection on eeg,'' in \emph{2021 IEEE Signal Processing in Medicine and Biology Symposium (SPMB)}.\hskip 1em plus 0.5em minus 0.4em\relax IEEE, 2021, pp. 1--6.

\bibitem{Tang2021graph}
S.~Tang, J.~A. Dunnmon, K.~Saab, X.~Zhang, Q.~Huang, F.~Dubost, D.~L. Rubin, and C.~Lee-Messer, ``{Self-Supervised Graph Neural Networks for Improved Electroencephalographic Seizure Analysis},'' \emph{ICLR 2022 - 10th Int. Conf. Learning Representations}, pp. 1--23, apr 2021.

\bibitem{Zhang2021graph}
X.~Zhang, M.~Zeman, T.~Tsiligkaridis, and M.~Zitnik, ``{Graph-Guided Network for Irregularly Sampled Multivariate Time Series},'' \emph{ICLR 2022 - 10th International Conference on Learning Representations}, pp. 1--21, oct 2021.

\bibitem{Censi2021graph}
A.~M. Censi, D.~Ienco, Y.~J.~E. Gbodjo, R.~G. Pensa, R.~Interdonato, and R.~Gaetano, ``{Attentive spatial temporal graph CNN for land cover mapping from multi temporal remote sensing data},'' \emph{IEEE Access}, vol.~9, pp. 23\,070--23\,082, 2021.

\bibitem{Azevedo2022graph}
T.~Azevedo, A.~Campbell, R.~Romero-Garcia, L.~Passamonti, R.~A. Bethlehem, P.~Li{\`{o}}, and N.~Toschi, ``{A deep graph neural network architecture for modelling spatio-temporal dynamics in resting-state functional MRI data},'' \emph{Medical Image Analysis}, vol.~79, p. 102471, jul 2022.

\bibitem{Duan2022graph}
Z.~Duan, H.~Xu, Y.~Wang, Y.~Huang, A.~Ren, Z.~Xu, Y.~Sun, and W.~Wang, ``{Multivariate time-series classification with hierarchical variational graph pooling},'' \emph{Neural Networks}, vol. 154, pp. 481--490, oct 2022.

\bibitem{Zha2022graph}
D.~Zha, K.-h. Lai, K.~Zhou, and X.~Hu, ``{Towards Similarity-Aware Time-Series Classification},'' in \emph{Proceedings of the 2022 SIAM International Conference on Data Mining (SDM)}, Philadelphia, PA, jan 2022, pp. 199--207.

\bibitem{Tulczyjew2022graph}
L.~Tulczyjew, M.~Kawulok, N.~Longepe, B.~{Le Saux}, and J.~Nalepa, ``{Graph Neural Networks Extract High-Resolution Cultivated Land Maps From Sentinel-2 Image Series},'' \emph{IEEE Geoscience and Remote Sensing Letters}, vol.~19, pp. 1--5, 2022.

\bibitem{Sun2023graph}
L.~Sun, C.~Li, B.~Liu, and Y.~Zhang, ``{Class-driven Graph Attention Network for Multi-label Time Series Classification in Mobile Health Digital Twins},'' \emph{IEEE Journal on Selected Areas in Communications}, vol.~41, no.~10, pp. 3267--3278, 2023.

\bibitem{Dufourg2023graph}
C.~Dufourg, C.~Pelletier, S.~May, and S.~Lef{\`{e}}vre, ``{Graph Dynamic Earth Net: Spatio-Temporal Graph Benchmark for Satellite Image Time Series},'' in \emph{IGARSS 2023 - 2023 IEEE International Geoscience and Remote Sensing Symposium}.\hskip 1em plus 0.5em minus 0.4em\relax IEEE, jul 2023, pp. 7164--7167.

\bibitem{Keogh2005exact}
E.~Keogh and C.~A. Ratanamahatana, ``{Exact indexing of dynamic time warping},'' \emph{Knowl. Inform. Systems}, vol.~7, no.~3, pp. 358--386, 2005.

\bibitem{Kipf2016graph}
T.~N. Kipf and M.~Welling, ``{Semi-Supervised Classification with Graph Convolutional Networks},'' \emph{5th International Conference on Learning Representations, ICLR 2017 - Conference Track Proceedings}, pp. 1--14, sep 2016.

\bibitem{yang2022unsupervised}
L.~Yang and S.~Hong, ``Unsupervised time-series representation learning with iterative bilinear temporal-spectral fusion,'' in \emph{ICML}, 2022, pp. 25\,038--25\,054.

\bibitem{hyvarinen2016unsupervised}
A.~Hyvarinen and H.~Morioka, ``Unsupervised feature extraction by time-contrastive learning and nonlinear ica,'' \emph{Advances in neural information processing systems}, vol.~29, 2016.

\bibitem{franceschi2019unsupervised}
J.-Y. Franceschi, A.~Dieuleveut, and M.~Jaggi, ``Unsupervised scalable representation learning for multivariate time series,'' \emph{NeurIPS}, vol.~32, 2019.

\bibitem{tonekaboni2021unsupervised}
S.~Tonekaboni, D.~Eytan, and A.~Goldenberg, ``Unsupervised representation learning for time series with temporal neighborhood coding,'' \emph{arXiv preprint arXiv:2106.00750}, 2021.

\bibitem{wickstrom2022mixing}
K.~Wickstr{\o}m, M.~Kampffmeyer, K.~{\O}. Mikalsen, and R.~Jenssen, ``Mixing up contrastive learning: Self-supervised representation learning for time series,'' \emph{Pattern Recognition Letters}, vol. 155, pp. 54--61, 2022.

\bibitem{yang2022timeclr}
X.~Yang, Z.~Zhang, and R.~Cui, ``Timeclr: A self-supervised contrastive learning framework for univariate time series representation,'' \emph{Knowledge-Based Systems}, vol. 245, p. 108606, 2022.

\bibitem{zhang2022self}
X.~Zhang, Z.~Zhao, T.~Tsiligkaridis, and M.~Zitnik, ``Self-supervised contrastive pre-training for time series via time-frequency consistency,'' in \emph{Proceedings of Neural Information Processing Systems, NeurIPS}, 2022.

\bibitem{meng2023mhccl}
Q.~Meng, H.~Qian, Y.~Liu, L.~Cui, Y.~Xu, and Z.~Shen, ``Mhccl: masked hierarchical cluster-wise contrastive learning for multivariate time series,'' in \emph{Proceedings of the AAAI Conference on Artificial Intelligence}, vol.~37, no.~8, 2023, pp. 9153--9161.

\bibitem{chowdhury2022tarnet}
R.~R. Chowdhury, X.~Zhang, J.~Shang, R.~K. Gupta, and D.~Hong, ``Tarnet: Task-aware reconstruction for time-series transformer,'' in \emph{28th ACM SIGKDD Conference on Knowledge Discovery and Data Mining, Washington, DC, USA}, 2022, pp. 14--18.

\bibitem{cheng2023timemae}
M.~Cheng, Q.~Liu, Z.~Liu, H.~Zhang, R.~Zhang, and E.~Chen, ``Timemae: Self-supervised representations of time series with decoupled masked autoencoders,'' \emph{arXiv preprint arXiv:2303.00320}, 2023.

\bibitem{zhang2023self}
W.~Zhang, L.~Yang, S.~Geng, and S.~Hong, ``Self-supervised time series representation learning via cross reconstruction transformer,'' \emph{IEEE Transactions on Neural Networks and Learning Systems}, 2023.

\bibitem{ismail2023finding}
A.~Ismail-Fawaz, M.~Devanne, S.~Berretti, J.~Weber, and G.~Forestier, ``Finding foundation models for time series classification with a pretext task,'' \emph{arXiv preprint arXiv:2311.14534}, 2023.

\bibitem{shorten2019survey}
C.~Shorten and T.~M. Khoshgoftaar, ``A survey on image data augmentation for deep learning,'' \emph{Journal of big data}, vol.~6, no.~1, pp. 1--48, 2019.

\bibitem{um2017data}
T.~T. Um, F.~M. Pfister, D.~Pichler, S.~Endo, M.~Lang, S.~Hirche, U.~Fietzek, and D.~Kuli{\'c}, ``Data augmentation of wearable sensor data for parkinson’s disease monitoring using convolutional neural networks,'' in \emph{Proc. 19th ACM int. conf. multimodal interaction}, 2017, pp. 216--220.

\bibitem{rashid2019window}
K.~M. Rashid and J.~Louis, ``Window-warping: a time series data augmentation of imu data for construction equipment activity identification,'' in \emph{ISARC. Proceedings of the International Symposium on Automation and Robotics in Construction}, vol.~36.\hskip 1em plus 0.5em minus 0.4em\relax IAARC Publications, 2019, pp. 651--657.

\bibitem{iwana2021time}
B.~K. Iwana and S.~Uchida, ``Time series data augmentation for neural networks by time warping with a discriminative teacher,'' in \emph{2020 25th International Conference on Pattern Recognition (ICPR)}.\hskip 1em plus 0.5em minus 0.4em\relax IEEE, 2021, pp. 3558--3565.

\bibitem{nguyen2020improving}
T.-S. Nguyen, S.~Stueker, J.~Niehues, and A.~Waibel, ``Improving sequence-to-sequence speech recognition training with on-the-fly data augmentation,'' in \emph{ICASSP 2020-2020 IEEE International Conference on Acoustics, Speech and Signal Processing (ICASSP)}.\hskip 1em plus 0.5em minus 0.4em\relax IEEE, 2020, pp. 7689--7693.

\bibitem{vachhani2018data}
B.~Vachhani, C.~Bhat, and S.~K. Kopparapu, ``Data augmentation using healthy speech for dysarthric speech recognition.'' in \emph{Interspeech}, 2018, pp. 471--475.

\bibitem{gao2020robusttad}
J.~Gao, X.~Song, Q.~Wen, P.~Wang, L.~Sun, and H.~Xu, ``Robusttad: Robust time series anomaly detection via decomposition and convolutional neural networks,'' 2020.

\bibitem{cui2016multiscale}
Z.~Cui, W.~Chen, and Y.~Chen, ``Multi-scale convolutional neural networks for time series classification,'' 2016.

\bibitem{leguennec2016data}
A.~Le~Guennec, S.~Malinowski, and R.~Tavenard, ``{Data Augmentation for Time Series Classification using Convolutional Neural Networks},'' in \emph{{ECML/PKDD on Advanced Analytics and Learning on Temporal Data}}, 2016.

\bibitem{forestier2017generating}
G.~Forestier, F.~Petitjean, H.~A. Dau, G.~I. Webb, and E.~Keogh, ``Generating synthetic time series to augment sparse datasets,'' in \emph{2017 IEEE international conference on data mining (ICDM)}.\hskip 1em plus 0.5em minus 0.4em\relax IEEE, 2017, pp. 865--870.

\bibitem{fawaz2018data}
H.~I. Fawaz, G.~Forestier, J.~Weber, L.~Idoumghar, and P.-A. Muller, ``Data augmentation using synthetic data for time series classification with deep residual networks,'' 2018.

\bibitem{terefe2020time}
T.~Terefe, M.~Devanne, J.~Weber, D.~Hailemariam, and G.~Forestier, ``Time series averaging using multi-tasking autoencoder,'' in \emph{2020 IEEE 32nd International Conference on Tools with Artificial Intelligence (ICTAI)}.\hskip 1em plus 0.5em minus 0.4em\relax IEEE, 2020, pp. 1065--1072.

\bibitem{iwana2021empirical}
B.~K. Iwana and S.~Uchida, ``An empirical survey of data augmentation for time series classification with neural networks,'' \emph{Plos one}, vol.~16, no.~7, p. e0254841, 2021.

\bibitem{pialla2022data}
G.~Pialla, M.~Devanne, J.~Weber, L.~Idoumghar, and G.~Forestier, ``Data augmentation for time series classification with deep learning models,'' in \emph{International Workshop on Advanced Analytics and Learning on Temporal Data}.\hskip 1em plus 0.5em minus 0.4em\relax Springer, 2022, pp. 117--132.

\bibitem{gao2023data}
Z.~Gao, L.~Li, and T.~Xu, ``Data augmentation for time-series classification: An extensive empirical study and comprehensive survey,'' \emph{arXiv preprint arXiv:2310.10060}, 2023.

\bibitem{deng2009imagenet}
J.~Deng, W.~Dong, R.~Socher, L.-J. Li, K.~Li, and L.~Fei-Fei, ``Imagenet: A large-scale hierarchical image database,'' in \emph{2009 IEEE conf. comp. vision patt. recognit.}\hskip 1em plus 0.5em minus 0.4em\relax Ieee, 2009, pp. 248--255.

\bibitem{fawaz2018transfer}
H.~I. Fawaz, G.~Forestier, J.~Weber, L.~Idoumghar, and P.-A. Muller, ``Transfer learning for time series classification,'' in \emph{2018 IEEE international conference on big data (Big Data)}.\hskip 1em plus 0.5em minus 0.4em\relax IEEE, 2018, pp. 1367--1376.

\bibitem{spiegel2016transfer}
S.~Spiegel, ``Transfer learning for time series classification in dissimilarity spaces,'' \emph{Proceedings of AALTD}, vol.~78, 2016.

\bibitem{li2020deep}
F.~Li, K.~Shirahama, M.~A. Nisar, X.~Huang, and M.~Grzegorzek, ``Deep transfer learning for time series data based on sensor modality classification,'' \emph{Sensors}, vol.~20, no.~15, p. 4271, 2020.

\bibitem{rotem2022transfer}
Y.~Rotem, N.~Shimoni, L.~Rokach, and B.~Shapira, ``Transfer learning for time series classification using synthetic data generation,'' in \emph{International Symposium on Cyber Security, Cryptology, and Machine Learning}.\hskip 1em plus 0.5em minus 0.4em\relax Springer, 2022, pp. 232--246.

\bibitem{senanayaka2022similarity}
A.~Senanayaka, A.~Al~Mamun, G.~Bond, W.~Tian, H.~Wang, S.~Fuller, T.~Falls, S.~Rahimi, and L.~Bian, ``Similarity-based multi-source transfer learning approach for time series classification,'' \emph{International Journal of Prognostics and Health Management}, vol.~13, no.~2, 2022.

\bibitem{kashiparekh2019convtimenet}
K.~Kashiparekh, J.~Narwariya, P.~Malhotra, L.~Vig, and G.~Shroff, ``Convtimenet: A pre-trained deep convolutional neural network for time series classification,'' in \emph{2019 International Joint Conference on Neural Networks (IJCNN)}.\hskip 1em plus 0.5em minus 0.4em\relax IEEE, 2019, pp. 1--8.

\bibitem{MerlinPraveena2022app}
D.~{Merlin Praveena}, D.~{Angelin Sarah}, and S.~{Thomas George}, ``{Deep Learning Techniques for EEG Signal Applications–A Review},'' \emph{IETE Journal of Research}, vol.~68, no.~4, pp. 3030--3037, 2022.

\bibitem{Liu2021app}
X.~Liu, H.~Wang, Z.~Li, and L.~Qin, ``{Deep learning in ECG diagnosis: A review},'' \emph{Knowledge-Based Systems}, vol. 227, p. 107187, 2021.

\bibitem{Zaini2022app}
N.~Zaini, L.~W. Ean, A.~N. Ahmed, and M.~A. Malek, ``{A systematic literature review of deep learning neural network for time series air quality forecasting},'' \emph{Environmental Science and Pollution Research}, vol.~29, no.~4, pp. 4958--4990, jan 2022.

\bibitem{Zhang2022app}
B.~Zhang, Y.~Rong, R.~Yong, D.~Qin, M.~Li, G.~Zou, and J.~Pan, ``{Deep learning for air pollutant concentration prediction: A review},'' \emph{Atmospheric Environment}, vol. 290, p. 119347, dec 2022.

\bibitem{Toh2020app}
G.~Toh and J.~Park, ``{Review of Vibration-Based Structural Health Monitoring Using Deep Learning},'' \emph{Appl. Sci.}, vol.~10, no.~5, p. 1680, 2020.

\bibitem{Thoppil2021app}
N.~M. Thoppil, V.~Vasu, and C.~S.~P. Rao, ``{Deep Learning Algorithms for Machinery Health Prognostics Using Time-Series Data: A Review},'' \emph{Journal of Vibration Engineering \& Technologies}, vol.~9, no.~6, pp. 1123--1145, sep 2021.

\bibitem{Ren2023app}
L.~Ren, Z.~Jia, Y.~Laili, and D.~Huang, ``{Deep Learning for Time-Series Prediction in IIoT: Progress, Challenges, and Prospects},'' \emph{IEEE Transactions on Neural Networks and Learning Systems}, vol.~PP, pp. 1--20, 2023.

\bibitem{Himeur2021app}
Y.~Himeur, K.~Ghanem, A.~Alsalemi, F.~Bensaali, and A.~Amira, ``{Artificial intelligence based anomaly detection of energy consumption in buildings: A review, current trends and new perspectives},'' \emph{Applied Energy}, vol. 287, p. 116601, 2021.

\bibitem{Stowell2022app}
D.~Stowell, ``{Computational bioacoustics with deep learning: a review and roadmap},'' \emph{PeerJ}, vol.~10, p. e13152, mar 2022.

\bibitem{Gupta2022survey}
N.~Gupta, S.~K. Gupta, R.~K. Pathak, V.~Jain, P.~Rashidi, and J.~S. Suri, ``{Human activity recognition in artificial intelligence framework: a narrative review},'' \emph{Artificial Intelligence Review}, vol.~55, no.~6, pp. 4755--4808, aug 2022.

\bibitem{Ramanujam2021survey}
E.~Ramanujam, T.~Perumal, and S.~Padmavathi, ``{Human activity recognition with smartphone and wearable sensors using deep learning techniques: A review},'' \emph{IEEE Sensors Journal}, vol.~21, no.~12, pp. 13\,029--13\,040, jun 2021.

\bibitem{Lockhart2012appl}
J.~W. Lockhart, T.~Pulickal, and G.~M. Weiss, ``{Applications of mobile activity recognition},'' in \emph{2012 ACM Conference on Ubiquitous Computing - UbiComp '12}.\hskip 1em plus 0.5em minus 0.4em\relax New York, New York, USA: ACM Press, 2012, p. 1054.

\bibitem{Tapia2004appl}
E.~M. Tapia, S.~S. Intille, and K.~Larson, ``{Activity recognition in the home using simple and ubiquitous sensors},'' in \emph{Lecture Notes in Computer Science}.\hskip 1em plus 0.5em minus 0.4em\relax Berlin, Heidelberg: Springer, 2004, vol. 3001, pp. 158--175.

\bibitem{Kong2022vision}
Y.~Kong and Y.~Fu, ``{Human action recognition and prediction: A survey},'' \emph{International Journal of Computer Vision}, vol. 130, no.~5, pp. 1366--1401, may 2022.

\bibitem{Zhang2019vision}
H.-B. Zhang, Y.-X. Zhang, B.~Zhong, Q.~Lei, L.~Yang, J.-X. Du, and D.-S. Chen, ``{A comprehensive survey of vision-based human action recognition methods},'' \emph{Sensors}, vol.~19, no.~5, p. 1005, feb 2019.

\bibitem{Ordonez2016deep}
F.~Ord{\'{o}}{\~{n}}ez and D.~Roggen, ``{Deep convolutional and LSTM recurrent neural networks for multimodal wearable activity recognition},'' \emph{Sensors}, vol.~16, no.~1, p. 115, jan 2016.

\bibitem{Reiss2012pamap}
A.~Reiss and D.~Stricker, ``{Introducing a new benchmarked dataset for activity monitoring},'' in \emph{16th Int. Symp. Wearable Computers}, 2012, pp. 108--109.

\bibitem{Zhang2012uschad}
M.~Zhang and A.~A. Sawchuk, ``{USC-HAD: A daily activity dataset for ubiquitous activity recognition using wearable sensors},'' in \emph{2012 ACM Conference on Ubiquitous Computing - UbiComp '12}.\hskip 1em plus 0.5em minus 0.4em\relax New York, New York, USA: ACM Press, 2012, p. 1036.

\bibitem{Roggen2010opp}
D.~Roggen, A.~Calatroni, M.~Rossi, T.~Holleczek, K.~F{\"o}rster, G.~Tr{\"o}ster, P.~Lukowicz, D.~Bannach, G.~Pirkl \emph{et~al.}, ``Collecting complex activity datasets in highly rich networked sensor environments,'' in \emph{Seventh international conference on networked sensing systems}.\hskip 1em plus 0.5em minus 0.4em\relax IEEE, 2010, pp. 233--240.

\bibitem{Sztyler2017position}
T.~Sztyler, H.~Stuckenschmidt, and W.~Petrich, ``{Position-aware activity recognition with wearable devices},'' \emph{Pervasive and Mobile Computing}, vol.~38, pp. 281--295, jul 2017.

\bibitem{Lara2013survey}
O.~D. Lara and M.~A. Labrador, ``{A survey on human activity recognition using wearable sensors},'' \emph{IEEE Communications Surveys {\&} Tutorials}, vol.~15, no.~3, pp. 1192--1209, 2013.

\bibitem{Gu2022survey}
F.~Gu, M.-H. Chung, M.~Chignell, S.~Valaee, B.~Zhou, and X.~Liu, ``{A survey on deep learning for human activity recognition},'' \emph{ACM Computing Surveys}, vol.~54, no.~8, pp. 1--34, nov 2022.

\bibitem{Hammerla2016deep}
N.~Y. Hammerla, S.~Halloran, and T.~Ploetz, ``{Deep, convolutional, and recurrent Models for human activity recognition using wearables},'' \emph{IJCAI International Joint Conference on Artificial Intelligence}, vol. 2016-Janua, pp. 1533--1540, apr 2016.

\bibitem{Zeng2014cnn}
M.~Zeng, L.~T. Nguyen, B.~Yu, O.~J. Mengshoel, J.~Zhu, P.~Wu, and J.~Zhang, ``{Convolutional neural networks for human activity recognition using mobile sensors},'' in \emph{6th International Conference on Mobile Computing, Applications and Services}.\hskip 1em plus 0.5em minus 0.4em\relax ICST, 2014, pp. 718--737.

\bibitem{Jiang2015cnn}
W.~Jiang and Z.~Yin, ``{Human activity recognition using wearable sensors by deep convolutional neural Networks},'' in \emph{23rd ACM international conference on Multimedia}.\hskip 1em plus 0.5em minus 0.4em\relax New York, NY, USA: ACM, oct 2015, pp. 1307--1310.

\bibitem{Yang2015cnn}
J.~B. Yang, M.~N. Nguyen, P.~P. San, X.~L. Li, and S.~Krishnaswamy, ``{Deep convolutional neural networks on multichannel time series for human activity recognition},'' \emph{IJCAI International Joint Conference on Artificial Intelligence}, vol. 2015-Janua, pp. 3995--4001, 2015.

\bibitem{Ronao2016har}
C.~A. Ronao and S.-B. Cho, ``{Human activity recognition with smartphone sensors using deep learning neural networks},'' \emph{Expert Systems with Applications}, vol.~59, pp. 235--244, oct 2016.

\bibitem{Guan2017ensemble}
Y.~Guan and T.~Pl{\"{o}}tz, ``{Ensembles of deep LSTM learners for activity recognition using wearables},'' \emph{ACM on Interactive, Mobile, Wearable and Ubiquitous Technologies}, vol.~1, no.~2, pp. 1--28, jun 2017.

\bibitem{Lee2017har}
S.-M. Lee, S.~M. Yoon, and H.~Cho, ``{Human activity recognition from accelerometer data using Convolutional Neural Network},'' in \emph{2017 IEEE International Conference on Big Data and Smart Computing (BigComp)}, vol.~83.\hskip 1em plus 0.5em minus 0.4em\relax IEEE, feb 2017, pp. 131--134.

\bibitem{Murad2017deep}
A.~Murad and J.-Y. Pyun, ``{Deep recurrent neural networks for human activity recognition},'' \emph{Sensors}, vol.~17, no.~11, p. 2556, nov 2017.

\bibitem{Ignatov2018realtime}
A.~Ignatov, ``{Real-time human activity recognition from accelerometer data using Convolutional Neural Networks},'' \emph{Applied Soft Computing}, vol.~62, pp. 915--922, jan 2018.

\bibitem{Rueda2018cnn}
F.~{Moya Rueda}, R.~Grzeszick, G.~Fink, S.~Feldhorst, and M.~ten Hompel, ``{Convolutional neural networks for human activity recognition using body-worn sensors},'' \emph{Informatics}, vol.~5, no.~2, p.~26, may 2018.

\bibitem{Yao2018cnn}
R.~Yao, G.~Lin, Q.~Shi, and D.~C. Ranasinghe, ``{Efficient dense labelling of human activity sequences from wearables using fully convolutional networks},'' \emph{Pattern Recognition}, vol.~78, pp. 252--266, jun 2018.

\bibitem{Zeng2018rnn}
M.~Zeng, H.~Gao, T.~Yu, O.~J. Mengshoel, H.~Langseth, I.~Lane, and X.~Liu, ``{Understanding and improving recurrent networks for human activity recognition by continuous attention},'' in \emph{ACM International Symposium on Wearable Computers}, New York, NY, USA, 2018, pp. 56--63.

\bibitem{Ma2019attnsense}
H.~Ma, W.~Li, X.~Zhang, S.~Gao, and S.~Lu, ``{AttnSense: Multi-level attention mechanism for multimodal human activity recognition},'' in \emph{Twenty-Eighth International Joint Conference on Artificial Intelligence}, California, 2019, pp. 3109--3115.

\bibitem{Xu2019innohar}
C.~Xu, D.~Chai, J.~He, X.~Zhang, and S.~Duan, ``{InnoHAR: A deep neural network for complex human activity recognition},'' \emph{IEEE Access}, vol.~7, pp. 9893--9902, 2019.

\bibitem{Zhang2020novel}
H.~Zhang, Z.~Xiao, J.~Wang, F.~Li, and E.~Szczerbicki, ``{A novel IoT-perceptive human activity recognition (HAR) approach using multihead convolutional attention},'' \emph{IEEE Internet of Things Journal}, vol.~7, no.~2, pp. 1072--1080, feb 2020.

\bibitem{Challa2021multi}
S.~K. Challa, A.~Kumar, and V.~B. Semwal, ``{A multibranch CNN-BiLSTM model for human activity recognition using wearable sensor data},'' \emph{The Visual Computer}, no. 0123456789, aug 2021.

\bibitem{Mekruksavanich2021deep}
S.~Mekruksavanich and A.~Jitpattanakul, ``{Deep Convolutional Neural Network with RNNs for complex activity recognition using wrist-worn wearable sensor data},'' \emph{Electronics}, vol.~10, no.~14, p. 1685, jul 2021.

\bibitem{Chen2021har}
L.~Chen, X.~Liu, L.~Peng, and M.~Wu, ``{Deep learning based multimodal complex human activity recognition using wearable devices},'' \emph{Applied Intelligence}, vol.~51, no.~6, pp. 4029--4042, jun 2021.

\bibitem{Mekruksavanich2021lstm}
S.~Mekruksavanich and A.~Jitpattanakul, ``{LSTM networks using smartphone data for sensor-based human activity recognition in smart homes},'' \emph{Sensors}, vol.~21, no.~5, p. 1636, feb 2021.

\bibitem{Mekruksavanich2021biometric}
------, ``{Biometric user identification based on human activity recognition using wearable sensors: An experiment using deep learning models},'' \emph{Electronics}, vol.~10, no.~3, p. 308, jan 2021.

\bibitem{Nafea2021sensor}
O.~Nafea, W.~Abdul, G.~Muhammad, and M.~Alsulaiman, ``{Sensor-based human activity recognition with spatio-temporal deep learning},'' \emph{Sensors}, vol.~21, no.~6, p. 2141, mar 2021.

\bibitem{Singh2021convlstm}
S.~P. Singh, M.~K. Sharma, A.~Lay-Ekuakille, D.~Gangwar, and S.~Gupta, ``{Deep ConvLSTM with self-attention for human activity decoding using wearable sensors},'' \emph{IEEE Sensors Journal}, vol.~21, no.~6, pp. 8575--8582, mar 2021.

\bibitem{Wang2022deep}
X.~Wang, L.~Zhang, W.~Huang, S.~Wang, H.~Wu, J.~He, and A.~Song, ``{Deep convolutional networks with tunable speed–accuracy tradeoff for human activity recognition using wearables},'' \emph{IEEE Transactions on Instrumentation and Measurement}, vol.~71, pp. 1--12, 2022.

\bibitem{Xu2022deform}
S.~Xu, L.~Zhang, W.~Huang, H.~Wu, and A.~Song, ``{Deformable convolutional networks for multimodal human activity recognition using wearable sensors},'' \emph{IEEE Transactions on Instrumentation and Measurement}, vol.~71, pp. 1--14, 2022.

\bibitem{Dai2017deform}
J.~Dai, H.~Qi, Y.~Xiong, Y.~Li, G.~Zhang, H.~Hu, and Y.~Wei, ``{Deformable convolutional networks},'' in \emph{2017 IEEE Int. Conf. Computer Vision (ICCV)}, 2017, pp. 764--773.

\bibitem{Wulder2008landsat}
M.~A. Wulder, J.~C. White, S.~N. Goward, J.~G. Masek, J.~R. Irons, M.~Herold, W.~B. Cohen, T.~R. Loveland, and C.~E. Woodcock, ``{Landsat continuity: Issues and opportunities for land cover monitoring},'' \emph{Remote Sensing of Environment}, vol. 112, no.~3, pp. 955--969, mar 2008.

\bibitem{Emery2017emr}
W.~Emery and A.~Camps, ``{Basic electromagnetic concepts and applications to optical sensors},'' in \emph{Introduction to Satellite Remote Sensing}, W.~Emery and A.~Camps, Eds.\hskip 1em plus 0.5em minus 0.4em\relax Elsevier, jan 2017, ch.~2, pp. 43--83.

\bibitem{Gorelick2017gee}
N.~Gorelick, M.~Hancher, M.~Dixon, S.~Ilyushchenko, D.~Thau, and R.~Moore, ``{Google Earth Engine: Planetary-scale geospatial analysis for everyone},'' \emph{Remote Sensing of Environment}, vol. 202, pp. 18--27, dec 2017.

\bibitem{Giuliani2017cube}
G.~Giuliani, B.~Chatenoux, A.~{De Bono}, D.~Rodila, J.-P. Richard, K.~Allenbach, H.~Dao, and P.~Peduzzi, ``{Building an Earth observations data cube: lessons learned from the Swiss data cube (SDC) on generating analysis ready data (ARD)},'' \emph{Big Earth Data}, vol.~1, no. 1-2, pp. 100--117, dec 2017.

\bibitem{Lewis2017cube}
A.~Lewis, S.~Oliver, L.~Lymburner, B.~Evans, L.~Wyborn, N.~Mueller, G.~Raevksi, J.~Hooke, R.~Woodcock, J.~Sixsmith \emph{et~al.}, ``The australian geoscience data cube—foundations and lessons learned,'' \emph{Remote Sensing of Environment}, vol. 202, pp. 276--292, 2017.

\bibitem{Ienco2020tassel}
D.~Ienco, Y.~J.~E. Gbodjo, R.~Interdonato, and R.~Gaetano, ``{Attentive weakly supervised land cover mapping for object-based satellite image time series data with spatial interpretation},'' \emph{arXiv}, pp. 1--12, 2020.

\bibitem{Garnot2020tae}
V.~{Sainte Fare Garnot}, L.~Landrieu, S.~Giordano, and N.~Chehata, ``{Satellite image time Series classification With Pixel-Set encoders and temporal self-attention},'' in \emph{2020 IEEE/CVF Conference on Computer Vision and Pattern Recognition (CVPR)}.\hskip 1em plus 0.5em minus 0.4em\relax IEEE, jun 2020, pp. 12\,322--12\,331.

\bibitem{Kulshrestha2022hole}
A.~Kulshrestha, L.~Chang, and A.~Stein, ``{Use of LSTM for sinkhole-related anomaly detection and classification of InSAR deformation time series},'' \emph{IEEE Journal of Selected Topics in Applied Earth Observations and Remote Sensing}, vol.~15, pp. 4559--4570, 2022.

\bibitem{Ban2020fire}
Y.~Ban, P.~Zhang, A.~Nascetti, A.~R. Bevington, and M.~A. Wulder, ``{Near real-time wildfire progression monitoring with Sentinel-1 SAR time series and deep learning},'' \emph{Scientific Reports}, vol.~10, no.~1, p. 1322, dec 2020.

\bibitem{Rambour2020flood}
C.~Rambour, N.~Audebert, E.~Koeniguer, B.~{Le Saux}, M.~Crucianu, and M.~Datcu, ``{Flood detection in time series of optical and SAR images},'' \emph{Int. Archives Photogrammetry, Remote Sens. \& Spatial Inf. Sci.}, vol. XLIII-B2-2, no.~B2, pp. 1343--1346, aug 2020.

\bibitem{KamdemDeTeyou2020road}
G.~Kamdem De~Teyou, Y.~Tarabalka, I.~Manighetti, R.~Almar, and S.~Tripodi, ``Deep neural networks for automatic extraction of features in time series optical satellite images,'' \emph{Int. Archives Photogrammetry, Remote Sens. \& Spatial Inf. Sci.}, vol.~43, 2020.

\bibitem{Matosak2022forest}
B.~M. Matosak, L.~M.~G. Fonseca, E.~C. Taquary, R.~V. Maretto, H.~D.~N. Bendini, and M.~Adami, ``{Mapping deforestation in Cerrado based on hybrid deep learning architecture and medium spatial resolution satellite time series},'' \emph{Remote Sensing}, vol.~14, no.~1, pp. 1--22, 2022.

\bibitem{Minh2017vege}
D.~{Ho Tong Minh}, D.~Ienco, R.~Gaetano, N.~Lalande, E.~Ndikumana, F.~Osman, and P.~Maurel, ``{Deep recurrent neural networks for winter vegetation quality mapping via multitemporal SAR Sentinel-1},'' \emph{IEEE Geoscience and Remote Sensing Letters}, vol.~15, no.~3, pp. 464--468, mar 2018.

\bibitem{Labenski2022under}
P.~Labenski, M.~Ewald, S.~Schmidtlein, and F.~E. Fassnacht, ``{Classifying surface fuel types based on forest stand photographs and satellite time series using deep learning},'' \emph{International Journal of Applied Earth Observation and Geoinformation}, vol. 109, p. 102799, may 2022.

\bibitem{Rao2020lfmc}
K.~Rao, A.~P. Williams, J.~F. Flefil, and A.~G. Konings, ``{SAR-enhanced mapping of live fuel moisture content},'' \emph{Remote Sens. Environ.}, vol. 245, p. 111797, 2020.

\bibitem{Zhu2020lfmc}
L.~Zhu, G.~I. Webb, M.~Yebra, G.~Scortechini, L.~Miller, and F.~Petitjean, ``{Live fuel moisture content estimation from MODIS: A deep learning approach},'' \emph{ISPRS J. Photogramm. Remote Sens.}, vol. 179, pp. 81--91, sep 2021.

\bibitem{Miller2022lfmc}
L.~Miller, L.~Zhu, M.~Yebra, C.~R{\"{u}}diger, and G.~I. Webb, ``{Multi-modal temporal CNNs for live fuel moisture content estimation},'' \emph{Environmental Modelling {\&} Software}, vol. 156, p. 105467, oct 2022.

\bibitem{Xie2022lfmc}
J.~Xie, T.~Qi, W.~Hu, H.~Huang, B.~Chen, and J.~Zhang, ``{Retrieval of live fuel moisture content based on multi-source remote sensing data and ensemble deep learning model},'' \emph{Remote Sensing}, vol.~14, no.~17, p. 4378, sep 2022.

\bibitem{Lahssini2022wood}
K.~Lahssini, F.~Teste, K.~R. Dayal, S.~Durrieu, D.~Ienco, and J.-M. Monnet, ``{Combining LiDAR metrics and Sentinel-2 imagery to estimate basal area and wood volume in complex forest environment via neural networks},'' \emph{IEEE J. Selected Topics Applied Earth Obs. Remote Sens.}, vol.~15, pp. 4337--4348, 2022.

\bibitem{Sun2020yield}
J.~Sun, Z.~Lai, L.~Di, Z.~Sun, J.~Tao, and Y.~Shen, ``{Multilevel deep learning network for county-level corn yield estimation in the U.S. Corn Belt},'' \emph{IEEE Journal of Selected Topics in Applied Earth Observations and Remote Sensing}, vol.~13, pp. 5048--5060, 2020.

\bibitem{Li2019tan}
Z.~Li, G.~Chen, and T.~Zhang, ``{Temporal attention networks for multitemporal multisensor crop classification},'' \emph{IEEE Access}, vol.~7, pp. 134\,677--134\,690, 2019.

\bibitem{Li2020tga}
Z.~Li, G.~Zhou, and Q.~Song, ``{A temporal group attention approach for multitemporal multisensor crop classification},'' \emph{Infrared Physics and Technology}, vol. 105, p. 103152, 2020.

\bibitem{Ji20183dcnn}
S.~Ji, C.~Zhang, A.~Xu, Y.~Shi, and Y.~Duan, ``{3D convolutional neural networks for crop classification with multi-temporal remote sensing images},'' \emph{Remote Sensing}, vol.~10, no.~2, p.~75, jan 2018.

\bibitem{Xu2020dcm}
J.~Xu, Y.~Zhu, R.~Zhong, Z.~Lin, J.~Xu, H.~Jiang, J.~Huang, H.~Li, and T.~Lin, ``{DeepCropMapping: A multi-temporal deep learning approach with improved spatial generalizability for dynamic corn and soybean mapping},'' \emph{Remote Sensing of Environment}, vol. 247, p. 111946, sep 2020.

\bibitem{Barriere2022lstm}
V.~Barriere and M.~Claverie, ``Multimodal crop type classification fusing multi-spectral satellite time series with farmers crop rotations and local crop distribution,'' \emph{arXiv preprint:2208.10838}, 2022.

\bibitem{Garnot2020ltae}
V.~S.~F. Garnot and L.~Landrieu, ``{Lightweight temporal self-attention for classifying satellite images time series},'' in \emph{Lecture Notes in Computer Science}.\hskip 1em plus 0.5em minus 0.4em\relax Springer International Publishing, 2020, vol. 12588 LNAI, pp. 171--181.

\bibitem{Ofori-Ampofo2021att}
S.~Ofori-Ampofo, C.~Pelletier, and S.~Lang, ``{Crop type mapping from optical and radar time series using attention-based deep learning},'' \emph{Remote Sensing}, vol.~13, no.~22, p. 4668, nov 2021.

\bibitem{Yuan2021sitsbert}
Y.~Yuan and L.~Lin, ``{Self-Supervised pretraining of transformers for satellite image time series classification},'' \emph{IEEE Journal of Selected Topics in Applied Earth Observations and Remote Sensing}, vol.~14, pp. 474--487, 2021.

\bibitem{DiMauro2017tiselc}
N.~Di~Mauro, A.~Vergari, T.~M.~A. Basile, F.~G. Ventola, and F.~Esposito, ``End-to-end learning of deep spatio-temporal representations for satellite image time series classification.'' in \emph{DC@ PKDD/ECML}, 2017.

\bibitem{Kussul2017cnn}
N.~Kussul, M.~Lavreniuk, S.~Skakun, and A.~Shelestov, ``{Deep learning classification of land cover and crop types using remote sensing data},'' \emph{IEEE Geoscience and Remote Sensing Letters}, vol.~14, no.~5, pp. 778--782, may 2017.

\bibitem{Pelletier2019tempcnn}
C.~Pelletier, G.~Webb, and F.~Petitjean, ``{Temporal convolutional neural network for the classification of satellite image time series},'' \emph{Remote Sensing}, vol.~11, no.~5, p. 523, mar 2019.

\bibitem{Dou2021tsi}
P.~Dou, H.~Shen, Z.~Li, and X.~Guan, ``{Time series remote sensing image classification framework using combination of deep learning and multiple classifiers system},'' \emph{International Journal of Applied Earth Observation and Geoinformation}, vol. 103, p. 102477, 2021.

\bibitem{Ienco2019twinns}
D.~Ienco, R.~Interdonato, R.~Gaetano, and D.~{Ho Tong Minh}, ``{Combining Sentinel-1 and Sentinel-2 satellite image time series for land cover mapping via a multi-source deep learning architecture},'' \emph{ISPRS J. Photogramm. Remote Sens.}, vol. 158, pp. 11--22, 2019.

\bibitem{Interdonato2019duplo}
R.~Interdonato, D.~Ienco, R.~Gaetano, and K.~Ose, ``{DuPLO: A DUal view Point deep Learning architecture for time series classificatiOn},'' \emph{ISPRS J. Photogramm. Remote Sens.}, vol. 149, pp. 91--104, mar 2019.

\bibitem{Russwurm2018seqrnn}
M.~Ru{\ss}wurm and M.~K{\"{o}}rner, ``{Multi-Temporal land cover classification with sequential recurrent encoders},'' \emph{ISPRS International Journal of Geo-Information}, vol.~7, no.~4, p. 129, mar 2018.

\bibitem{Stoian2019fgunet}
A.~Stoian, V.~Poulain, J.~Inglada, V.~Poughon, and D.~Derksen, ``{Land cover maps production with high resolution satellite image time series and convolutional neural networks: Adaptations and limits for operational systems},'' \emph{Remote Sensing}, vol.~11, no.~17, pp. 1--26, 2019.

\bibitem{Ienco2017rnn}
D.~Ienco, R.~Gaetano, C.~Dupaquier, and P.~Maurel, ``{Land cover classification via multitemporal spatial data by deep recurrent neural networks},'' \emph{IEEE Geoscience and Remote Sensing Letters}, vol.~14, no.~10, pp. 1685--1689, oct 2017.

\bibitem{Gbodjo2020hob2srnn}
Y.~J.~E. Gbodjo, D.~Ienco, L.~Leroux, R.~Interdonato, R.~Gaetano, and B.~Ndao, ``{Object-based multi-temporal and multi-source land cover mapping leveraging hierarchical class relationships},'' \emph{Remote Sensing}, vol.~12, no.~17, p. 2814, aug 2020.

\bibitem{Ienco2019od2rnn}
D.~Ienco, R.~Gaetano, R.~Interdonato, K.~Ose, and D.~{Ho Tong Minh}, ``{Combining Sentinel-1 and Sentinel-2 time series via RNN for object-based land cover classification},'' in \emph{IGARSS 2019 - 2019 IEEE International Geoscience and Remote Sensing Symposium}.\hskip 1em plus 0.5em minus 0.4em\relax IEEE, jul 2019, pp. 4881--4884.

\bibitem{Yuan2022sitsformer}
Y.~Yuan, L.~Lin, Q.~Liu, R.~Hang, and Z.-G. Zhou, ``{SITS-Former: A pre-trained spatio-spectral-temporal representation model for Sentinel-2 time series classification},'' \emph{International Journal of Applied Earth Observation and Geoinformation}, vol. 106, p. 102651, feb 2022.

\bibitem{Qiao2021yield}
M.~Qiao, X.~He, X.~Cheng, P.~Li, H.~Luo, L.~Zhang, and Z.~Tian, ``Crop yield prediction from multi-spectral, multi-temporal remotely sensed imagery using recurrent 3d convolutional neural networks,'' \emph{International Journal of Applied Earth Observation and Geoinformation}, vol. 102, p. 102436, 2021.

\bibitem{Russwurm2020att}
M.~Ru{\ss}wurm and M.~K{\"{o}}rner, ``{Self-attention for raw optical satellite time series classification},'' \emph{ISPRS J. Photogramm. Remote Sens.}, vol. 169, pp. 421--435, 2020.

\bibitem{Tuia2016da}
D.~Tuia, C.~Persello, and L.~Bruzzone, ``{Domain adaptation for the classification of remote sensing data: An overview of recent advances},'' \emph{IEEE Geoscience and Remote Sensing Magazine}, vol.~4, no.~2, pp. 41--57, 2016.

\bibitem{Garnot2019time}
V.~S.~F. Garnot, L.~Landrieu, S.~Giordano, and N.~Chehata, ``Time-space tradeoff in deep learning models for crop classification on satellite multi-spectral image time series,'' in \emph{IGARSS 2019-2019 IEEE International Geoscience and Remote Sensing Symposium}.\hskip 1em plus 0.5em minus 0.4em\relax IEEE, 2019, pp. 6247--6250.

\bibitem{Fawaz2019ensembles}
H.~{Ismail Fawaz}, G.~Forestier, J.~Weber, L.~Idoumghar, and P.-A. Muller, ``{Deep neural network ensembles for time series classification},'' in \emph{2019 International Joint Conference on Neural Networks (IJCNN)}, vol. 2019-July.\hskip 1em plus 0.5em minus 0.4em\relax IEEE, jul 2019, pp. 1--6.

\bibitem{Wolpert1992stack}
D.~H. Wolpert, ``{Stacked generalization},'' \emph{Neural Networks}, vol.~5, no.~2, pp. 241--259, jan 1992.

\bibitem{Freund1996ada}
Y.~Freund and R.~E. Schapire, ``{Experiments with a new boosting algorithm},'' in \emph{13th Int. conf. mach. learn.}, 1996, pp. 148--156.

\bibitem{Gomez2016review}
C.~G{\'{o}}mez, J.~C. White, and M.~A. Wulder, ``{Optical remotely sensed time series data for land cover classification: A review},'' \emph{ISPRS J. Photogramm. Remote Sens.}, vol. 116, pp. 55--72, 2016.

\bibitem{Zhu2017review}
X.~X. Zhu, D.~Tuia, L.~Mou, G.~S. Xia, L.~Zhang, F.~Xu, and F.~Fraundorfer, ``{Deep learning in remote sensing: A comprehensive review and list of resources},'' \emph{IEEE Geoscience and Remote Sensing Magazine}, vol.~5, no.~4, pp. 8--36, 2017.

\bibitem{Ma2019review}
L.~Ma, Y.~Liu, X.~Zhang, Y.~Ye, G.~Yin, and B.~A. Johnson, ``{Deep learning in remote sensing applications: A meta-analysis and review},'' \emph{ISPRS J. Photogramm. Remote Sens.}, vol. 152, pp. 166--177, jun 2019.

\bibitem{Yuan2020review}
Q.~Yuan, H.~Shen, T.~Li, Z.~Li, S.~Li, Y.~Jiang, H.~Xu, W.~Tan, Q.~Yang, J.~Wang, J.~Gao, and L.~Zhang, ``{Deep learning in environmental remote sensing: Achievements and challenges},'' \emph{Remote Sensing of Environment}, vol. 241, p. 111716, may 2020.

\bibitem{Chaves2020review}
M.~{E. D. Chaves}, M.~{C. A. Picoli}, and I.~{D. Sanches}, ``{Recent applications of Landsat 8/OLI and Sentinel-2/MSI for land use and land cover mapping: A systematic review},'' \emph{Remote Sensing}, vol.~12, no.~18, p. 3062, sep 2020.

\bibitem{Moskolai2021apps}
W.~R. Moskola{\"{i}}, W.~Abdou, A.~Dipanda, and Kolyang, ``{Application of deep learning architectures for satellite image time series prediction: A review},'' \emph{Remote Sensing}, vol.~13, no.~23, p. 4822, nov 2021.

\bibitem{lines2015time}
J.~Lines and A.~Bagnall, ``Time series classification with ensembles of elastic distance measures,'' \emph{Data Min. Knowl. Discov.}, vol.~29, no.~3, pp. 565--592, 2015.

\bibitem{tan2020fastee}
C.~W. Tan, F.~Petitjean, and G.~I. Webb, ``{FastEE: Fast Ensembles of Elastic Distances for time series classification},'' \emph{Data Min. Knowl. Discov.}, vol.~34, no.~1, pp. 231--272, 2020.

\bibitem{herrmann2021amercing}
M.~Herrmann and G.~I. Webb, ``Amercing: An intuitive, elegant and effective constraint for dynamic time warping,'' \emph{arXiv preprint:2111.13314}, 2021.

\bibitem{bagnall2020usage}
A.~Bagnall, M.~Flynn, J.~Large, J.~Lines, and M.~Middlehurst, ``{On the usage and performance of the Hierarchical Vote Collective of Transformation-based Ensembles version 1.0 (hive-cote v1. 0)},'' in \emph{International Workshop on Advanced Analytics and Learning on Temporal Data}, 2020, pp. 3--18.

\bibitem{middlehurst2021hive}
M.~Middlehurst, J.~Large, M.~Flynn, J.~Lines, A.~Bostrom, and A.~Bagnall, ``{HIVE-COTE 2.0: a new meta ensemble for time series classification},'' \emph{Machine Learning}, vol. 110, no.~11, pp. 3211--3243, 2021.

\bibitem{bagnall2015time}
A.~Bagnall, J.~Lines, J.~Hills, and A.~Bostrom, ``{Time-series classification with COTE: the collective of transformation-based ensembles},'' \emph{IEEE Transactions on Knowledge and Data Engineering}, vol.~27, no.~9, pp. 2522--2535, 2015.

\bibitem{lines2018time}
J.~Lines, S.~Taylor, and A.~Bagnall, ``{Time series classification with HIVE-COTE: The hierarchical vote collective of transformation-based ensembles},'' \emph{ACM Transactions on Knowledge Discovery from Data}, vol.~12, no.~5, 2018.

\bibitem{lines2016hive}
------, ``{Hive-Cote: The hierarchical vote collective of transformation-based ensembles for time series classification},'' in \emph{2016 IEEE 16th international conference on data mining (ICDM)}.\hskip 1em plus 0.5em minus 0.4em\relax IEEE, 2016, pp. 1041--1046.

\bibitem{kate2016using}
R.~J. Kate, ``Using dynamic time warping distances as features for improved time series classification,'' \emph{Data Min. Knowl. Discov.}, vol.~30, no.~2, pp. 283--312, 2016.

\bibitem{bostrom2015binary}
A.~Bostrom and A.~Bagnall, ``Binary shapelet transform for multiclass time series classification,'' in \emph{Int. conf. big data analytics .knowl. disco.}\hskip 1em plus 0.5em minus 0.4em\relax Springer, 2015, pp. 257--269.

\bibitem{schafer2015boss}
P.~Sch{\"a}fer, ``The boss is concerned with time series classification in the presence of noise,'' \emph{Data Min. Knowl. Discov.}, vol.~29, no.~6, pp. 1505--1530, 2015.

\bibitem{hills2014classification}
J.~Hills, J.~Lines, E.~Baranauskas, J.~Mapp, and A.~Bagnall, ``Classification of time series by shapelet transformation,'' \emph{Data Min. Knowl. Discov.}, vol.~28, no.~4, pp. 851--881, 2014.

\bibitem{deng2013time}
H.~Deng, G.~Runger, E.~Tuv, and M.~Vladimir, ``A time series forest for classification and feature extraction,'' \emph{Inf. Sci.}, vol. 239, pp. 142--153, 2013.

\bibitem{baydogan2013bag}
M.~G. Baydogan, G.~Runger, and E.~Tuv, ``A bag-of-features framework to classify time series,'' \emph{IEEE transactions on pattern analysis and machine intelligence}, vol.~35, no.~11, pp. 2796--2802, 2013.

\bibitem{dempster2021minirocket}
A.~Dempster, D.~F. Schmidt, and G.~I. Webb, ``{Minirocket: A very fast (almost) deterministic transform for time series classification},'' in \emph{27th ACM SIGKDD Conference on Knowledge Discovery \& Data Mining}, 2021, pp. 248--257.

\bibitem{tan2021multirocket}
C.~W. Tan, A.~Dempster, C.~Bergmeir, and G.~I. Webb, ``{MultiRocket: multiple pooling operators and transformations for fast and effective time series classification},'' \emph{Data Min. Knowl. Discov.}, jun 2022.

\bibitem{dempster2023hydra}
A.~Dempster, D.~F. Schmidt, and G.~I. Webb, ``Hydra: Competing convolutional kernels for fast and accurate time series classification,'' \emph{Data Mining and Knowledge Discovery}, pp. 1--27, 2023.

\bibitem{lucas2019proximity}
B.~Lucas, A.~Shifaz, C.~Pelletier, L.~O’Neill, N.~Zaidi, B.~Goethals, F.~Petitjean, and G.~I. Webb, ``Proximity forest: an effective and scalable distance-based classifier for time series,'' \emph{Data Mining and Knowledge Discovery}, vol.~33, no.~3, pp. 607--635, 2019.

\bibitem{herrmann2023proximity}
M.~Herrmann, C.~W. Tan, M.~Salehi, and G.~I. Webb, ``Proximity forest 2.0: A new effective and scalable similarity-based classifier for time series,'' \emph{arXiv preprint arXiv:2304.05800}, 2023.

\bibitem{fukushima1982neocognitron}
K.~Fukushima and S.~Miyake, ``Neocognitron: A self-organizing neural network model for a mechanism of visual pattern recognition,'' in \emph{Competition and cooperation in neural nets}.\hskip 1em plus 0.5em minus 0.4em\relax Springer, 1982, pp. 267--285.

\bibitem{hubel1962receptive}
D.~H. Hubel and T.~N. Wiesel, ``Receptive fields, binocular interaction and functional architecture in the cat's visual cortex,'' \emph{The Journal of physiology}, vol. 160, no.~1, p. 106, 1962.

\bibitem{lecun1998gradient}
Y.~LeCun, L.~Bottou, Y.~Bengio, and P.~Haffner, ``Gradient-based learning applied to document recognition,'' \emph{Proc. IEEE}, vol.~86, no.~11, pp. 2278--2324, 1998.

\bibitem{nair2010rectified}
V.~Nair and G.~E. Hinton, ``Rectified linear units improve restricted boltzmann machines,'' in \emph{Icml}, 2010.

\bibitem{hihi1995hierarchical}
S.~Hihi and Y.~Bengio, ``Hierarchical recurrent neural networks for long-term dependencies,'' \emph{Advances neural inf. process. syst.}, vol.~8, 1995.

\bibitem{pascanu2013construct}
R.~Pascanu, C.~Gulcehre, K.~Cho, and Y.~Bengio, ``How to construct deep recurrent neural networks,'' \emph{arXiv preprint:1312.6026}, 2013.

\bibitem{kawakami2008supervised}
K.~Kawakami, ``Supervised sequence labelling with recurrent neural networks,'' Ph.D. dissertation, Technical University of Munich, 2008.

\bibitem{bahdanau2014neural}
D.~Bahdanau, K.~Cho, and Y.~Bengio, ``Neural machine translation by jointly learning to align and translate,'' \emph{arXiv preprint:1409.0473}, 2014.

\bibitem{cho2014learning}
K.~Cho, B.~Van~Merri{\"e}nboer, C.~Gulcehre, D.~Bahdanau, F.~Bougares, H.~Schwenk, and Y.~Bengio, ``Learning phrase representations using rnn encoder-decoder for statistical machine translation,'' \emph{arXiv preprint:1406.1078}, 2014.

\bibitem{luong2015effective}
M.-T. Luong, H.~Pham, and C.~D. Manning, ``Effective approaches to attention-based neural machine translation,'' \emph{arXiv preprint:1508.04025}, 2015.

\bibitem{Bruna2013graph}
\BIBentryALTinterwordspacing
J.~Bruna, W.~Zaremba, A.~Szlam, and Y.~LeCun, ``{Spectral Networks and Locally Connected Networks on Graphs},'' pp. 1--14, 2013. [Online]. Available: \url{http://arxiv.org/abs/1312.6203}
\BIBentrySTDinterwordspacing

\bibitem{Shuman2013graph}
\BIBentryALTinterwordspacing
D.~I. Shuman, S.~K. Narang, P.~Frossard, A.~Ortega, and P.~Vandergheynst, ``{The emerging field of signal processing on graphs: Extending high-dimensional data analysis to networks and other irregular domains},'' \emph{IEEE Signal Processing Magazine}, vol.~30, no.~3, pp. 83--98, 2013. [Online]. Available: \url{http://ieeexplore.ieee.org/document/6494675/}
\BIBentrySTDinterwordspacing

\bibitem{Longa2023graph}
\BIBentryALTinterwordspacing
A.~Longa, V.~Lachi, G.~Santin, M.~Bianchini, B.~Lepri, P.~Lio, F.~Scarselli, and A.~Passerini, ``{Graph Neural Networks for temporal graphs: State of the art, open challenges, and opportunities},'' 2023. [Online]. Available: \url{http://arxiv.org/abs/2302.01018}
\BIBentrySTDinterwordspacing

\bibitem{Bachlin2009daphnet}
M.~Bachlin, D.~Roggen, G.~Troster, M.~Plotnik, N.~Inbar, I.~Meidan, T.~Herman, M.~Brozgol, E.~Shaviv, N.~Giladi, and J.~M. Hausdorff, ``{Potentials of enhanced context awareness in wearable assistants for Parkinson's Disease patients with the freezing of gait syndrome},'' in \emph{2009 International Symposium on Wearable Computers}.\hskip 1em plus 0.5em minus 0.4em\relax IEEE, sep 2009, pp. 123--130.

\bibitem{Micucci2017unimib}
D.~Micucci, M.~Mobilio, and P.~Napoletano, ``{UniMiB SHAR: A dataset for human activity recognition using acceleration data from smartphones},'' \emph{Applied Sciences}, vol.~7, no.~10, p. 1101, oct 2017.

\bibitem{Zappi2008skoda}
P.~Zappi, C.~Lombriser, T.~Stiefmeier, E.~Farella, D.~Roggen, L.~Benini, and G.~Tr{\"o}ster, ``Activity recognition from on-body sensors: accuracy-power trade-off by dynamic sensor selection,'' in \emph{European Conference on Wireless Sensor Networks}.\hskip 1em plus 0.5em minus 0.4em\relax Springer, 2008, pp. 17--33.

\bibitem{Chavarriaga2013opp}
R.~Chavarriaga, H.~Sagha, A.~Calatroni, S.~T. Digumarti, G.~Tr{\"{o}}ster, J.~D.~R. Mill{\'{a}}n, and D.~Roggen, ``{The Opportunity challenge: A benchmark database for on-body sensor-based activity recognition},'' \emph{Pattern Recognition Letters}, vol.~34, no.~15, pp. 2033--2042, nov 2013.

\bibitem{Reiss2012pamap2}
A.~Reiss and D.~Stricker, ``{Creating and benchmarking a new dataset for physical activity monitoring},'' in \emph{5th Int. Conf. PErvasive Technologies Related to Assistive Environments - PETRA '12}.\hskip 1em plus 0.5em minus 0.4em\relax New York, New York, USA: ACM Press, 2012, p.~1.

\bibitem{Anguita2013ucihar}
D.~Anguita, A.~Ghio, L.~Oneto, X.~Parra, and J.~L. Reyes-Ortiz, ``{A public domain dataset for human activity recognition using smartphones},'' in \emph{21th European Symposium on Artificial Neural Networks, Computational Intelligence and Machine Learning, ESANN}, Bruges, Belgium, 2013, pp. 437--442.

\bibitem{Kwapisz2011wisdm}
J.~R. Kwapisz, G.~M. Weiss, and S.~A. Moore, ``{Activity recognition using cell phone accelerometers},'' \emph{ACM SIGKDD Explorations Newsletter}, vol.~12, no.~2, pp. 74--82, mar 2011.

\bibitem{USGALandsat}
\BIBentryALTinterwordspacing
{U.S. Geological Survey}, ``{Landsat Satellite Missions}.'' [Online]. Available: \url{https://www.usgs.gov/landsat-missions/landsat-satellite-missions}
\BIBentrySTDinterwordspacing

\bibitem{NASAMODIS}
\BIBentryALTinterwordspacing
NASA, ``{MODIS Moderate Resolution Imaging Spectrometer}.'' [Online]. Available: \url{https://modis.gsfc.nasa.gov/}
\BIBentrySTDinterwordspacing

\bibitem{ESA2019Sentinel}
\BIBentryALTinterwordspacing
{European Space Agency}, ``{Sentinel Online},'' 2019. [Online]. Available: \url{https://sentinel.esa.int/web/sentinel/home}
\BIBentrySTDinterwordspacing

\bibitem{ESAPleiades}
\BIBentryALTinterwordspacing
------, ``{Pleiades - Earth Online}.'' [Online]. Available: \url{https://earth.esa.int/eogateway/missions/pleiades}
\BIBentrySTDinterwordspacing

\bibitem{NSPO2020Formosat}
\BIBentryALTinterwordspacing
{National Space Organization}, ``{FORMOSAT-2},'' 2020. [Online]. Available: \url{https://www.nspo.narl.org.tw/history{\_}prog.php?c=20030402{\&}ln=en}
\BIBentrySTDinterwordspacing

\bibitem{EoPortal2014Gaofen1}
\BIBentryALTinterwordspacing
EoPortal, ``{Gaofen-1},'' 2014. [Online]. Available: \url{https://www.eoportal.org/satellite-missions/gaofen-1}
\BIBentrySTDinterwordspacing

\bibitem{EoPortal2015Gaofen2}
\BIBentryALTinterwordspacing
------, ``{Gaofen-2},'' 2015. [Online]. Available: \url{https://www.eoportal.org/satellite-missions/gaofen-2}
\BIBentrySTDinterwordspacing

\end{thebibliography}

\newpage

\appendix
\section*{Appendix}
\section{Non-Deep Learning Time Series Classification} \label{App:non-deep}
In this section, we aim to give a brief introduction to the field of TSC and discuss its current status.
We refer interested readers to the `bake-off' papers \cite{bagnall2017great,middlehurst2023bake, ruiz2020great} that describes TSC methods in much details and benchmark them. 

Research in TSC started with distance-based approaches that find discriminating patterns in the shape of the time series. Distance-based approaches usually consist of coupling a 1-nearest neighbour (1NN) classifier with a time series distance measure~\cite{lines2015time,tan2020fastee}.
Small distortions in the time series can lead to false matches when measuring the distance between time series using standard distance measurements such as Euclidean distance~\cite{lines2015time}. 
A time series distance measure aims to compensate for these distortions by aligning two time series such that the alignment cost between the two are minimised.
There are many time series distances proposed in the literature; among these, the Dynamic Time Warping ($DTW$) distance is one of the most popular choices for many time series tasks, due to its intuitiveness and effectiveness in aligning two time series.
The 1NN-$DTW$ has been the go-to method for TSC for decades.
However, by comparing several time series distance measures, the work in~\cite{lines2015time}  showed that as of 2015, there was no single distance that significantly outperformed $DTW$ when used with a 1NN classifier. 
The recent Amerced $DTW$~\cite{herrmann2021amercing} distance is the first distance that is significantly more accurate than $DTW$. 
These individual 1NN classifiers with different distances can be ensembled together to create an ensemble, such as the Ensemble of Elastic distances (EE), that significantly outperforms each of them individually~\cite{tan2020fastee,lines2015time}. However, since most distances have a complexity of $O(L^2)$ where $L$ is the length of the series, performing a nearest neighbour search becomes very costly. Hence, distance-based approaches are considered to be one of the slowest methods for TSC~\cite{bagnall2020usage,middlehurst2021hive}.

As a result of EE, recent studies have focused mainly on developing ensembling methods that significantly outperform 1NN-$DTW$~\cite{tan2020fastee,bagnall2020usage,middlehurst2021hive,bagnall2015time,lines2018time,lines2016hive,kate2016using,bostrom2015binary,schafer2015boss,hills2014classification,deng2013time,baydogan2013bag}.
These approaches use either an ensemble of tree-based approaches~\cite{baydogan2013bag,deng2013time} or an ensemble of different types of discriminant classifiers, such as NN with several distances and Support Vector Machine (SVM) on one or several feature spaces~\cite{bagnall2015time,bostrom2015binary,schafer2015boss,kate2016using}. 
All these approaches share a common property -- the data transformation phase where the time series is transformed into a new feature space such as the shapelets transform~\cite{bostrom2015binary} or DTW features~\cite{kate2016using}. 
Taking advantage of this notion led to the development of the Hierarchical Vote Collective of Transformation-based Ensembles (HIVE-COTE)~\cite{lines2016hive,middlehurst2021hive}. HIVE-COTE is a meta ensemble for TSC and forms its ensemble from ensemble classifiers of multiple domains. Since its introduction in 2016~\cite{lines2016hive}, HIVE-COTE has gone through a few iterations. 
Recently, the latest HIVE-COTE version, HIVE-COTEv2.0 (HC2) was proposed~\cite{middlehurst2021hive}. 
It is comprised of 4 ensemble members, each of them being the then state of the art in their respective domains. 
It is currently one of the most accurate classifiers for both univariate and multivariate TSC tasks~\cite{middlehurst2021hive}. 
Despite being accurate on 26 multivariate and 142 univariate TSC benchmark datasets, that are relatively small, HC2 scales poorly on large datasets with long time series as well as datasets with large numbers of channels.

Various work has been done on speeding up TSC methods without sacrificing accuracy \cite{dempster2019rocket,dempster2021minirocket,tan2020fastee,tan2021multirocket,dempster2023hydra,lucas2019proximity,herrmann2023proximity}.
A recent breakthrough is the development of Rocket \cite{dempster2019rocket} that was able to process 109 univariate time series datasets under 4 hours while the previous fastest took days. 
Rocket leverages large number of random convolutional filters to extract features from each series that might be relevant to classifying a 
series. 
These features are then passed to a linear model for classification. 
Rocket has been improved to be faster (Minirocket \cite{dempster2021minirocket}) and more accurate (Multirocket \cite{tan2021multirocket} and Hydra \cite{dempster2023hydra}).
Hydra when combined with Multirocket is now one of the fastest and most accurate method for TSC. 

\section{DNN Architectures for Time Series} \label{APx:DNN}

In this section, we provide a descriptive overview of deep learning-based models for TSC. The focus is on clarifying their architectures and outlining their adaptations to the specific characteristics of time series data.

\subsection{Multi-Layer Perceptron (MLP)} \label{Apx:MLP}
\begin{figure}[!t]
    \centering
    \includegraphics[trim=1cm 4cm 10cm 1cm, width=0.45\columnwidth]{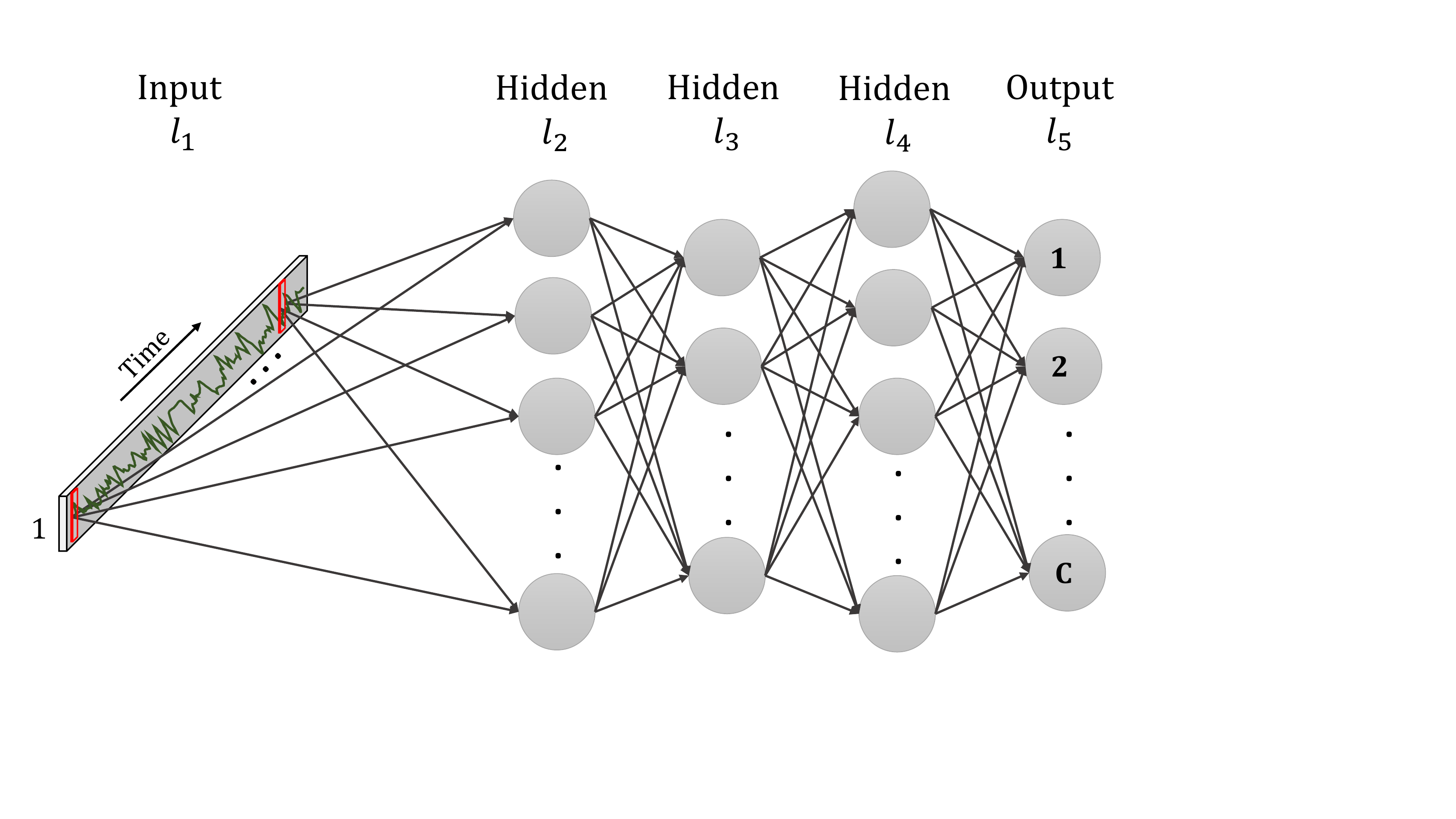}
    \caption{Multilayer perceptron for univariate time series classification.}
    \label{fig:MLP}
\end{figure}

The simplest Neural Network architecture is a fully connected network (FC), also known as a multilayer perceptron (MLP). As shown in Fig. \ref{fig:MLP} all neurons of one layer $l-1$ are connected to all neurons of the following layer $l$ with $l\in [1,L]$. The weights model these connections in a neural network. A general equation for applying a non-linearity to an input
$A^{l-1}$ is: 
\begin{equation}
    A^{l} = f(W^{l}\times A^{l-1} +b)
\end{equation}
where $A^{l}$ the activation of the neurons in layer $l$ where $A^{1}$ is equal to input series $X$. Also, $W$ and $b$ are the neuron weights and biases, and $f$ is the nonlinear activation function.

One of the main limitations of using multilayer perceptrons (MLPs) for time series data is that they are not well-suited to capturing the temporal dependencies in this type of data. MLPs are feedforward networks that process input data in a fixed and predetermined order without considering the temporal relationships between the input values. As shown in Fig. \ref{fig:MLP}, each time step is weighted individually, and time series elements are treated independently from each other.

\subsection{Convolution Neural Networks (CNNs)}
The convolutional neural network (CNN) was first proposed by Kunihiko Fukushima in 1982~\cite{fukushima1982neocognitron}. It was inspired by the structure and function of the visual cortex in animals, specifically the cat's cortex, as described by David Hubel and Torsten Wiesel in their influential work from 1962~\cite{hubel1962receptive}. Convolutional neural networks have been widely used for visual pattern recognition, but their ability to process large images was constrained by computational limitations until the emergence of GPU technology. Following the development of Graphics Processing Unit (GPU) technology, Krizhevsky et al.~\cite{krizhevsky2012imagenet} implemented an efficient GPU-based program and won the ImageNet competition in 2012, bringing the convolution neural network back into the spotlight.

\begin{figure}[!t]
    \centering
    \includegraphics[trim=3cm 3cm 5cm 2cm, width=0.5\columnwidth]{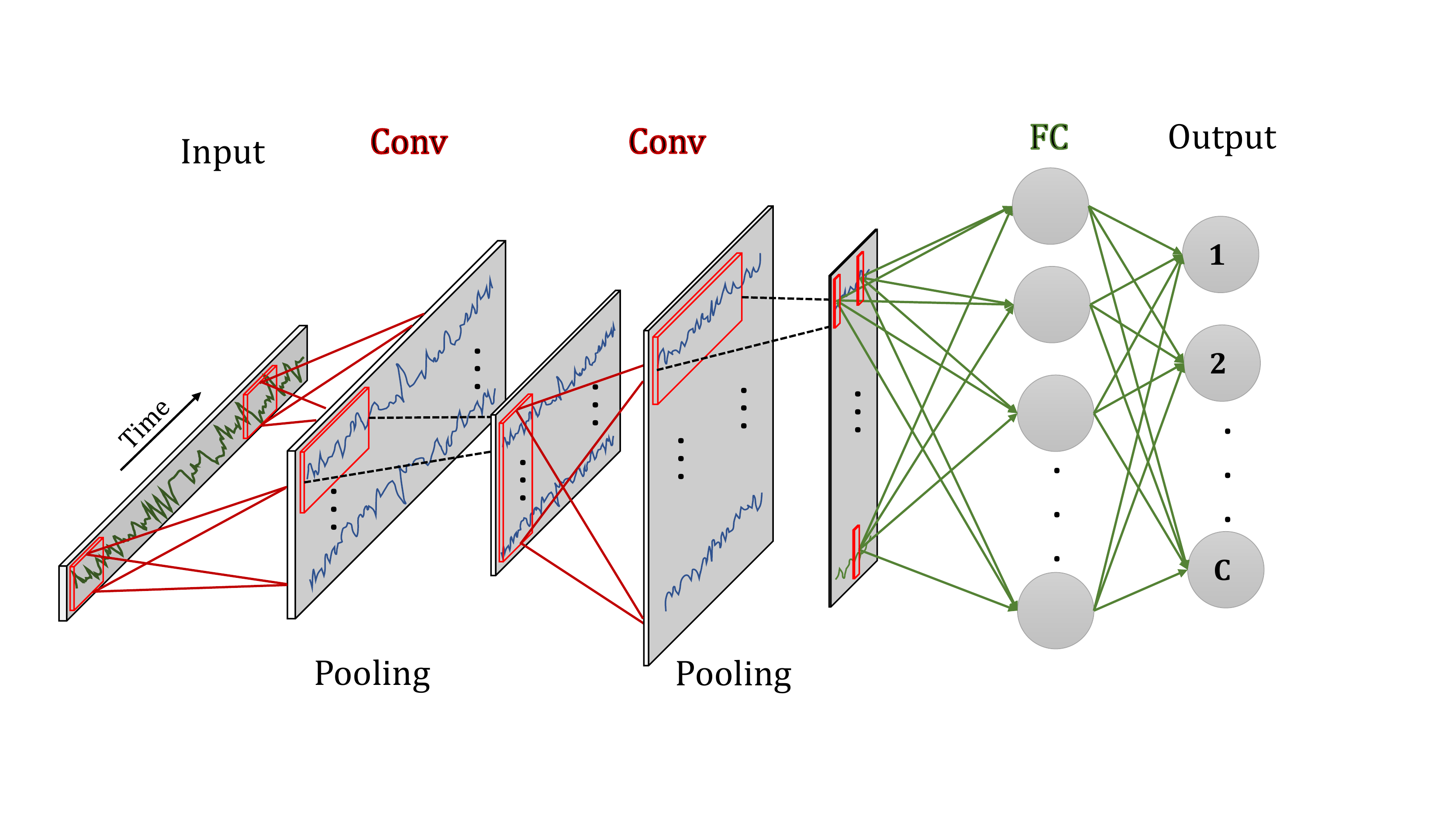}
    \caption{The architecture of the t-LeNet network (time series specific version of LeNet)}
    \vspace{-0.4cm}
    \label{fig:t-LeNet}
\end{figure}

Many variants of CNN architectures have been proposed in the literature, but their primary components are very similar. Using the LeNet-5~\cite{lecun1998gradient} as an example, it consists of three types of layers: convolutional, pooling, and fully connected. The purpose of the convolutional layer is to learn feature representations of the inputs. Fig. \ref{fig:t-LeNet} shows the architecture of the t-LeNet network, which is a time series-specific version of LeNet. This figure shows that the convolution layer is composed of several convolution kernels (or filters) used to compute different feature maps. In particular, each neuron of a feature map is connected to a region of neighboring neurons in the previous layer called the receptive field. Feature maps can be created by first convolving inputs with learned kernels and then applying an element-wise nonlinear activation function to the convolved results. It is important to note that all spatial locations of the input share the kernel for each feature map, and several kernels are used to obtain the entire feature map. 

The feature value of the $l$\textsuperscript{th} layer of $k$\textsuperscript{th} feature map at location $(i,j)$ is obtained by:
\begin{equation}
    Z^l_{i,j,k}={\textbf{W}^l_k}^T \textbf{A}^{l-1}_{i,j}+b^l_k
\end{equation}
Where $\textbf{W}^l_k$ and $b^l_k$ are the weight vector and bias term of the $k$\textsuperscript{th} filter of the $l$\textsuperscript{th} layer, respectively, and $\textbf{A}^{l-1}_{i,j}$ is the input patch centered at location $(i, j)$ of the $l$ layer. Note that the kernel $\textbf{W}^l_k$ that generates the feature map $Z^l_{:,:,k}$ is shared. A weight-sharing mechanism has several advantages, such as reducing model complexity and making the network easier to train. Let $f\left(.\right)$ denote the nonlinear activation function. The activation value  of convolutional feature $Z^l_{i,j,k}$ can be computed as:
\begin{equation}
    \textbf{A}^{l}_{i,j,k} = f(Z^l_{i,j,k})
\end{equation}

The most common activation functions are sigmoid, tanh and ReLU~\cite{nair2010rectified}. As shown in Fig. \ref{fig:t-LeNet}, a pooling layer is often placed between two convolution layers to reduce the resolution of the feature maps and to achieve shift-invariance.
Following several convolution stages $-$the block comprising convolution, activation, and pooling is called convolution $stage$ $-$ there may be one or more fully-connected layers that aim to perform high-level reasoning. As discussed in section \ref{sec:MLP}, each neuron in the previous layer is connected to every neuron in the current layer to generate global semantic information. In the final layer of CNNs, there is the output layer in which the Softmax operators are commonly used for classification tasks~\cite{gu2018recent}.

\subsection{Recurrent Neural Networks (RNN)}
\begin{figure}[!t]
    \centering
    \includegraphics[trim=2cm 1cm 8cm 1cm, width=0.5\columnwidth]{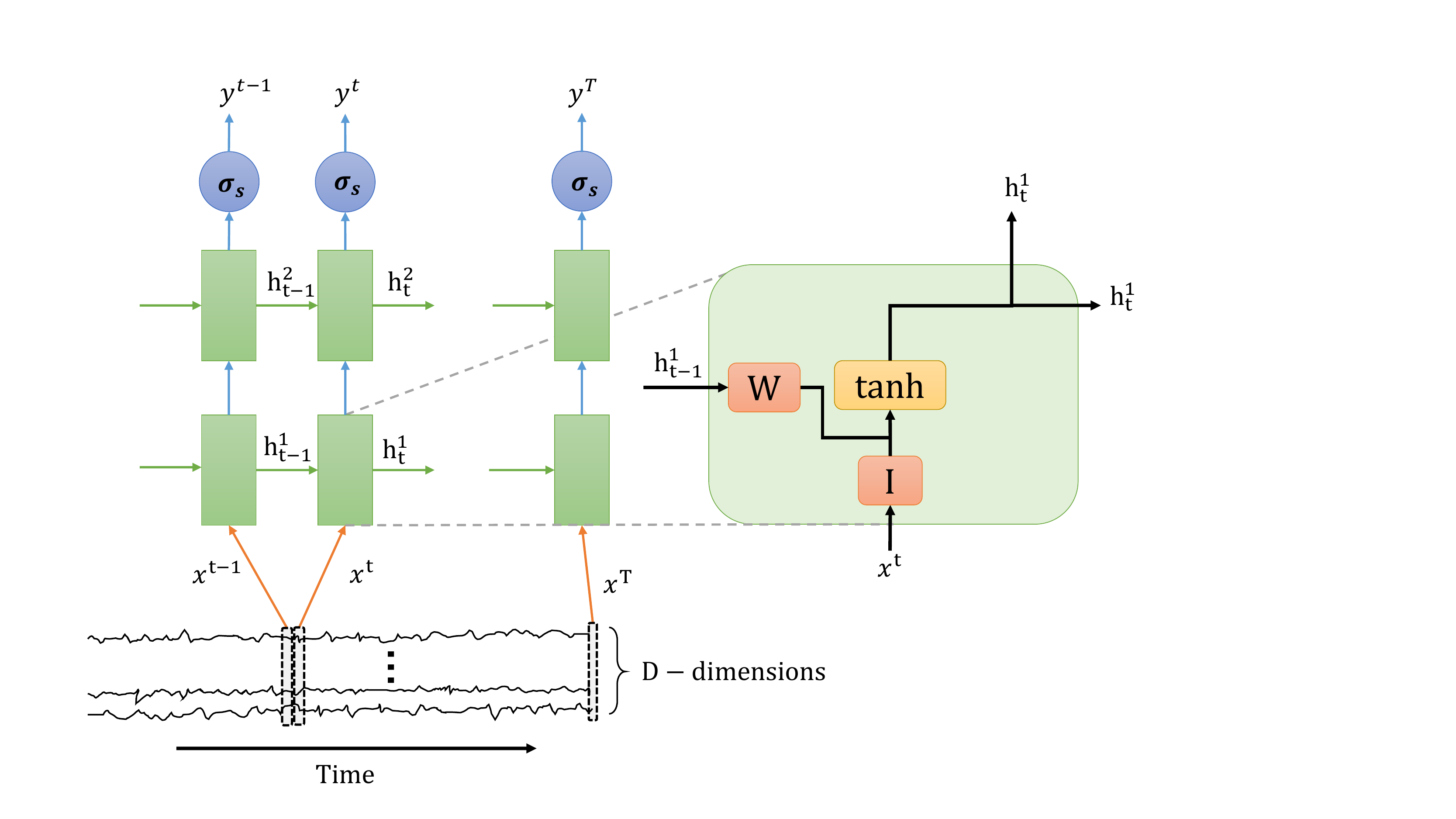}
    \caption{The architecture of two layer Recurrent Neural Network}
    \vspace{-0.5cm}
    \label{fig:RNN}
\end{figure}

 RNNs are types of neural networks that are specifically designed to process time series and other sequential data. RNNs are conceptually similar to feed-forward neural networks (FFNs). While FFNs map from fixed-size inputs to fixed-size outputs, RNNs can process variable-length inputs and produce variable-length outputs. This capability is enabled by sharing parameters over time through directed connections between individual layers. RNN models for TSC can be classified as sequence to sequence or sequence-to-one based on their outputs. Fig. \ref{fig:RNN} shows sequence to sequence architectures for RNN models, with an output for each input sub-series. On the other hand, in sequence-to-one architecture, decisions are made using only $y^T$ and ignoring the other outputs.

At each time step $t$, RNNs maintain a hidden vector $h$ which updates as follows~\cite{hihi1995hierarchical,pascanu2013construct}:
\begin{equation}
    h_t = tanh(Wh_{t-1} + Ix^{t})
\end{equation}
Where $X =\left\{x^1,...,x^{t-1},x^t,...,x^T\right\}$ contains all of the observation, $tanh$ denotes the hyperbolic tangent function, and the recurrent weight and the projection matrix are shown by  $W$ and $I$, respectively. The hidden-to-hidden connections also model the short-term time dependency. The hidden state $h$ is used to make a prediction as:
\begin{equation}
    y^t = \sigma_s(Wh_{t-1})
\end{equation}
where $\sigma_s$ is a softmax function and provides a normalized probability distribution over the possible classes. As depicted in Fig. \ref{fig:RNN}, the hidden state $h$ can be used to stack RNNs in order to build deeper networks:
\begin{equation}
    h_t^l = \sigma(Wh^l_{t-1} + Ih^{l-1}_t )
\end{equation}
where $\sigma$ is the logistic sigmoid function. As an alternative to feeding each time step to the RNN, the data can be divided into time windows of $\omega$ observations, with the option for variable overlaps. Each time window is labeled with the majority response labels within the $\omega$ window. 

\subsubsection{Long Short Term Memory (LSTM)}
LSTM deals with the vanishing/exploding gradient problem commonly found in standard recurrent neural networks through the incorporation of gate-controlled memory cells into their state dynamics~\cite{hochreiter1997long}. As shown in Fig.~\ref{fig:LSTM-GRU}~(a) LSTM uses a hidden vector $h$ and a memory vector $m$ to control state updates and outputs for each time step. Specifically, the computation at time step $t$ is formulated as follows~\cite{kawakami2008supervised}:

\begin{equation} 
\begin{split}
&\Gamma^c = tanh(W^c h_{t-1} + I^c x^t) \\ 
&\Gamma^u = \sigma(W^u h_{t-1} + I^u x^t) \\
&\Gamma^f = \sigma(W^f h_{t-1} + I^f x^t) \\
&\Gamma^o = \sigma(W^o h_{t-1} + I^o x^t) \\
& m_t = \Gamma^f\otimes m_{t-1} + \Gamma^u \otimes \Gamma^c \\
& h_t =tanh(\Gamma^o\otimes m_t)
\end{split}
\end{equation}
where $\Gamma^c$ is a cell state gate and $\Gamma^u,\Gamma^f$ and $\Gamma^o$ are the activation vector of the input, forget and output gate respectively.  $\sigma$ is the logistic sigmoid function and $\otimes$ shows the element-wise product. $W^u, W^f, W^o, W^c$ represent the recurrent weight matrices, and $I^u, I^f, I^o, I^c$ represent the projection matrices. 

\subsubsection{Gated Recurrent Unit (GRU)}
\begin{figure}[!t]
    \centering
    \includegraphics[trim=2cm 6cm 0cm 0cm, width=0.65\columnwidth]{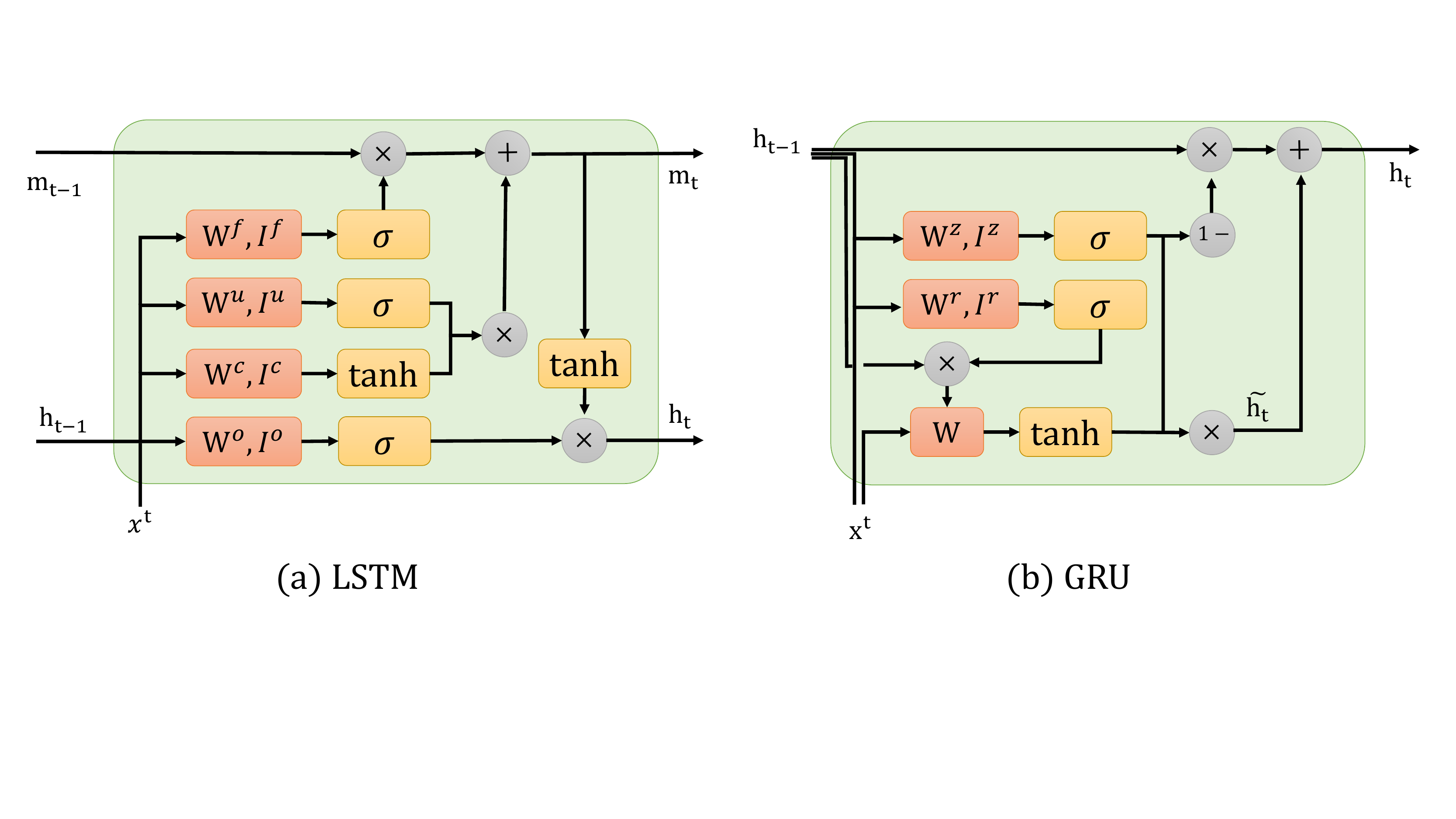}
    \caption{The architecture of LSTM (a) and GRU (b) units}
    \vspace{-0.6cm}
    \label{fig:LSTM-GRU}
\end{figure}

GRU is another widely-used variant of RNNs, and similar to LSTM it can control the flow of information like memorizing the context over multiple time steps~\cite{chung2014empirical}. While GRU was introduced later than LSTM, it has a simpler architecture. As shown in Fig~\ref{fig:LSTM-GRU}~(b), compared to LSTMs, GRUs have two gates, reset and update gates, which are more computationally efficient and require fewer data to generalize and defined as follows: 
\begin{equation} 
\begin{split}
&\Gamma^z = \sigma(W^z h_{t-1} + I^z x^t) \\ 
&\Gamma^r = \sigma(W^r h_{t-1} + I^r x^t) \\
& \widetilde{h}_t = tanh(W [h_{t-1}\otimes\Gamma^r,x^t])\\
& h_t =(1-\Gamma^z)\otimes h_{t-1}+\Gamma^z\otimes \widetilde{h}_t
\end{split}
\end{equation}

Where $W^z$ and $W^r$ are the weight matrices associated with gates, and $\Gamma^z$ and $\Gamma^r$ represent the update and reset gates, respectively. The function $\sigma$ denotes the logistic sigmoid, and $\otimes$ shows the element-wise product.

\subsection{Attention Based Model}
\subsubsection{Self-Attention}
\begin{figure}[!t]
    \centering
    \includegraphics[trim=2cm 3cm 12cm 1cm, width=0.45\columnwidth]{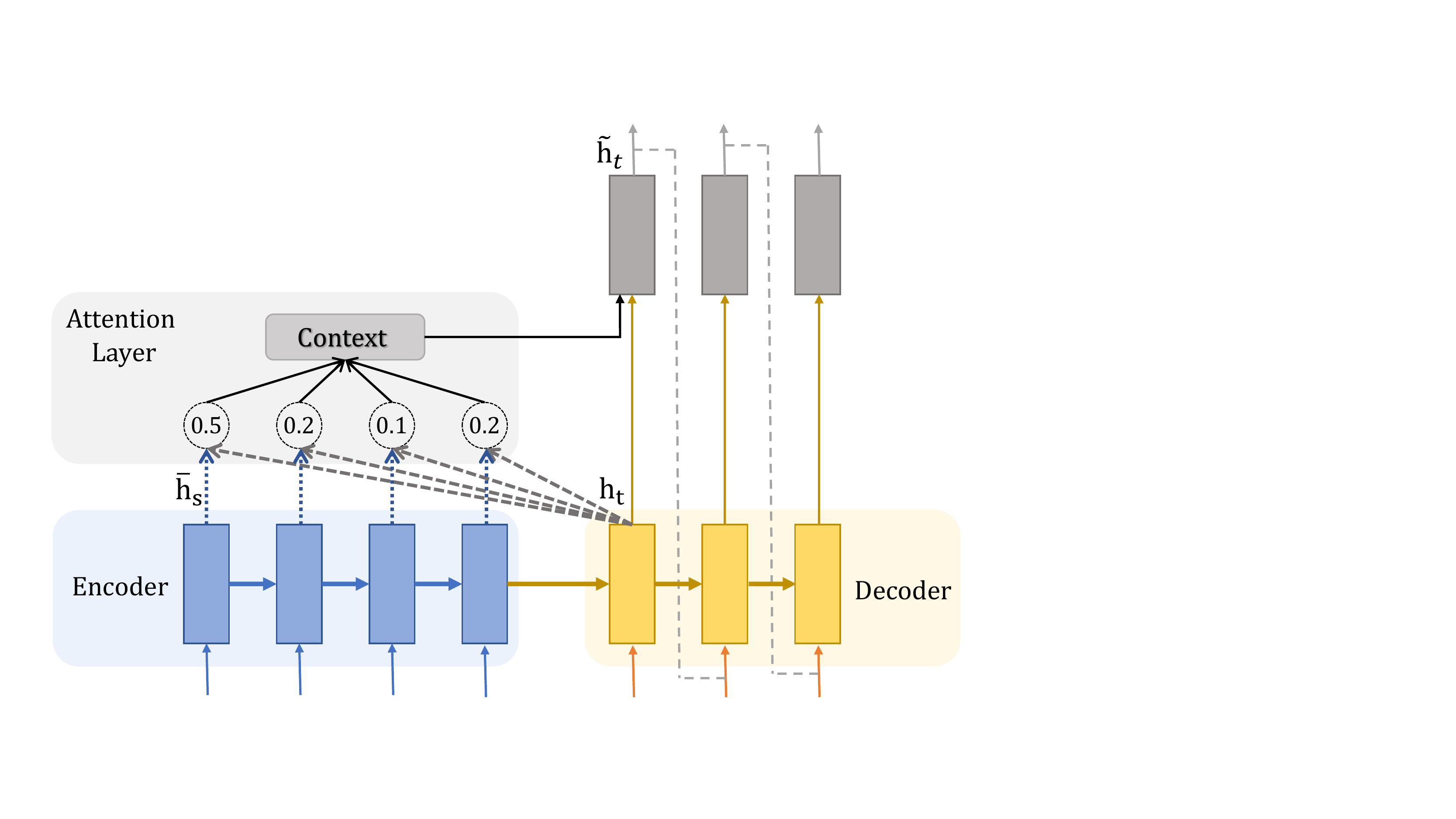}
    \caption{Self-Attention mechanism}
    \vspace{-0.5cm}
    \label{fig:SA}
\end{figure}

The attention mechanism was introduced by~\cite{bahdanau2014neural} for improving the performance of encoder-decoder models~\cite{cho2014learning} in neural machine translation. The encoder-decoder in neural machine translation encodes a source sentence into a vector in latent space and decodes the latent vector into a target language sentence.
As shown in Fig. \ref{fig:SA}, the attention mechanism allows the decoder to pay attention to the segments of the source for each target through a context vector $c_t$. For this model, a variable-length attention vector $\alpha_t$, equal to the number of source time steps, is derived by comparing the current target hidden state $h_t$ with each source hidden state $\overline{h}_s$ as follows~\cite{luong2015effective}:
 \begin{equation}
     \alpha_t(s)= \frac{{exp(score(h_t,\overline{h}_s)})}{\sum_{s'}exp(score(h_t,\overline{h}_{s'}))}
 \end{equation}
The term $score$ is referred to as an alignment model and used to compare the target hidden state $h_t$ with each of the source hidden states $\overline{h}_s$, and the result is normalized to produced attention weights (a distribution over source positions). There are various choices of the scoring function:


\begin{equation}
    score(h_t,\overline{h}_s)= \begin{cases}h_t^TW\overline{h}_s 
    \\ 
    v_{\alpha}^Ttanh(W_{\alpha}[h_t;\overline{h}_s])\end{cases}
\end{equation}
These scores influence the attention distribution, impacting how the model attends to different parts of the input sequence during predictions. As shown above, the score function is parameterized as a feedforward neural network that is jointly trained with all the other components of the model. The model directly computes soft attention, allowing the cost function's gradient to be backpropagated~\cite{bahdanau2014neural}. 

Given the alignment vector as weights, the context vector $c_t$ is computed as the weighted average over all the source hidden state:
\begin{equation}
    c_t = \sum_{s}\alpha_{ts}\overline{h}_s
\end{equation}
Accordingly, the computation path goes from $h_t\rightarrow \alpha_t \rightarrow c_t \rightarrow \widetilde{h}_t$ then make a prediction using a $Softmax$ function~\cite{luong2015effective}. Note that $\widetilde{h}_t$ is a refined hidden state that incorporates both the original hidden state $h_t$ and the context information $c_t$ obtained through attention mechanisms.

\subsubsection{Transformers}
Similar to self-attention and other competitive neural sequence models, the original transformer developed for NLP (hereinafter the vanilla transformer) has an encoder-decoder structure that takes as input a sequence of words from the source language and then generates the translation in the target language~\cite{vaswani2017attention}.
Both the encoder and decoder are composed of multiple identical blocks. Each encoder block consists of a multi-head self-attention module and a position-wise feed-forward network (FFN), while each decoder block inserts cross-attention models between the multi-head self-attention module and the position-wise feed-forward network (FFN). Unlike RNNs, Transformers do not use recurrence and instead model sequence information using the positional encoding in the input embeddings.

\begin{figure}[!t]
    \centering
    \includegraphics[trim=1cm 1cm 2cm 0cm, width=0.7\columnwidth]{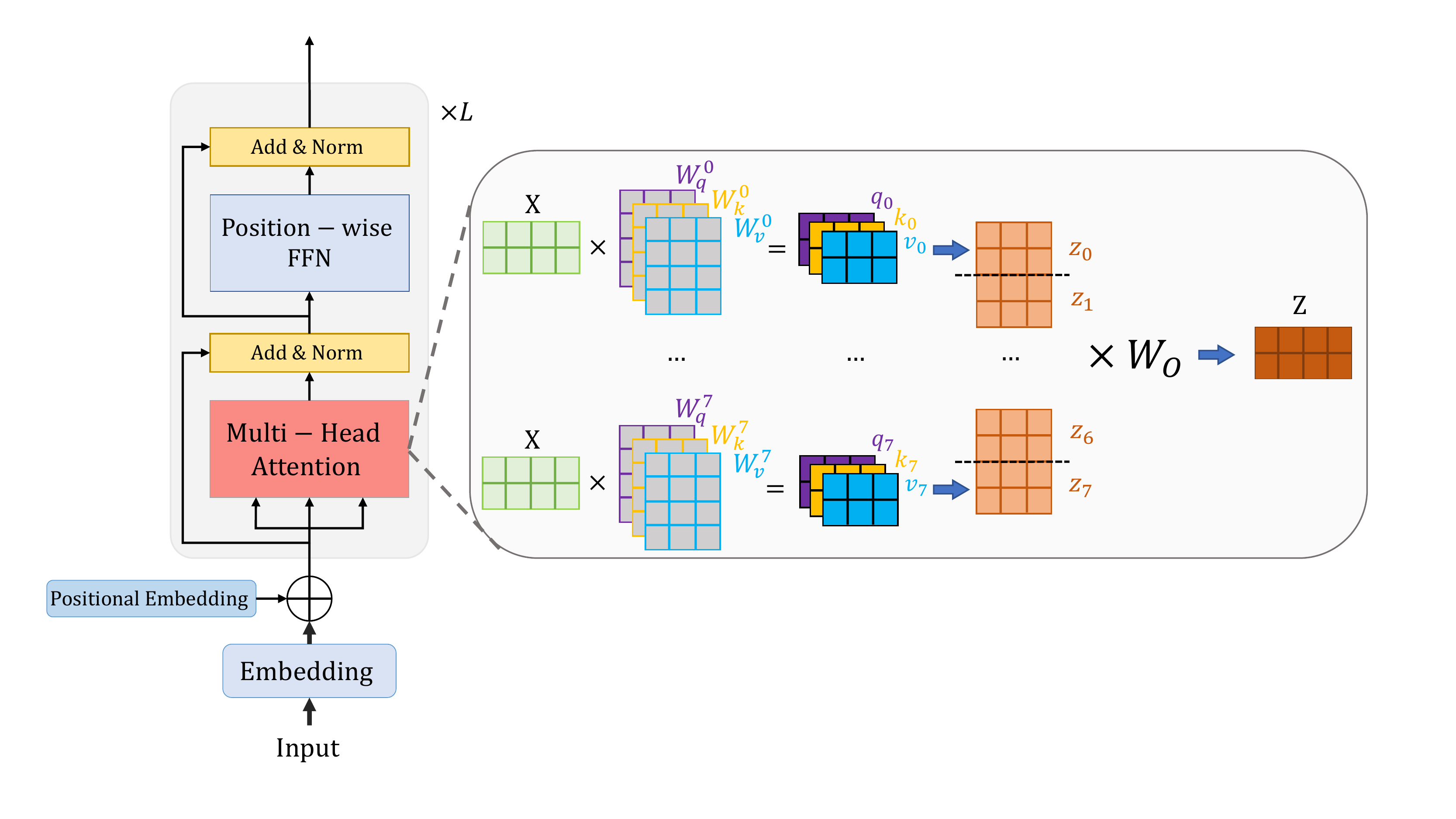}
    \caption{Multi-head attention block: the example consists of eight heads, and the input sequence comprises two time steps.}
    \vspace{-0.4cm}
    \label{fig:Multi-Head}
\end{figure}

The transformer architecture is based on finding associations or correlations between various input segments using the dot product. As shown in Fig. \ref{fig:Multi-Head}, the attention operation in transformers starts with building three different linearly-weighted vectors from the input $x_i$, referred to as query ($q_i$), key ($k_i$), and value ($v_i$):
\begin{equation}
    \textbf{q}_i = W_q\textbf{x}_i , \quad 
    \textbf{k}_i = W_k\textbf{x}_i , \quad 
    \textbf{v}_i = W_v\textbf{x}_i
\end{equation}
where $W_q,W_k$ and $W_v$ learnable weight matrices. The output vectors $\textbf{z}_i$ are given by:
\begin{equation}
\textbf{z}_i=\sum_{j}softmax\left(\frac{\textbf{q}_i^T\textbf{k}_j}{\sqrt{d_q}}\right)\textbf{v}_i
\end{equation}
Note that the weighting of the value vector $\textbf{v}_i$ depends on the mapped correlation between the query vector  $\textbf{q}_i$ at position $i$ and the key vector $\textbf{k}_j$ at position $j$. The value of the dot product tends to grow with the increasing size of the query and key vectors. As the softmax function is sensitive to large values, the attention weights are scaled by the square root of the size of the query and key vectors $d_q$.
The input data may contain several levels of correlation information, and the learning process may benefit from processing the input data in multiple different ways. Multiple attention heads are introduced that operate on the same input in parallel and use different weight matrices $W_q$,$W_k$, and $W_v$ to extract various levels of correlation between the input data.

\subsection{Graph Neural Networks}
A graph consists of a set of nodes and a set of edges, each of which connects two nodes. Both nodes and edges may have attributes associated with them. The edges may be directional or unidirectional, and may be weighted. Graphs are useful for representing data that cannot be represented in Euclidean space, such as molecular structures, social networks, and spatial temporal data (for example, electroencephalogram (EEG) or traffic monitoring networks). 

Graph neural networks (GNNs) were first proposed by Scarcelli et al.~\cite{Scarselli2009graph} to learn directly from graph representations of data. Prior to the use of GNNs, techniques such as recursive neural networks and Markov chains (random walk models) were used to incorporate graph structures. However, these methods required a pre-processing step, rather than learning directly from the graph structure. GNNs take as input the graph structure and any associated node and edge attributes. Depending on the required task, the GNN output can be per node, per edge or a single output per graph~\cite{Wu2021graph}.

Scarcelli et al.’s proposed GNN combines recursive neural networks and Markov chains to deal directly with the graph structure without any pre-processing requirement. While the network structure is predefined, the edge weights are parameters learned during training. During training, units exchange information and update their states until reaching equilibrium.

Graph convolutional networks (GCN) were proposed by Bruna et al.~\cite{Bruna2013graph} and extend CNNs to graph structures. Bruna et al. proposed two methods of constructing the graph, spatial and spectral. The spatial technique simply applies the convolution operator to the local neighbourhood of each node, followed by a pooling operator. Although this reduces the spatial resolution, successive layers compensate for this by increasing the number of filters. Spectral construction first transforms the graph to a matrix V, which consists of the eigenvectors of the graph Laplacian, order by eigenvalue. The eigenvectors represent frequency components of the original graphs, so lower-order eigenvectors modulate slowly, thus neighbouring nodes have similar values. Higher-order eigenvectors modulate more rapidly and connected nodes are likely to have dissimilar values~\cite{Shuman2013graph}.

Many real-world graph datasets evolve over time – edges and nodes may come into existence, disappear or attributes may change value. Dynamic or temporal graphs build this information into the graph structure, for instance by the use of temporal nodes and edges, that include initial and final timestamps~\cite{Longa2023graph}. Alternatively, spatiotemporal GNNs model the spatial and temporal aspects in separate layers, using GCN layers to learn spatial representations and RNN or 1D-CNN layers for the temporal representations~\cite{Wu2021graph}.


\section{Datasets}
\subsection{HAR Datasets} \label{har-datasets}
Many of the studies reviewed in the subsection reviewing the use of deep learning for human activity recognition time series use publicly available datasets. Some of the most commonly used datasets are listed in table \ref{table:hardata}, together with the number of participants, the sensors used to collect the data, a description of the activities recorded, and references to the studies using each dataset. Larger lists of datasets are provided in~\cite{chen2021deep,Ramanujam2021survey}. Common activity sets include activities of daily living (ADL) or basic activities (e.g. walking, running, sitting, standing, ascending/descending stairs. However, activities for more specialised events such as gait freezing in Parkinson's Disease patients~\cite{Bachlin2009daphnet}, falls~\cite{Micucci2017unimib}, and manufacturing activities~\cite{Zappi2008skoda} are also collected.
{
\setlength{\tabcolsep}{3pt}
\begin{table}[!htp]
\begin{tabulary}{\textwidth \tymin=2cm}{L p{1.3cm} L L L}
\hline
 \bfseries{Name} & \bfseries{Partici- \newline pants} & \bfseries{Sensors} & \bfseries{Activities/ \newline Description} & \bfseries{Used by} \\
\hline
 \noalign{\smallskip}
 DAPHNet FoG~\cite{Bachlin2009daphnet} & 10 & 3 accelerometers attached to the body & Detection of freezing of gait events in Parkinson's Disease patients. &~\cite{Murad2017deep,Zeng2018rnn,Hammerla2016deep} \\
 \noalign{\smallskip}
 Opportunity challenge~\cite{Chavarriaga2013opp,Roggen2010opp} & 4 & 5 on-body IMUs each with 3 sensors & A variety of activities of daily living. &~\cite{Ordonez2016deep,Wang2022deep,Rueda2018cnn,Zeng2014cnn,Xu2022deform}
~\cite{Yao2018cnn,Murad2017deep,Guan2017ensemble,Yang2015cnn,Hammerla2016deep} \\
 \noalign{\smallskip}
 PAMAP2~\cite{Reiss2012pamap,Reiss2012pamap2} & 9 & 3 IMUs and heart rate monitor & 18 basic and daily living activities. & ~\cite{Wang2022deep,Rueda2018cnn,Zeng2018rnn,Guan2017ensemble}~\cite{Challa2021multi,Chen2021har,Ma2019attnsense,Hammerla2016deep}\\
 \noalign{\smallskip}
 Skoda~\cite{Zappi2008skoda} & 1 & 19 body-worn sensors & 10 quality assurance activities from car production plant. &~\cite{Ordonez2016deep,Zeng2014cnn,Murad2017deep}~\cite{Zeng2018rnn,Guan2017ensemble,Ma2019attnsense} \\
 \noalign{\smallskip}
 UCI-HAR~\cite{Anguita2013ucihar} & 30 & Smartphone accelerometer and gyroscope & 6 basic activities. &~\cite{Wang2022deep,Ronao2016har,Ignatov2018realtime,Jiang2015cnn,Murad2017deep}~\cite{Challa2021multi,Nafea2021sensor,Mekruksavanich2021lstm,Mekruksavanich2021biometric} \\
 \noalign{\smallskip}
 UniMiB SHAR~\cite{Micucci2017unimib} & 30 & Smartphone accelerometers & Activities of daily living and falls. &~\cite{Wang2022deep,Xu2022deform} \\
 \noalign{\smallskip}
 USC-HAD~\cite{Zhang2012uschad} & 14 & Motion node (accelerometer, gyroscope, and magnetometer) attached to hip & 12 basic activities. &~\cite{Jiang2015cnn,Xu2022deform,Murad2017deep,Singh2021convlstm,Mekruksavanich2021biometric} \\
 \noalign{\smallskip}
 WISDM~\cite{Kwapisz2011wisdm} & 29 & Smartphone accelerometer & 6 basic activities. &~\cite{Wang2022deep,Ignatov2018realtime,Zhang2020novel,Xu2022deform}~\cite{Singh2021convlstm,Challa2021multi,Nafea2021sensor} \\
\hline
\noalign{\smallskip}
\end{tabulary}
\caption{Commonly used human activity recognition datasets}
\label{table:hardata}
\end{table}
}

\subsection{Earth observation satellites and instruments} \label{earth-observation}
Table \ref{table:satellites} lists the main satellites and instruments used in the studies reviewed for this survey. The table lists references for each source, which provide more details about the data collected, plus a list of the studies using each source.

\begin{table}[!htp]
\noindent
\makebox[\textwidth]{
\begin{tabular}{ p{2.8cm} p{4cm} p{1.7cm} p{4.3cm} }
\hline
 \noalign{\smallskip}
 \bfseries{Agency} & \bfseries{Satellite \& Instruments} & \bfseries{Type} & \bfseries{Used by} \\
\hline
 \noalign{\smallskip}
 NASA           & Landsat-7 ETM+~\cite{USGALandsat}                 & Optical &~\cite{Xu2020dcm,Dou2021tsi}  \\
 NASA           & Landsat-8 OLI~\cite{USGALandsat}                  & Optical &~\cite{Xu2020dcm,Dou2021tsi,Li2019tan,Li2020tga,Kussul2017cnn}, \newline~\cite{DiMauro2017tiselc,Matosak2022forest,Rao2020lfmc,Xie2022lfmc} \\  
 NASA           & Terra/Aqua MODIS~\cite{NASAMODIS}                 & Optical &~\cite{Xie2022lfmc,Zhu2020lfmc,Miller2022lfmc,Sun2020yield,Qiao2021yield} \\
 ESA            & Sentinel-1A/1B \newline SAR-C~\cite{ESA2019Sentinel}       & Microwave \newline SAR &~\cite{Kussul2017cnn,Rao2020lfmc,Xie2022lfmc,Ofori-Ampofo2021att,Ienco2019twinns,Gbodjo2020hob2srnn}, \newline~\cite{Ienco2019od2rnn,Ban2020fire,Rambour2020flood,Kulshrestha2022hole,Minh2017vege} \\
 ESA            & Sentinel-2A/2B MSI~\cite{ESA2019Sentinel}         & Optical &~\cite{Li2019tan,Li2020tga,Matosak2022forest,Ofori-Ampofo2021att,Ienco2019twinns,Ienco2019od2rnn,Rambour2020flood}, \newline~\cite{Garnot2019time,Garnot2020tae,Garnot2020ltae,Yuan2021sitsbert,Ienco2020tassel,Interdonato2019duplo,Barriere2022lstm},~\cite{Russwurm2018seqrnn,Russwurm2020att,Yuan2022sitsformer,Stoian2019fgunet,Labenski2022under,KamdemDeTeyou2020road,Lahssini2022wood} \\
 CNES (France)  & Pléiades-1A/1B HiRI~\cite{ESAPleiades}            & Optical &~\cite{Ienco2017rnn} \\
 NSPO (Taiwan)  & Formosat-2 RSI~\cite{NSPO2020Formosat}            & Optical &~\cite{Pelletier2019tempcnn} \\
 CRESDA (China) & Gaofen-1/2 MUX, PAN, \newline WFV~\cite{EoPortal2014Gaofen1,EoPortal2015Gaofen2} & Optical &~\cite{Ji20183dcnn} \\
 \noalign{\smallskip}
\hline
 \noalign{\smallskip}
\end{tabular}}
\caption{Earth Observation satellites and instruments collecting the data used in the studies reviewed.}
\label{table:satellites}
\end{table}

\end{document}